\documentclass[journal,oneside]{IEEEtran}
\usepackage{graphicx,subfigure,amssymb,bm,color,listings,url}
\usepackage[utf8]{inputenc}
\graphicspath{{images/}}

\usepackage{cite}
\usepackage[cmex10]{amsmath} 
\renewcommand{\Re}{\mathbb{R}}
\renewcommand{\vec}[1]{\ensuremath{\boldsymbol{#1}}}
\newcommand{\mat}[1]{\ensuremath{\boldsymbol{#1}}}
\renewcommand{\exp}[1]{\ensuremath{ e^{#1}}}

\title{Locally Orderless Registration}
\author{Sune Darkner and Jon Sporring %
\thanks{Manuscript submitted to IEEE TPAMI April 2012}}
\begin{document}

\maketitle

\begin{abstract}
  Image registration is an important tool for medical image analysis and is used to bring images into the same reference frame by warping the coordinate field of one image, such that some similarity measure is minimized. We study similarity in image registration in the context of Locally Orderless Images (LOI), which is the natural way to study density estimates and reveals the 3 fundamental scales: the measurement scale, the intensity scale, and the integration scale.

This paper has three main contributions:  Firstly, we rephrase a large set of popular similarity measures into a common framework, which we refer to as Locally Orderless Registration, and which makes full use of the features of local histograms.  Secondly, we extend the theoretical understanding of the local histograms.  Thirdly, we use our framework to compare two state-of-the-art intensity density estimators for image registration: The Parzen Window (PW) and the Generalized Partial Volume (GPV), and we demonstrate their differences on a popular similarity measure, Normalized Mutual Information (NMI). 

We conclude, that complicated similarity measures such as NMI may be evaluated almost as fast as simple measures such as Sum of Squared Distances (SSD) regardless of the choice of PW and GPV.  Also, GPV is an asymmetric measure, and PW is our preferred choice.
\end{abstract}


\section{Introduction}
\label{sec:introduction}
\IEEEPARstart{I}{mage} similarity measures are crucial components in image registration, and Mutual information (MI) \cite{collignon1995,wells1996mmv} and Normalized Mutual Information (NMI) \cite{studholme1999oie} are considered state-of-the-art for image registration. MI and NMI are particularly useful for registering Magnetic Resonance Images (MRI) to MRI as well as for multi-modal image registration in general. MI and NMI are entropy based measures and hence rely on probability distributions. Probability distributions are most often approximated by discrete histograms, which poses a challenge to gradient based optimization schemes. The most common estimation techniques are: The Parzen Window (PW) \cite{wells1996mmv} and the Generalized Partial Volume (GPV) \cite{maes1997mir,chen2003mutual}. Empirical comparisons have previously been presented \cite{loeckx2006comparison}, and recently we investigated their theoretical connection \cite{darkner.sporring.ipmi2011}.

In this paper, we present Locally Orderless Registration (LOR).  LOR is a framework for performing registration of $N$-dimensional images, and it includes a common framework for a wide range of image similarity measures such as Correlation Ratio, MI NMI, Huber Norm etc..  
%These as well as our generalizations are discussed in Appendix~\ref{sec:dissimilarityMeasures}. 
The framework is centered around local histograms, and we use Locally Orderless Images (LOI) \cite{griffin97,koenderink1999structure}, which makes the 3 natural scale parameters available for image registration: measurement scale -- smoothing of the initial image, intensity scale - smoothing of the histogram, and integration scale - the local spatial extend of local histograms.  These 3 scales interact in a nontrivial manner, and we explore their relation theoretically by the local intensity moments as well as on a simple local image model.  We perform extensive empirical investigations on the influence of the scales on the density estimates as well as NMI.  We also summarize and extend our earlier theoretical work \cite{darkner.sporring.ipmi2011}, where LOR is used to compare PW and GPV, and we demonstrate both theoretically as well as empirically that GPV is asymmetric, and therefore the less preferred choice of the two.  Finally, we present a unifying algorithm for PW and GPV for various measures, and we present analytical as well as empirical investigations of its computational complexity. Timing results on our algorithm shows that NMI is almost as fast as Sum of Squared Differences (SSD), and that multi-threaded implementation only has 13\% overhead compared to the theoretical computational speed.

\subsection{Previous work}
The use of Mutual Information  for image registration was originally proposed by \cite{collignon1995,wells1996mmv}. An extensive overview was given in \cite{pluim2003mutual}.  Normalized Mutual Information was introduced as a more robust alternative especially designed for multi-modal image registration \cite{studholme1999oie}.  The first implementations relied on Powell's method \cite{maes1997mir}, hill climbing \cite{studholme1999oie}, or similar methods without gradients, which were accurate but slow. A GPU speed-up was suggested in \cite{modat2009fast}. Today, state-of-the-art implementations are gradient based methods and group in two algorithm types: The first type is based on PWs \cite{wells1996mmv} and relies on the fact that the marginal and the joint histograms are made continuous by using different kernels, e.g., Gaussian or B-splines \cite{thevenaz2000omi}. The second type is based on GPV, where the distribution is sampled from the image directly \cite{maes1997mir}.  Analytical derivatives of this method were presented in \cite{maes1999comparative} and a generalization using B-splines was presented in \cite{chen2003mutual}.  A variational method relating to Locally orderless images \cite{koenderink1999structure} for MI (and other measures) was presented in \cite{hermosillo2002variational}. GPV and PW was compared numerically in \cite{loeckx2006comparison} concluding that PW is precise and GPV has a larger convergence radius. MI and NMI are notorious for their local minima and difficulty of implementation, and the choice of interpolation scheme greatly influences the smoothness of the objective function. Some investigations into this can be found in \cite{pluim2000interpolation,haber2007intensity}.  An alternative approach is the Conditional Mutual Information \cite{loeckx2007nonrigid}.  In \cite{darkner.sporring.ipmi2011} we investigated PW and GPV for NMI, using differential calculus in a thorough step-by-step presentation. The derivations was an alternative to the variational approach in \cite{hermosillo2002variational}, and our approach revealed much faster algorithms and a direct comparison between PW and GVP. \cite{hermosillo2002variational} allowed for a local variant of MI, which was implemented in\cite{ourselin2011tmi}. Furthermore, a density alternative though computational complex estimation scheme was suggested in \cite{rajwade2009probability} but is however unsuited for fast gradient based optimization schemes.

 The remainder of this article is organized as follows: In Section~\ref{sec:registration} the general registration framework is described.  In Section~\ref{sec:locallyOrderlessImages} we revisit Locally orderless images as a basis for analysing GPV and PW as well as discuss relation between scales for local histograms.  In Section~\ref{sec:pwgpvCompare} we provide a theoretical comparison between GPV and PW, and in Section~\ref{sec:asym} we augment the theoretical comparison with empirical demonstrations of the asymmetry.  In Section~\ref{sec:scales}, we discuss empirical relations between scales.  In Section~\ref{sec:implementation} we present a fast algorithm for computing PW and GPV for a large range of similarity measures, and in Section~\ref{sec:conclusion} we summarize our findings and conclude on our work.

\section{Image registration}
\label{sec:registration}
Image registration is the process of transforming one image $\tilde{I}:\Omega\rightarrow\Gamma$, where $\Omega\subseteq\Re^N$ and $\Gamma\subseteq\Re$, w.r.t.\ a reference image $R:\Omega\rightarrow\Gamma$ such that some functional $\mathcal F(\tilde{I},R)$ is minimized. We consider diffeomorphic transformation of $M$ parameters, $\vec \phi:\Omega\times\Re^M\rightarrow\Omega$, and for brevity we write $ I = \tilde{I} \circ \vec \phi$.  We consider functionals,  $\mathcal F$, of the form,
\begin{align}
  \label{eq:F}
  \mathcal{F}&=\mathcal{M}(I,R)+\mathcal{S}(\vec \phi),
\end{align}
where $\mathcal{M}$ is a (dis-)similarity measure and $\mathcal{S}$ is a regularization term. Typical forms of $\mathcal{S}$ is elasticity \cite{darknerSCIA}, fluid \cite{christensen1994dsm} or the recent Kernel Bundle LDDMM \cite{sommer2011}. Our focus is solely on $\mathcal{M}$.

\subsection{The Similarity Mearsure}
Many similarity measures are on the form of,
\begin{align}
  \label{eq:SpaceIntegral}
  \mathcal{M}_\Omega=\int_{\Omega}F\big(\vec x,I(\vec x),R(\vec x)\big)\,d\vec x,
\end{align}
where we in this article make the distinction between differentials as the element wise differentials and the hyper volume elements $d\vec x = d x_1\wedge \dots \wedge dx_N$ used here for integration.   A popular choice of the loss-function, $F$, are monomials, $F(I(\vec x), R(\vec(x))) = (I(\vec x)-R(\vec x))^q$ for $q>0$.  Other similarity measures have the form of,
\begin{align}
  \label{eq:IntensityIntegral}
  \mathcal{M}_\Gamma=\int_{\Gamma^2}F\big(\vec x, i,j, h_{I,R}(i,j)\big)\,di\wedge dj,
\end{align}
where $h_{I,R}:\Gamma^2\rightarrow\Re_+$ is the joint histogram of image $I$ and $R$ with intensity variables $i$ and $j$.  A popular choice is mutual information \cite{shannon1948}, $\mathcal{M}_\text{MI}  = \mathcal{H}_I +\mathcal{H}_R-\mathcal{H}_{I,R}$, where $\mathcal{H}$ is the (joint) entropy, in which case $F=p(i,j)\ln p(i,j)-p(i)\ln p(i)-p(j)\ln p(j)$.  The natural logarithm is often used for convenience, and the distribution $p$ is the normalized (joint) histogram $p(i,j)=h(i,j)/\int_{\Gamma^2} h(i,j)\,di\wedge dj$, such that $p(i)=\int_\Gamma p(i,j)\,dj$, and $p(j) =\int_\Gamma p(i,j)\,di$.

A seemingly major difference between~\eqref{eq:SpaceIntegral} and~\eqref{eq:IntensityIntegral} is the integration domain.  However, we will show that by reordering the integral by the distribution of $I$ and $R$ values, we may rewrite~\eqref{eq:SpaceIntegral} in terms of local histograms $h(\vec x,i,j)$.  This has several advantages: 1) It creates a common form for both classes of similarity measures. 2) The histogram perspective reveals the 3 fundamental scales of images: intensity, measure, and integration.  3) The loss-function $F$ for q-norms and similar becomes linear w.r.t.\ the transformation parameters. 4) With the use of smooth kernels, the derivatives w.r.t.\ space as well as intensity are trivial, and thus are readily available for gradient descent schemes.  There is, however, a minor disadvantage: Continuous histograms have poles, corresponding to image values, where the spatial gradient of the image is zero.  In practice this is of little importance, since we consider generic images, i.e., images whose structure is stable w.r.t.\ negligible noise, and for such images, the set of areas with zero gradients are singular points with measure zero.  We will assume that the poles in the histograms likewise have zero measure, which is supported by our observations, but which we leave to be proven in later work. 

A wide range of loss-functions, $F$, linear as well as non-linear can be formulated in our common framework. We refer to this framework as the Locally Orderless Registration framework (LOI registration), and it has the form, 
\begin{align}
  \label{eq:pdnonlinearIntegral}
  \mathcal{M}=\int_{\Omega\times\Gamma^2} F\big(\vec x, i,j,h_{I,R}(\vec x,i,j)\big)\,d\vec x\wedge di\wedge dj.
\end{align}
Most functionals in the literature are positional independent, wherefore we will adopt the same focus, and further we denote the remainder as either $\mathcal{M}_\text{lin}$ or $\mathcal{M}_\text{nlin}$.  The similarity measure, $\mathcal{M}_\text{lin}$, uses a position independent, linear loss-functions,
\begin{align}
  \label{eq:linearIntegral}
  \mathcal{M}_\text{lin}=\int_{\Gamma^2}F(i,j)h_{I,R}(i,j)\,di\wedge dj.
\end{align}
This measure includes~\eqref{eq:SpaceIntegral} with any positional independent loss-function $F$ such as monomials, it is linear in $h$ w.r.t.\ $F$ and $h$, and the transformation parameters only influences $h$.  The similarity measure, $\mathcal{M}_\text{nlin}$, uses a position independent, non-linear loss-functions,
\begin{align}
  \label{eq:nonlinearIntegral}
  \mathcal{M}_\text{nlin}=\int_{\Gamma^2} F\big(h_{I,R}(i,j)\big)\,di\wedge dj,
\end{align}
where $F$ now denotes some non-linear functional, and this form includes Mutual Information.  As will be shown later, the added complexity from linear to non-linear measures has little influence on computation time.

Position independent, linear loss-functions, $\mathcal{M}_\text{lin}$,
%  have the form,
% \begin{align}
%   \label{eq:linearIntegral2}
%   \mathcal{M}_\text{lin}=\int_{\Gamma^2}F(i,j)h_{I,R}(i,j)\,di\wedge dj,
% \end{align}
% where $h:\Gamma^2\rightarrow\Re_+$ is the joint histogram or co-occurrence matrix of intensity values in $I$ and $J$.
%Such functionals and our generalizations 
are all linear in $h$ w.r.t.\ transformation parameters.  Examples are, $q\geq0$:
\begin{align}
&F_{\ell_q}(i,j)=|i-j|^q,\\
&F_\text{q-hinge}(i,j)=
\begin{cases}(|i-j|-k)^q & \text{if } |i-j|>k, \\ 0 & \text{otherwise},\end{cases}\\
&F_\text{q-Huber}(i,j)=
\begin{cases}|i-j|^q & \text{\hspace*{-5mm}if } |i-j|<k,\\ qk^{q-1}(i-j)-(q-1)k^q  & \text{otherwise},\end{cases}\\
&F_\text{q-trunc}(i,j)=
\begin{cases}|i-j|^q & \text{if } |i-j|<k,\\ k^q  & \text{otherwise}.\end{cases}
\end{align} 

Position independent, non-linear loss-functions, $\mathcal{M}_\text{nlin}$,
%  are written as
% \begin{align}
%   \label{eq:nonlinearIntegral2}
%   \mathcal{M}_\text{nlin}=\int_{\Gamma^2} F\big(h_{I,R}(i,j)\big)\,di\wedge dj,
% \end{align}
%where $F$ now denotes some non-linear functional.  As will be shown later, typical non-linearity has little influence on computation time.  These measures 
include $\mathcal{M}_\text{lin} $ as well as mutual information (MI), normalized mutual information (NMI), and Correlation Coefficient (CC),
\begin{align}
  \mathcal{M}_\text{MI}
  &=\mathcal{H}_I +\mathcal{H}_R - \mathcal{H}_{I,R},\\
  \label{eq:MI}
  \mathcal{M}_\text{NMI} &=\frac{\mathcal{H}_I +\mathcal{H}_R}{\mathcal{H}_{I,R}},\\
  \label{eq:NMI2}
  \mathcal{M}_\text{CC}&=\int_{\Gamma^2} \frac{(i-\mu_I)(j-\mu_R)}{\sigma_I \sigma_R} p_{I,R}\,di\wedge dj,
\end{align}
where $\mu_I=\int_{\Gamma^2} i p_{I,R}(i,j)\,di\wedge dj$, $\sigma_I^2=\int_{\Gamma^2} (i-\mu_I)^2p_{I,R}(i,j)\,di\wedge dj$, and similarly for $\mu_R$ and $\sigma_R$, and where $\mathcal{H}$ denotes the marginal and the joint entropy of the intensity distribution \cite{shannon1948},
\begin{align}
  \mathcal{H}_I &=- \int_\Gamma p_I(i) \ln p_I(i)\,di,\\ 
  \mathcal{H}_R &=-\int_\Gamma p_R(j) \ln p_R(j)\,dj,\\
  \mathcal{H}_{I,R} &=-\int_{\Gamma^2} p_{I,R}(i,j)\ln p_{I,R}(i,j)\,di\wedge dj,
\end{align}
The distributions are obtained by normalizing the histograms to unity,
\begin{align}
  p(i) &\simeq \frac{h(i)}{\int_\Gamma h(j) dj},\\
  p(i,j) &\simeq \frac{h(i,j)}{\int_{\Gamma^2} h(k,l)\,dk\wedge dl}.
\end{align}

Position dependent, non-linear loss-functions, $\mathcal{M}$,
%are given as,
%\begin{align}
%  \label{eq:pdnonlinearIntegral2}
%  \mathcal{M}=\int_{\Omega\times\Gamma^2} F\big(\vec x, h_{I,R}(\vec x,i,j)\big)\,d\vec x\wedge di\wedge dj.
%\end{align}
%These
include $\mathcal{M}_\text{nlin}$ as well as Correlation Ratio (CR).  Correlation Ratio for image registration was proposed in \cite{roche1998crn}, but originates from analysis of variance and is based on the factorization of the variance into variance within classes and between class averages \cite{fisher1918}.  E.g., consider an image $I$, segmented into regions denoted by $R$ such that region $\Omega_j=\{x:R(x)=j\}$, and assuming uniform spatial distributions.  Then correlation ratio is defined as,
\begin{equation}
    1-\sum_j \frac{|\Omega_j|}{|\Omega|} \frac{\sigma_j^2}{ \sigma^2} = \sum_j \frac{|\Omega_j|}{|\Omega|} \frac{(\mu_j-\mu)^2}{ \sigma^2},
\end{equation}
where 
\begin{align}
    |\Omega|\mu &= \int_\Omega I(\vec x)\,d\vec x =\int_\Gamma ih(i)\,di,\\
    |\Omega|\sigma^2 &= \int_\Omega I(\vec x)^2-\mu^2\,d\vec x =  \int_\Gamma (i^2-\mu^2)h(i)\,di.
\end{align}
and
\begin{align}
    |\Omega_j|\mu_j &= \int_{\Omega_j} I(\vec x)\,d\vec x= \int_{\Gamma} ih_j(i)\,di,\\
    |\Omega_j|\sigma_j^2 &= \int_{\Omega_j} I(\vec x)^2-\mu_j^2\,d\vec x=  \int_\Gamma (i^2-\mu^2)h_j(i)\,dv.
\end{align}
for the corresponding local histograms $h_j$ of intensity values in $I$ of segment $\Omega_j$.

In this paper we will consider Normalized Mutual Information (NMI) \cite{studholme1999oie}, which has proven to be very powerful for registration of medical images in general.

%  NMI is defined as,
% % 
% \begin{align}
%   \label{eq:NMI}
% %  \mathcal{M}_\text{MI}  &=\mathcal{H}_I +\mathcal{H}_R-\mathcal{H}_{I,R},\\
%   \mathcal{M}_\text{NMI}  &=\frac{\mathcal{H}_I +\mathcal{H}_R}{\mathcal{H}_{I,R}},
% \end{align}
% % 
% where the entropies $\mathcal{H}$ are given as,
% % 
% \begin{align}
%   \mathcal{H}_I &=- \int_\Gamma p_I(i) \ln p_I(i)\,di\\
%   \mathcal{H}_R &=-\int_\Gamma p_R(i) \ln p_R(i)\,di\\
%   \mathcal{H}_{I,R} &=-\int_{\Gamma^2} p_{I,R}(i,j)\ln p_{I,R}(i,j)\,di\wedge dj
% \end{align}
% % 
% using the natural logarithm for convenience, and the intensity and joint intensity distributions, $p(\cdot):\Gamma\rightarrow\Re_+$ and $p(\cdot,\cdot): \Gamma^2\rightarrow\Re_+$, are estimated by the histogram and joint histogram of the intensity values, and the estimation algorithm will be discussed in the following section.  It is possible and maybe even natural to extend the similarity measure to include any combination of similarity measures, but in this article, we will consider NMI to be of type $\mathcal{M}_\text{nlin}$, with the loss-function is $F=(p_I(i) \ln p_I(i)+ p_R(i) \ln p_R(i))/\int_{\Gamma^2} p_{I,R}(m,n)\ln p_{I,R}(m,n)\,dm\wedge dn$.  Explicit examples of all the above types of  similarity measures are given in Appendix~\ref{sec:dissimilarityMeasures}.

\section{Density estimation}
\label{sec:locallyOrderlessImages}
%To estimate the entropies, we need to estimate joint and marginal histograms.
A common algorithm for estimating the histogram of an image is counting:  Given an image $I(\vec x)=i$, a set of bin-widths and sample points, $\Delta i_n>0$ and $i_n$, $m>n\Rightarrow i_m>i_n$, and an indicator function,
\begin{equation}
  P_n(i) =
  \begin{cases}
  1,&\text{ if $i_n\leq i<i_n+\Delta i_n$}\\
  0,&\text{otherwise.}
  \end{cases}
\end{equation}
Then the histogram may be found as,
\begin{equation}
 h(n)=\int_{\Omega}P_n(I(\vec x))\,d\vec x,
\end{equation}
or as a sum using a suitable discretization of $\Omega$.  The bin-widths act as scale parameters in the sense that when increasing $\Delta i_n$, then the histogram will contain less detail about the intensity distribution.  This can be stated precisely: Select a discrete set of sample points and bin-widths such that $\Delta i_n=i_{n+1}-i_n$, and consider 2 neighboring histogram values, $h(n)$ and $h(n+1)$.  In this case, the sum, $h'(n)=h(n)+h(n+1)$, is equivalent to evaluate the integral with a modified indicator function
\begin{equation}
  P_n'(i)=
  \begin{cases}
  1,&\text{ if $i_n\leq i<i_{n+1}+\Delta i_{n+1}=i_n+\Delta i_n'$}\\
  0,&\text{otherwise,}
  \end{cases}
\end{equation}
where $\Delta i_n'=\Delta i_n+\Delta i_{n+1}$.  By induction it becomes clear that filtering $h(n)$ with a Boxcar function (0-order b-spline) of height 1 and width $m$ is equivalent to increasing the extend of the indicator function as $\Delta i_n'=\sum_{k=0}^{m-1} \Delta i_{n+k}$.  Thus, increasing $\Delta i$ is equivalent to smoothing the histogram with a Boxcar function.

In general, the interesting scales of $i$ are not given by the data, and therefore the only option is to study all scales, that is, all discretizations of intensity.  Along with the scale-space on the spatial parameter $\vec x$, this leads to a scale-space theory for space and intensity known as Imprecision Space \cite{griffin97}.  In the general case, histograms are local. Since the scale of the region of interest is not generally given, then we are also required to study all scales.  This scale we denote the integration scale, and the combined construction is called Locally Orderless Images (LOI) \cite{koenderink1999structure}.

\subsection{Estimating local histograms}
According to Locally Orderless Images, a local histogram is obtained as follows: First a (possibly deformed) image $I$ is smoothed with the kernel $K$, a soft isophote $i$ is extracted using kernel $P$, and finally the isophote mass is calculated in a neighborhood of a point $\vec x$ with kernel $W$.   Formally,
\begin{align}
  \label{eq:localHistogram}
  h_I(i,\vec x,\mat\Phi,\alpha,\beta,\sigma)
  &= P(I(\vec x,\mat\Phi,\sigma)-i,\beta)*W(\vec x, \alpha),\\
  I(\vec \psi,\mat\Phi,\sigma)
  &= I(\vec x)*K(\vec x,\sigma),
  \label{eq:spatialSmoothing}
\end{align}
where $P:\Re\times\Re_+\rightarrow [0,1]$ is an intensity measurement of scale $\beta$ and is often called the Parzen Window (PW), $K:\Re^N\times\Re_+\rightarrow\Re_+$ is a spatial measurement kernel of scale $\sigma$, $W:\Re^N\times\Re_+\rightarrow\Re_+$ is an integration window of integration scale $\alpha$, $\cdot *\cdot$ is the convolution operator taken w.r.t.\ the variable $\vec x$, and $\mat\Phi\in\Re^M$ denotes the parameters for the transformation.  The histogram $h_R$ is defined similarly independently of $\mat\Phi$.  In \cite{koenderink1999structure} it is proposed to use $P(i,\beta)=\exp{-i^2/(2\beta^2)}$, $K(\vec x,\sigma)=\exp{-\vec x^T\vec x/(2\sigma^2)}/(2\pi\sigma^2)^{N/2}$, and $W(\vec x,\alpha)=\exp{-\vec x^T\vec x/(2\alpha^2)}/(2\pi\alpha^2)^{N/2}$, which implies the structure of the heat diffusion in all 3 scale parameters and is considered the simplest structure imposable for studying data by all scales.  In typical registration scenarios, such as registering CT and MR images, intensity and spatial scale are of quite different nature. The spatial scales can often be related to a common frame of references, but this is not as easy for intensity scales, which indicates that information measures may be preferable for multimodal registration.

\begin{figure}
  \centering  \subfigure[][]{\label{fig:image:orig}\includegraphics[width=0.48\linewidth]{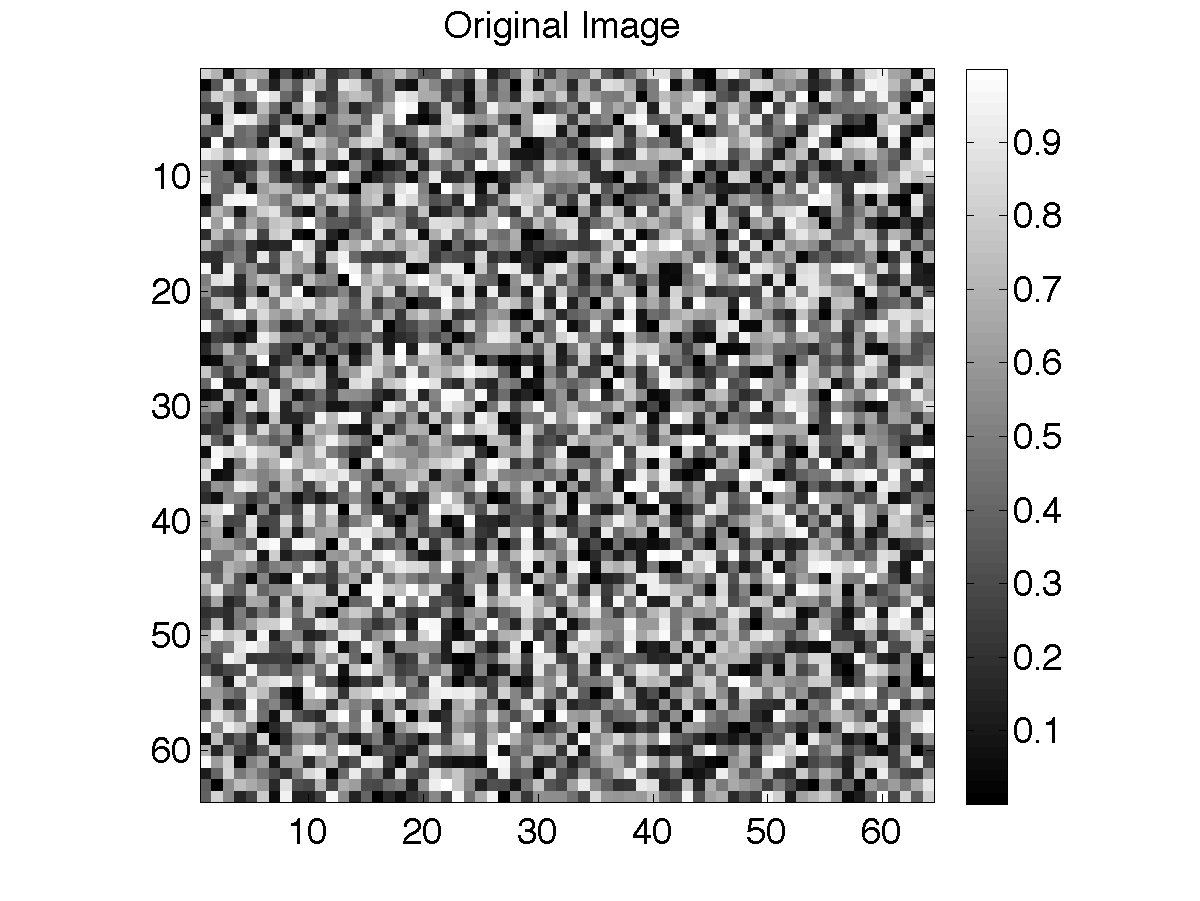}}
\subfigure[][]{\label{fig:image:origHist}\includegraphics[width=0.48\linewidth]{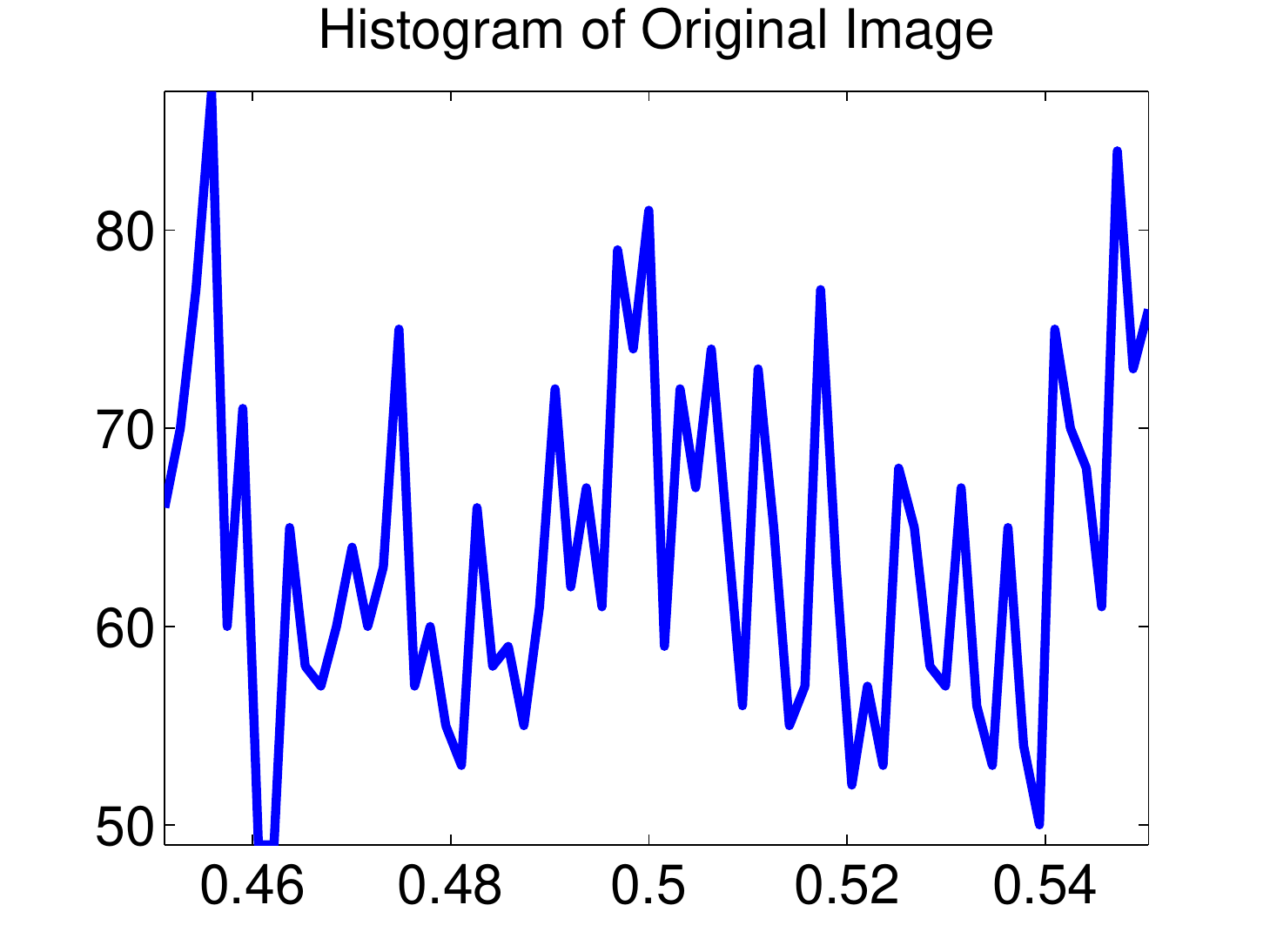}}
  \caption{(a): A random image and (b): its histograms.}
  \label{fig:original}
\end{figure}

To give intuition we will discuss the calculation of the local histograms in a step by step manner including the meaning of the various scale parameters.  Consider a  random image and its histogram as calculated by the Matlab \texttt{hist} function, shown in \figurename~\ref{fig:original}.
In terms of local histogram parameters, this corresponds to: $\alpha=\infty$, $\sigma=0$, and $\beta=1/\sqrt{12}$, the standard deviation of a Boxcar function of width 1.

\begin{figure}
  \centering
  \subfigure[][]{\label{fig:image:smoothed}\includegraphics[width=0.48\linewidth]{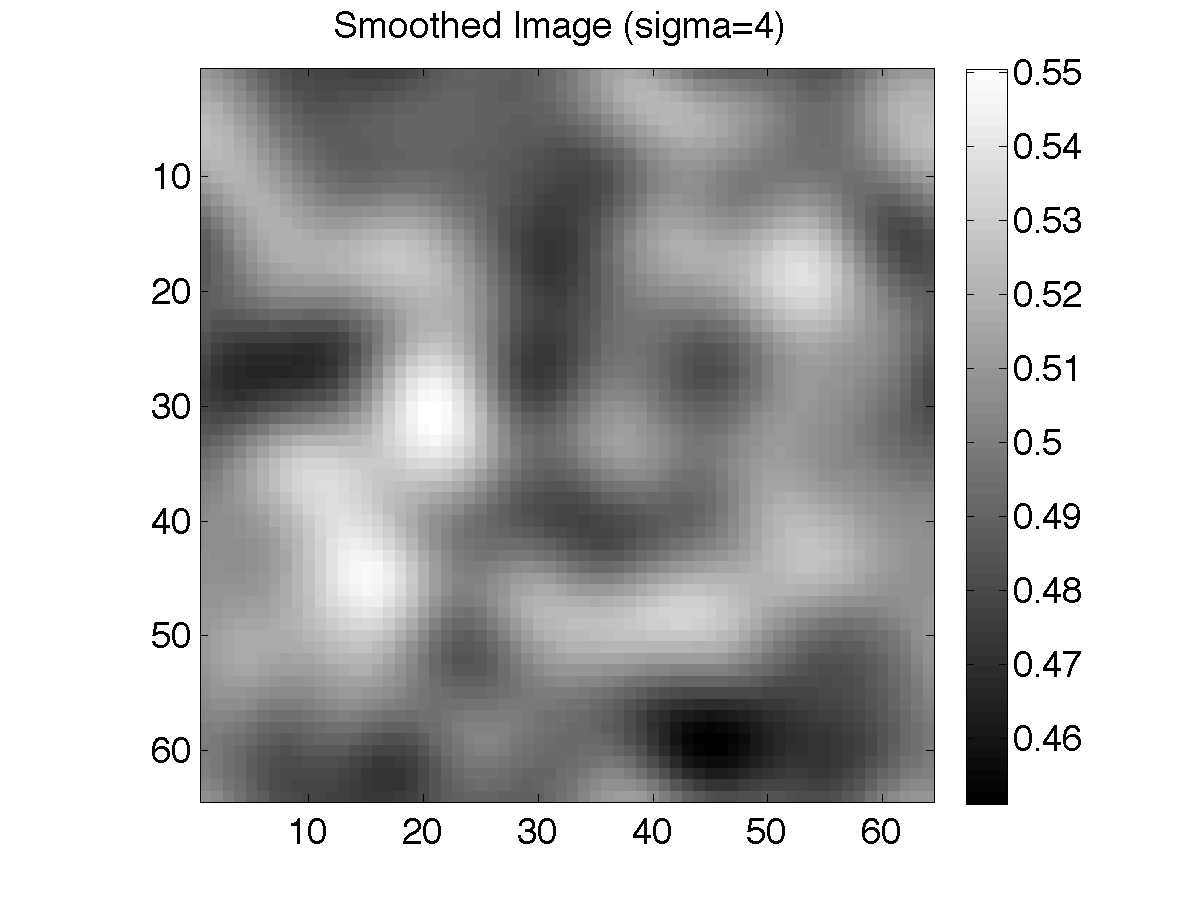}}
  \subfigure[][]{\label{fig:image:smoothedHist}\includegraphics[width=0.48\linewidth]{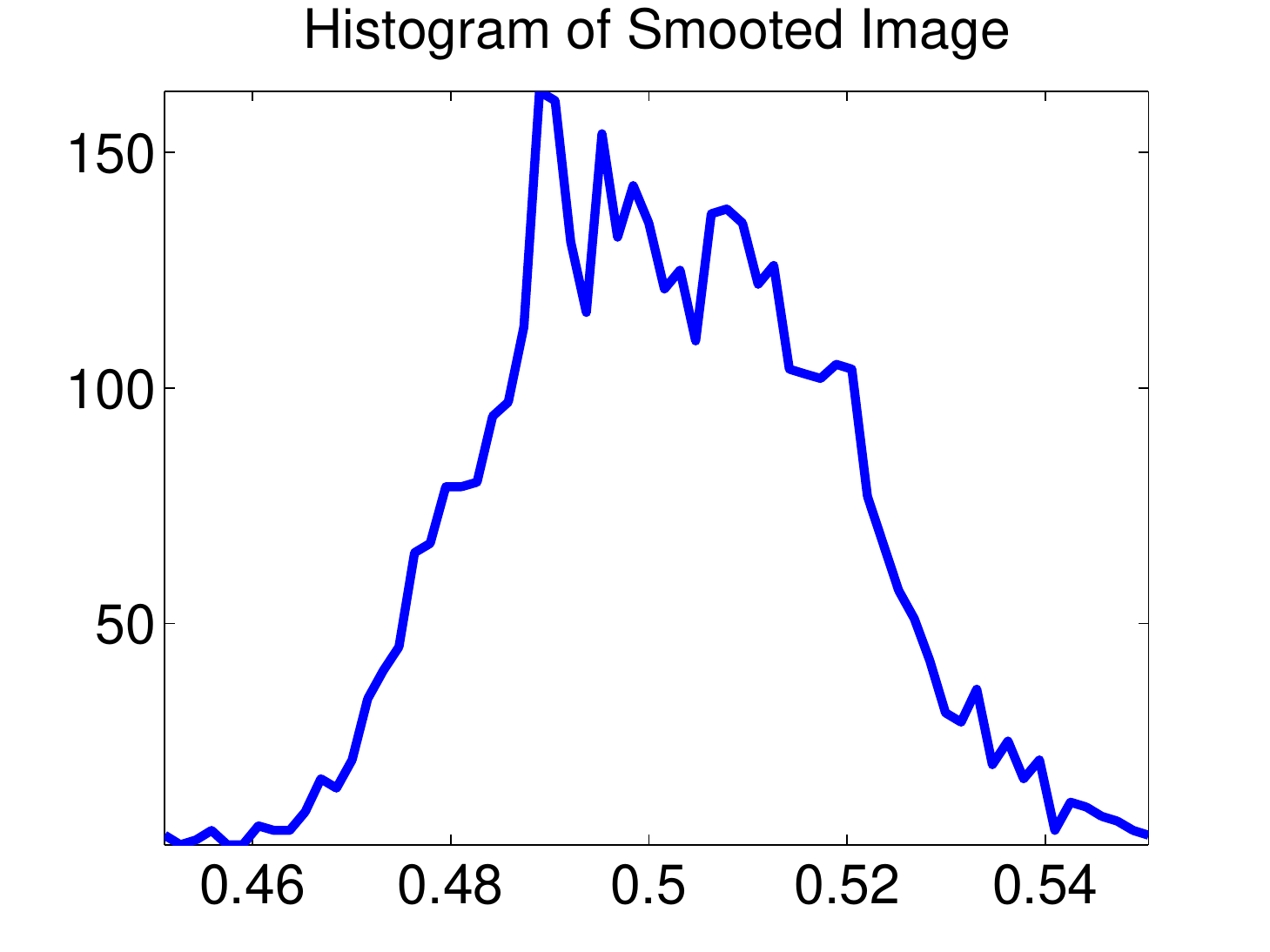}}
  \caption{(a): The image in \figurename~\ref{fig:image:orig} smoothed with $\sigma=4$ and (b): its histogram.}
  \label{fig:smoothedImage}
\end{figure}

\paragraph{Image smoothing with $K$}
First step in calculating a local histogram is to smooth the image with kernel $K$.  The kernel $K$ controls the image scale, $\sigma$. This is illustrated in \figurename~\ref{fig:smoothedImage} and corresponds to $\alpha=\infty$, $\sigma>0$, and $\beta=\Delta i/\sqrt{12}$, where $\Delta i$ is the original intensity scale. Since smoothing an image implies a monotonic contraction of image intensity around the mean value, we expect that the histogram is likewise contracted, when increasing $\sigma$.  This is confirmed by the experiment illustrated in \figurename~\ref{fig:image:smoothedHist}.

\begin{figure*}
  \centering
  \subfigure[][]{\label{fig:isophote:smoothed}\includegraphics[width=0.32\linewidth]{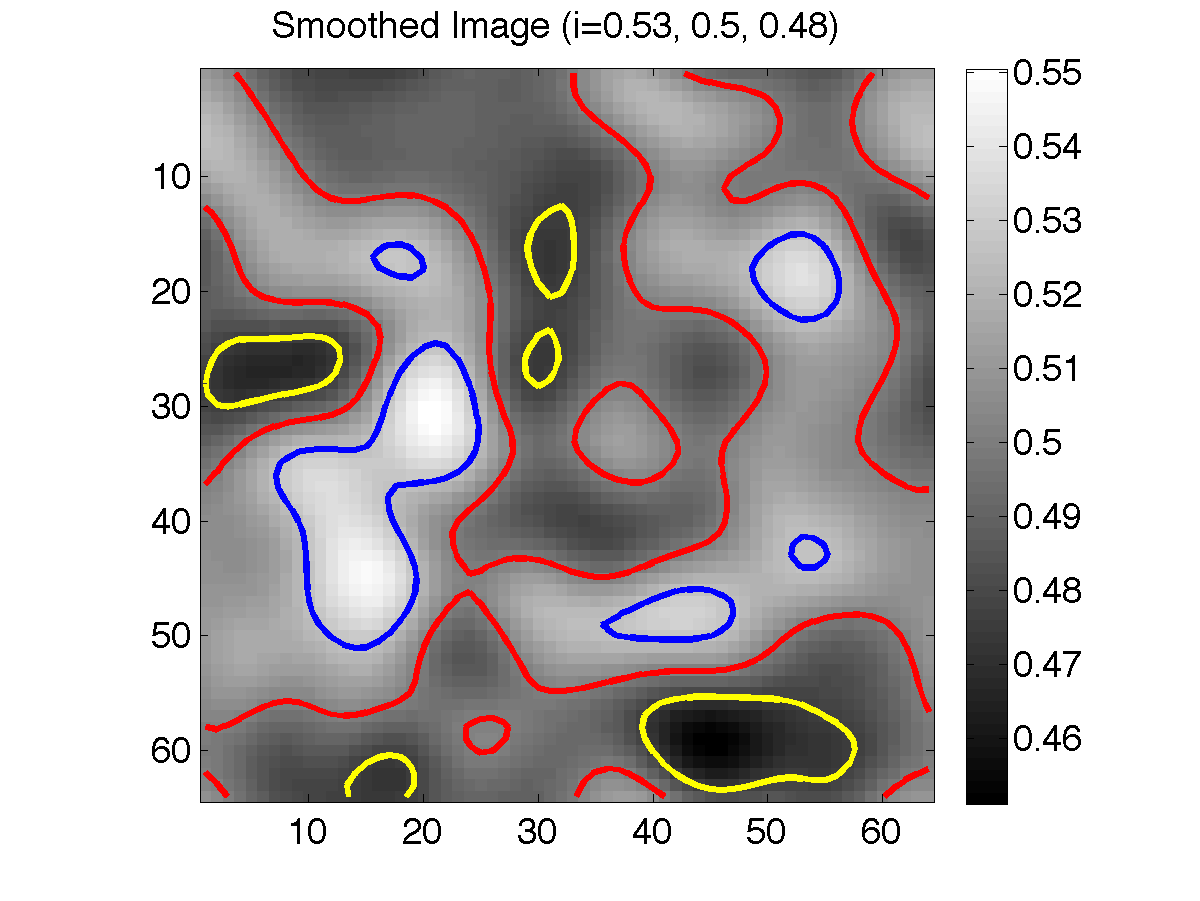}}
  \subfigure[][]{\label{fig:isophote:1}\includegraphics[width=0.32\linewidth]{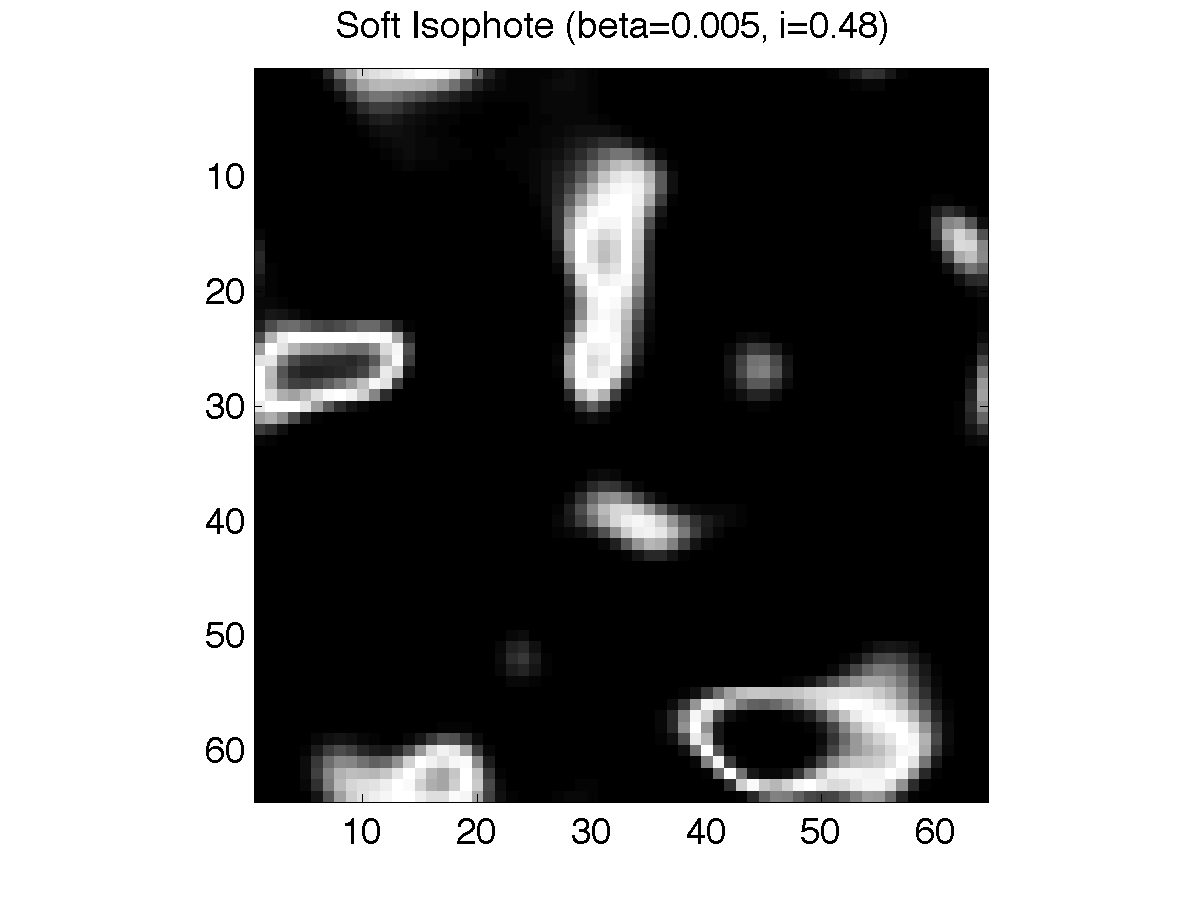}} 
  \subfigure[][]{\label{fig:isophote:2}\includegraphics[width=0.32\linewidth]{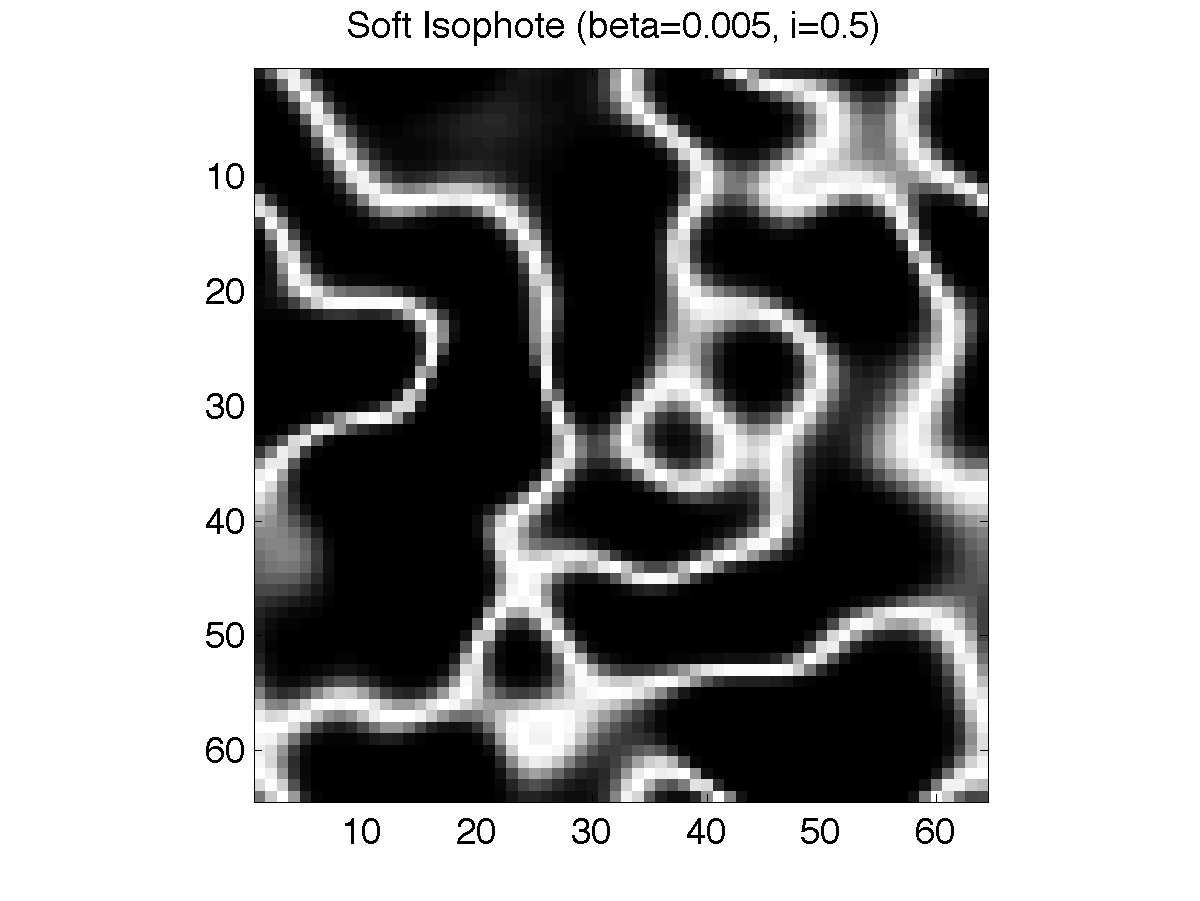}}
  \caption{Measuring isophotes in \figurename~\ref{fig:smoothedImage}. (a): 3 isophote lines as produced by Matlab's \texttt{contour} function, (b) and (c): the yellow and red isophotes as extracted with a kernel P using $i=0.48$ and $i=0.50$ and in both instances $\beta=0.005$.}
  \label{fig:softIsophote}
\end{figure*}

\begin{figure}
  \centering
  \subfigure[][]{\label{fig:softHistogram}\includegraphics[width=0.48\linewidth]{smoothedHist}}
  \subfigure[][]{\label{fig:histogramSmoothed}\includegraphics[width=0.48\linewidth]{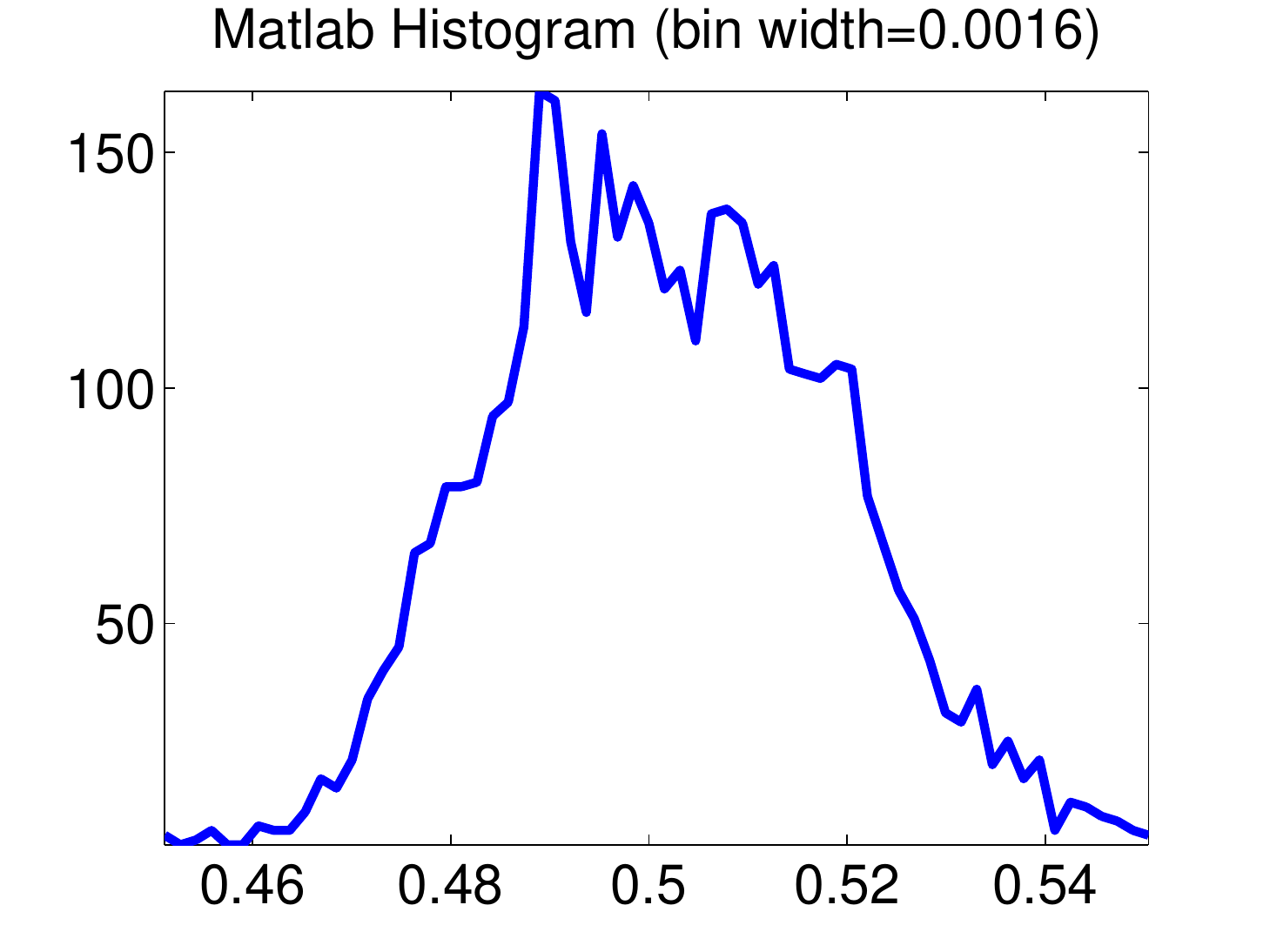}} 
  \caption{Measuring histograms of \figurename~\ref{fig:smoothedImage}. (a): A local histogram using $\alpha=\infty$ and $\beta=0.005$, and (b): a histogram using Matlab's \texttt{hist} function.}
  \label{fig:histogramsCompared}
\end{figure}

\paragraph{The Parzen window, $P$}
Second step is to calculate the soft isophote $i$ with kernel $P$:  The kernel $P$ controls intensity scale, $\beta$.  This is illustrated in \figurename~\ref{fig:softIsophote} and corresponds to $\alpha=0$, $\sigma>0$, and $\beta>0$. 
\figurename~\ref{fig:isophote:1} and~\ref{fig:isophote:2} show the spread of 2 fixed isophotes for the chosen $P$.  For a fixed position $\vec x$ the image contains the value of the local histogram at $\vec x$.  Hence, the stack of images for all isophotes gives all the local histograms.  The spread of a soft isophote depends on the image geometry at $I(\vec x,\sigma)=i$:  The spread will be large, where the gradient magnitude is small, and small, where the gradient magnitude is large.  In general the width $\beta$ acts as the bin-width in the histogram, and varying $\beta$ corresponds to varying the degree of smoothing of the histogram, which is illustrated in \figurename~\ref{fig:histogramsCompared}.

\begin{figure}
  \centering
  \subfigure[][]{\label{fig:loi:img1}\includegraphics[width=0.4\linewidth]{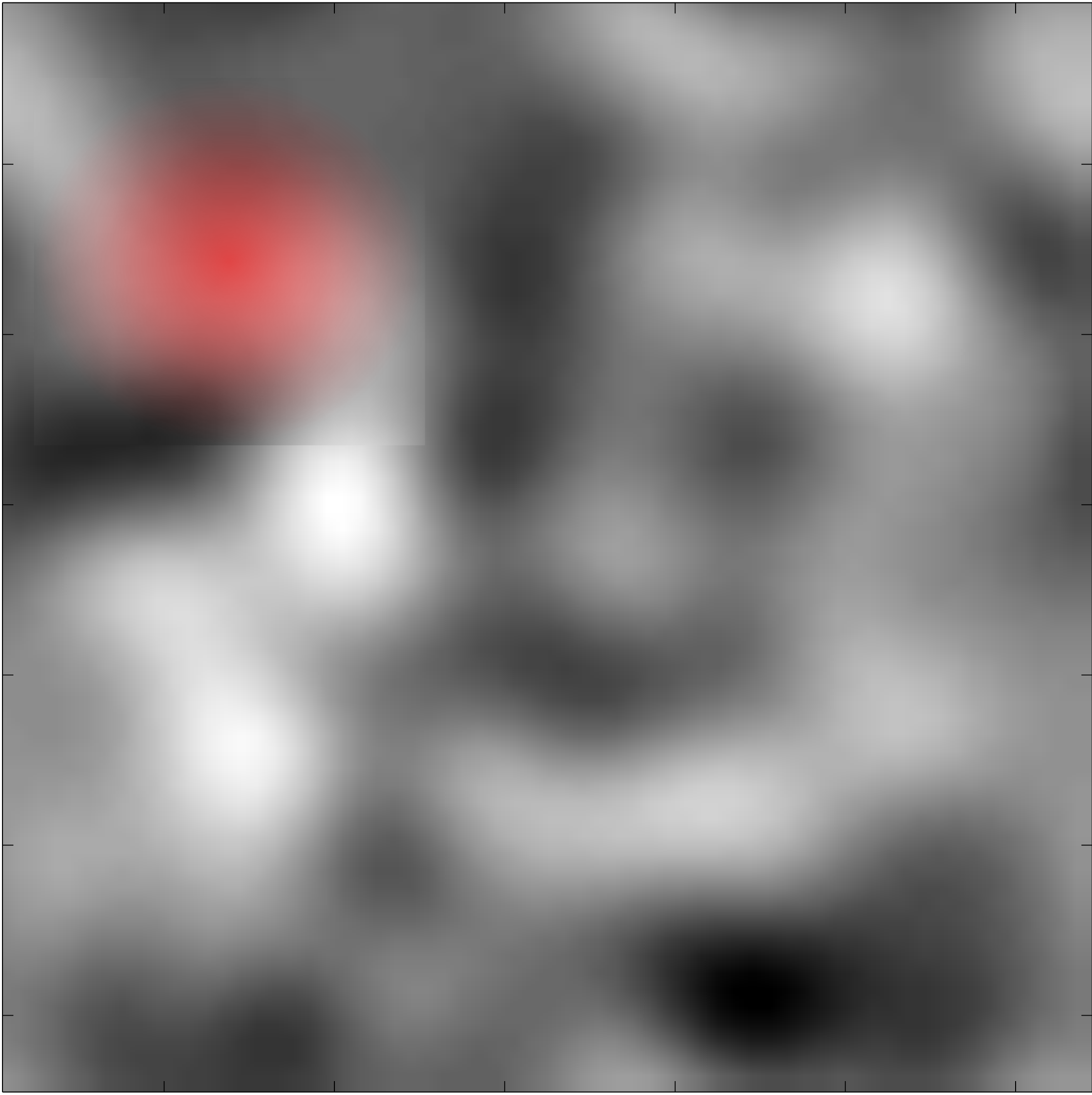}}
  \subfigure[][]{\label{fig:loi:hist1}\includegraphics[width=0.55\linewidth]{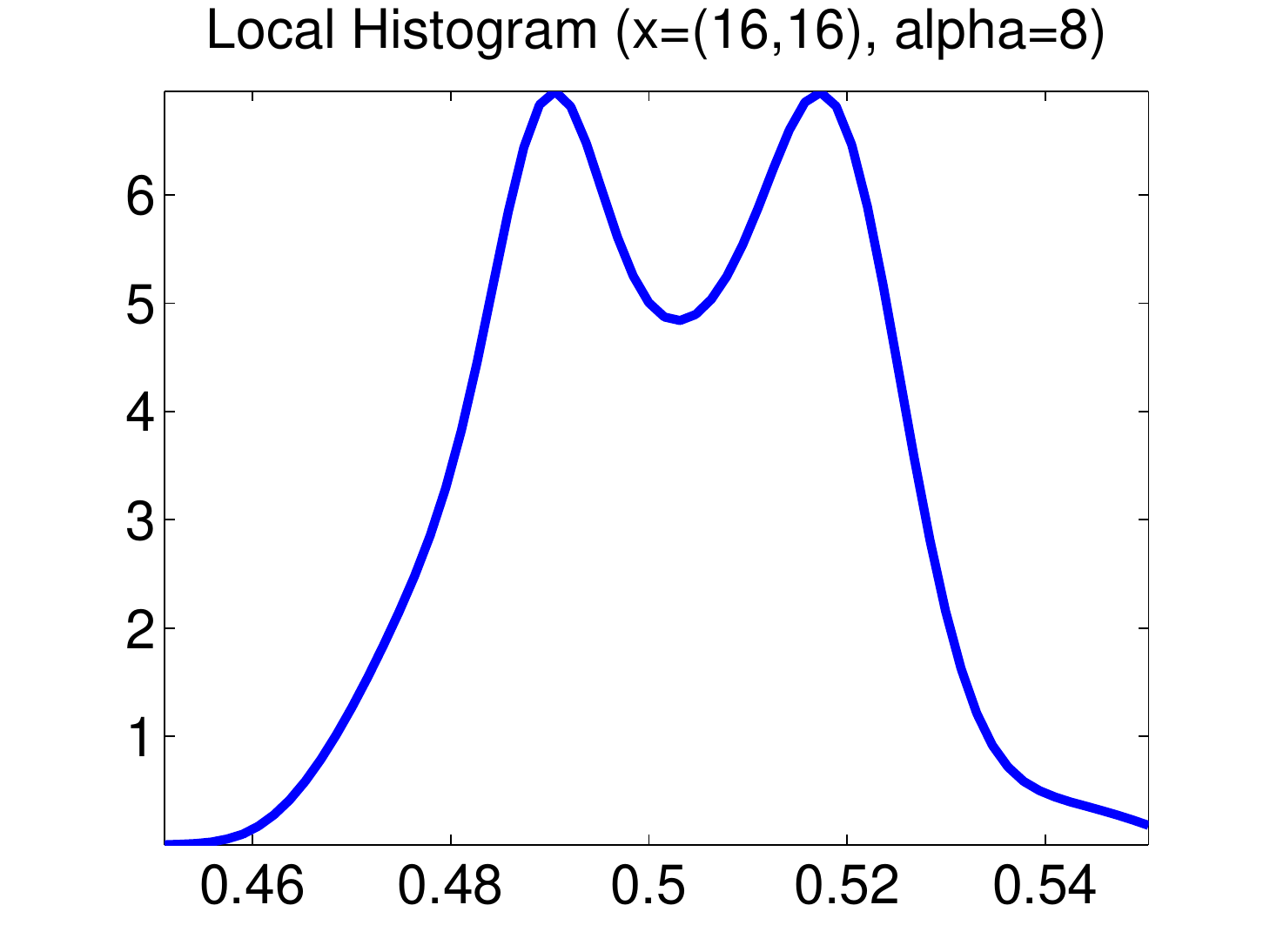}}
  \subfigure[][]{\label{fig:loi:img2}\includegraphics[width=0.4\linewidth]{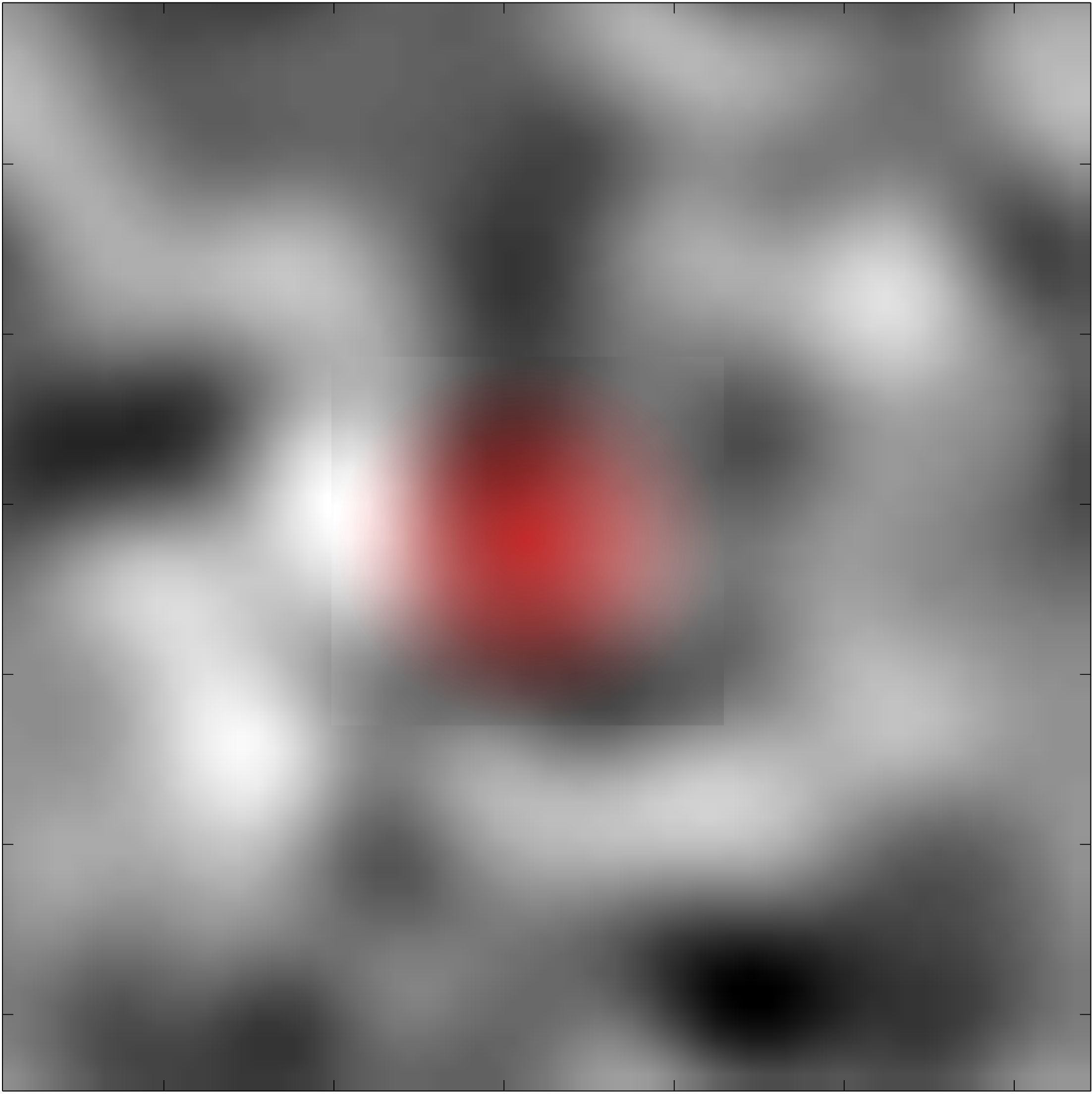}}
  \subfigure[][]{\label{fig:loi:hist2}\includegraphics[width=0.55\linewidth]{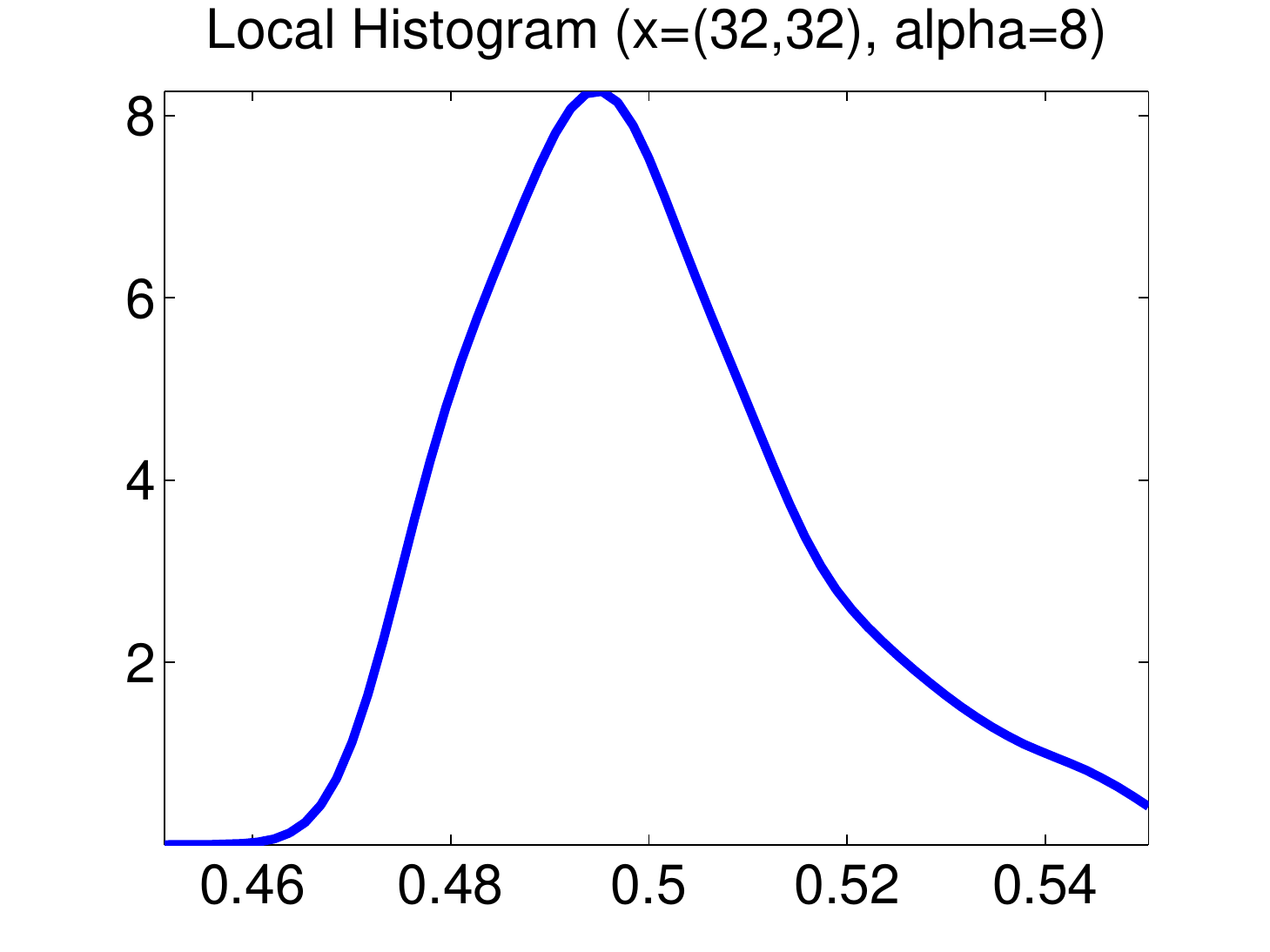}}
  \subfigure[][]{\label{fig:loi:img3}\includegraphics[width=0.4\linewidth]{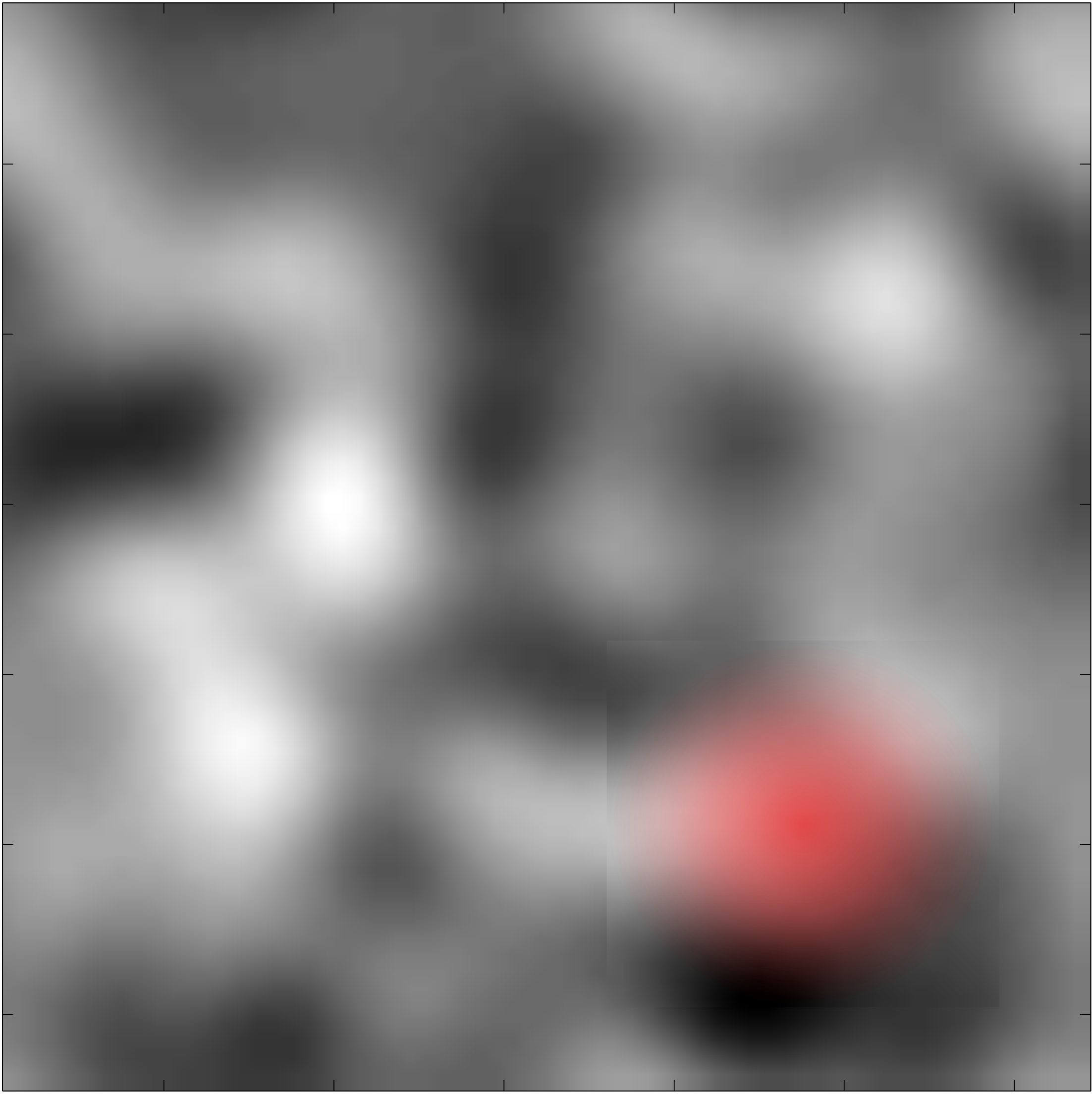}}
  \subfigure[][]{\label{fig:loi:hist3}\includegraphics[width=0.55\linewidth]{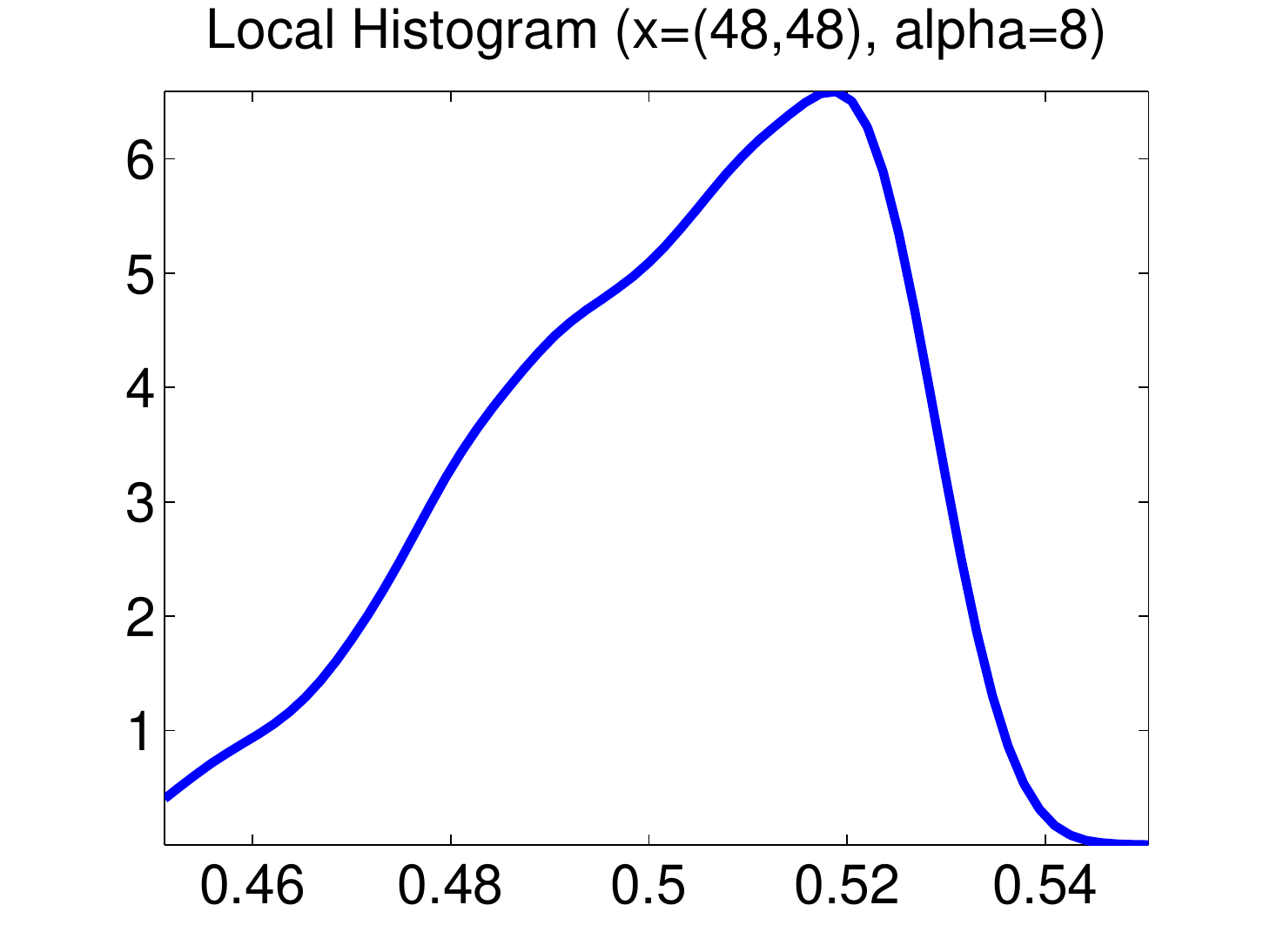}}
  \caption{Examples of local histograms generated by Locally orderless images in neighbourhoods as indicated by the red overlays.}
  \label{fig:loi}
\end{figure}

%\paragraph{Specifying region of interest with $W$}
\paragraph{Locality, $W$}
Last step is to calculate the local isophote area near $\vec x$ with kernel $W$: The kernel $W$ controls the locality of the local histogram, $\alpha$, illustrated in \figurename~\ref{fig:loi}.
Note that the histograms change quite significantly with the position of the kernel $W$.  

\subsection{Some relations between scales}
\label{sec:notesOnScales}
In general, varying $\beta$ and varying $\sigma$ yields different results, since the width of a soft isophote in a point is proportional to the gradient in the point, while the extend of local average is irrespective of the gradient in the point.  In addition, near the symmetry set \cite{bruce.giblin92}, the soft isophote will have a ridge like behavior.

The relation between $\alpha$ and $\sigma$ may be stated in terms of the histogram raw moments and central moments.  The raw moment and central moments of order $n\geq 0$ of the histogram $h$ at position $\vec x$ are given as,
\begin{align}
 \mu_n'&=\int_{-\infty}^\infty i^n h(i,\vec x)/k(\vec x)\,di\\
 \mu_n&=\int_{-\infty}^\infty (i-\mu(\vec x))^n h(i,\vec x)/k(\vec x)\,di
\end{align}
where $k(\vec x)=\int_{-\infty}^\infty h(i,\vec x)\,di$ and $\mu(\vec x)=\mu_1'$ is the mean value.  In the following, we will evaluate these moments.  
\begin{itemize}
\item \emph{Normalization constant $k$:} Convolution is linear, thus $k(\vec x)=\big(\int_{-\infty}^\infty P(L(\vec x)-i)\,di\big) *W(\vec x)$.  Since the image value $L(\vec x)$ under the integral acts as a translation of the Parzen window, and since the integral is over the entire domain, we conclude that the integral is independent on $\vec x$, and thus
  \begin{equation}
    k=\int_{-\infty}^\infty P(i)\,di
  \end{equation}
  independent on $\vec x$.  In case of a Gaussian Parzen window of variance $\beta^2$, then $k_\text{Gauss}=\beta\sqrt{2\pi}$. 

\item\emph{Mean value $\mu$:} If the Parzen window, $P$, has zero mean value, then the mean value of $P(L(\vec x)-i)$ is $L(\vec x)$, and thus
  \begin{align}
    \mu&=\int_{-\infty}^\infty i h(i,\vec x)/k\,di=k^{-1}L(\vec x)*W(\vec x)\\
    &=k^{-1}I(\vec x)*K(\vec x)*W(\vec x)\\
    &=k^{-1}I(\vec x)*W'(\vec x),
  \end{align}
  where $W'(\vec x)=K(\vec x)*W(\vec x)$. In case of Gaussian $K$ of variance $\sigma^2$ and $W$ of variance $\alpha^2$, then $W'(\vec x)$ is Gaussian of variance $\sigma^2+\alpha^2$. 

\item\emph{Raw moments $\mu_n'$:}
The raw moments of $h$ may be constructed from the moments of $P/k$.  Writing the raw and central moments of $P/k$ as $\eta_n'$ and $\eta_n$, we have that
\begin{align}
  \eta_n'&=\sum_{j=0}^n (-1)^j\binom{n}{j}\eta_j(\eta_1')^{n-j}\\
  &=\sum_{j=0}^n (-1)^j\binom{n}{j}\eta_jL(\vec x)^{n-j}.
\end{align}
The $n$'th central moments for a Gaussian of variance $\beta^2$ is $(n-1)!!\beta^n$ for even $n$ and 0 otherwise, where $(n-1)!!$ is the double factorial function.  Thus, the $n$'th raw moments of $h$ is $\mu_n'=\eta_n'*W(\vec x)$, which is linear combination of terms $L(\vec x)^{n-j}*W(\vec x),\; j=0\ldots n$, and the relation between $\sigma$ and $\alpha$ is non-linear for most cases of $n$ and $j$.  For Gaussian $K$ and $W$ this is a pseudo-linear scale-space~\cite{florack99}.  Examples are given in Table~\ref{tab:moments}.

\item\emph{Central moments $\mu_n$:}
The central moments of $h$ may be constructed from its raw moments, since
\begin{equation}
  \mu_n=\sum_{j=0}^n\binom{n}{j}(-1)^{n-j}\mu_j'\mu^{n-j}.
\end{equation}
Examples are given in Table~\ref{tab:moments}. 
\end{itemize}
\begin{table*}
\centering
\begin{tabular}{l|lll}
$n$ & $\eta_n$ & $\mu_n'$&$\mu_n$
\\\hline 0
&1
&1
&1
\\1
&0
&$L(\vec x)*W(\vec x)$
&0
\\2
&$\beta^2$
&$\big(L(\vec x)^2+\beta^2\big)*W(\vec x)$
&$-(\mu_1')^2+\mu_2'$
\\3
&
0
&$\big(L(\vec x)^3+3\beta^2L(\vec x)\big)*W(\vec x)$
&$2(\mu_1')^3-3\mu_1'\mu_2'+\mu_3'$
\\4
&
$3\beta^4$
&$\big(L(\vec x)^4+6\beta^2L(\vec x)^2+3\beta^4\big)*W(\vec x)$
&$-3(\mu_1')^4+6(\mu_1')^2\mu_2'-4\mu_1'\mu_3'+\mu_4'$
\\5
&
0
&$\big(L(\vec x)^5+10\beta^2L(\vec x)^3+15\beta^4L(\vec x)\big)*W(\vec x)$
&$4(\mu_1')^5-10(\mu_1')^3\mu_2'+10(\mu_1')^2\mu_3'-5\mu_1'\mu_4'+\mu_5'$\\
\end{tabular}
\caption{Examples of raw and central moments $\mu_n'$ and $\mu_n$ of order $n$, when the Parzen window has central moments $\eta_j,\, j=0\ldots n$, as does a Gaussian of zero mean and variance $\beta^2$.}
\label{tab:moments}
\end{table*}

Some intuition may be obtained by considering an image, which in the neighborhood of the point $\vec x_0$ is linear with gradient $\nabla I(x)$. The image in the neighborhood of $\vec x_0$ is then given as
\begin{equation}
  I(\vec x)\simeq I(\vec x_0)+(\vec x - \vec x_0)\cdot\nabla I(\vec x_0),
\end{equation}
and the isophotes near $I(\vec x_0)$ are all lines perpendicular to the gradient.  The image in the neighborhood around $\vec x_0$ is invariant w.r.t., smoothing with symmetric and normalized kernels, hence $\sigma$ has no influence on the local histograms for small values of $\sigma$. However, the interplay between $\beta$ and $\alpha$ is nontrivial:  The soft isophotes are constant perpendicular to the gradient.  Hence, we may consider this a one dimensional problem along the axis of the gradient, say $x$, and $I(x)\simeq a x + b$, where $a=|\nabla I(\vec x_0)|$, $ax=(\vec x - \vec x_0)\cdot\nabla I(\vec x_0)$, and $b=I(\vec x_0)$. The soft isophote $b$ using Gaussian $P$ is $P(ax,\beta)=P(x,\beta/a)$, convolution with a Gaussian integration kernel $W(x,\alpha)$ yields another Gaussian
\begin{equation}
  P(ax,\beta)*W(x,\alpha)=P(x,\sqrt{\beta^2/a^2+\alpha^2}),
\end{equation}
due to the semi-group properties of Gaussian convolution.  For non-linear images the interplay between $\sigma$, $\beta$, and $\alpha$ is more complicated.

\subsection{Estimating local densities}
The local density distributions are obtained by normalizing to unity,
\begin{align}
  p_I(i|\vec x,\mat\Phi,\alpha,\beta,\sigma) &\simeq \frac{h_I(i,\vec x,\mat\Phi,\alpha,\beta,\sigma)}{\int_\Gamma h_I(j,\vec x,\mat\Phi,\alpha,\beta,\sigma) dj},\\
  p_I(i|\mat\Phi,\alpha,\beta,\sigma) &= \frac{1}{|\Omega |}\int_\Omega p_I(i|\vec x,\mat\Phi,\alpha,\beta,\sigma)\,d\vec x,
\end{align}
where we have assumed (conditional) independence and uniformity such that $p_I(i,\vec x|\mat\Phi,\alpha,\beta,\sigma)=p_I(i|\vec x,\mat\Phi,\alpha,\beta,\sigma)/|\Omega|$.  The density $p_R$ is defined in a similar manner.  As \cite{hermosillo2002variational}, we extend the concept to the joint distributions as follows,
\begin{align}
  &h_{I,R}(i,j,\vec x,\mat\Phi,\alpha,\beta,\sigma) =\nonumber\\
&\left(P(I(\vec x,\mat\Phi,\sigma)-i,\beta)P(J(\vec x,\sigma)-j,\beta)\right)*W(\vec x, \alpha),\\
  &p_{I,R}(i,j|\vec x,\mat\Phi,\alpha,\beta,\sigma) \simeq \frac{h_{I,R}(i,j,\mat\Phi,\vec x,\alpha,\beta,\sigma)}{\int_{\Gamma^2} h_{I,R}(k,l,\vec x,\alpha,\beta,\sigma)\,dk\wedge dl},\\
  &p_{I,R}(i,j|\mat\Phi,\alpha,\beta,\sigma) = \frac{1}{|\Omega |}\int_\Omega p_{I,R}(i,j|\mat\Phi,\vec x,\alpha,\beta,\sigma)\,d\vec x,
\end{align}
where we also have assumed (conditional) independence and uniformity such that $p_{I,R}(i,j,\vec x|\mat\Phi,\alpha,\beta,\sigma)=p_{I,R}(i,j|\vec x,\mat\Phi,\alpha,\beta,\sigma)/|\Omega|$.

\section{Theoretical comparison of PWs and GPV density estimation}
\label{sec:pwgpvCompare}
Locally orderless image is the cornerstone in understanding the difference between the PW and GPV density estimators.% to be discussed in the following.

\subsection{The PW is a special case of Locally orderless images}
\label{sec:parzen}
The PW, originally proposed along with MI in \cite{wells1996mmv}, is a special case of Locally orderless images, often used in the literature. Consider~\eqref{eq:localHistogram}--\eqref{eq:spatialSmoothing} and let  $\alpha\rightarrow\infty$.  In that case, the window $h_I$ simplifies as,
\begin{align}
  \label{eq:hist_cont}
  h_I(i,\vec x,\mat\Phi,\alpha,\beta,\sigma)&\rightarrow \text{const.}\,\int_\Omega P(I(\vec \psi,\mat\Phi,\sigma)-i,\beta)\,d\vec\psi,\\
  p_I(i|\mat\Phi,\alpha,\beta,\sigma)&\rightarrow \frac{\int_\Omega P(I(\vec \psi,\mat\Phi,\sigma)-i,\beta)\,d\vec\psi}{\int_\Gamma\int_\Omega P(I(\vec \psi,\mat\Phi,\sigma)-j,\beta)\,d\vec\psi\wedge dj}.
\end{align}
Choosing
\begin{align}
  \label{eq:GaussParzenWindow} 
  P(i,\beta)&=\exp{-i^2/(2\beta^2)},
\end{align}
we find that
\begin{align}
  &\int_\Gamma\int_\Omega P(I(\vec \psi,\mat\Phi,\sigma)-j,\beta)\,d\vec\psi\wedge dj
  = |\Omega| \sqrt{2\pi\beta^2},
\end{align}
and
\begin{equation}
  \label{eq:single_bin_cont_conv}
  p_I(i|\mat\Phi,\alpha,\beta,\sigma)\rightarrow \frac{1}{|\Omega| \sqrt{2\pi\beta^2}}\int_\Omega \exp{-(I(\vec x,\mat\Phi,\sigma)-i)^2/(2\beta^2)}\,d\vec x.
\end{equation}
Likewise, we have
\begin{multline}
  \label{eq:joint_bin_cont_conv}
  p_{I,R}(i,j|\mat\Phi,\alpha,\beta,\sigma) \rightarrow\\ 
   \frac{\int_\Omega \exp{-(I(\vec x,\mat\Phi,\sigma)-i)^2+(R(\vec x,\sigma)-j)^2/(2\beta^2)}\,d\vec x}{|\Omega| 2\pi\beta^2}.
\end{multline}
This is precisely the PW method using a Gaussian kernel with infinite support \cite{wells1996mmv}. Similar results are obtained for any integrable Parzen window, $P(i,\beta)$. The PW can be interpreted as a globally orderless image, as $\mat W$ defining the locality extends globally. 

As a side note, since both~\eqref{eq:single_bin_cont_conv} and~\eqref{eq:joint_bin_cont_conv} obey the diffusion equation w.r.t.\ $\beta^2/2$, we may use Green's theorem and write,
\begin{align}
  p_I(i|\sqrt{\beta_0^2+\beta^2}) &= p_I(i|\beta_0)*G(i,\beta),\\
  p_{I,R}(i,j|\sqrt{\beta_0^2+\beta^2}) &= p_{I,R}(i,j|\beta_0)*G([i,j]^T,\beta),
\end{align}
for fast computation of a range of PW sizes. Further, $\alpha \rightarrow 0$ in MI for 2D images reduces to $-\log(\angle(\nabla I,\nabla R))$ \cite{griffin2006}, i.e., the angle between the gradients of the images at $x$, which is similar to Normalized gradient fields proposed in \cite{haber2007intensity}.

\subsection{GPV is an approximation of Locally orderless images}
Shortly after the introduction of PW, partial volume (PV) was introduced in \cite{maes1997mir} and extended to GPV in \cite{chen2003mutual}. Unlike PW, GPV estimates a global density as a sum of local densities and samples the intensity values directly from the image in the local neighborhood $W$. GPV may be derived from the joint histograms as follows.  First, calculate the joint histogram,
\begin{align}
  &h_{I,R}(i,j,\vec x,\alpha,\beta,\sigma) \nonumber\\
  &=\int_\Omega P(I(\vec \psi,\sigma)-i,\beta)P(J(\vec \psi,\sigma)-j,\beta)W(\vec x-\vec \psi, \alpha)\,d\vec\psi\label{eq:LOI}\\
  &\approx P(J(\vec x,\sigma)-j,\beta)\int_\Omega P(I(\vec \psi,\sigma)-i,\beta)W(\vec x-\vec \psi, \alpha)\,d\vec\psi \\\
  &=P(J(\vec x,\sigma)-j,\beta)\left[ P(I(\vec x,\sigma)-i,\beta)*W(\vec x,\alpha)\right]
  \label{eq:GPV}
\end{align}
Then set $P$ to a Boxcar function,
\begin{align}
  P(i,\beta)&= \begin{cases} 1 & \mathrm{if} \, -\frac{\beta}{2} \leq i < \frac{\beta}{2} \\ 0 & \mathrm{otherwise} \end{cases} 
\end{align}
where $\beta$ is chosen such that $I(\vec \psi,\mat\Phi,\sigma)$ is mapped into non-coinciding isophotes curves. The motivation for this is that all isophotes can be evaluated at $\vec{x}$ simultaneously and can be thought of as a 0-order b-spline PW. When integrating over the entire domain $\mat \Omega$ the GPV scheme is obtained. Thus GPV uses small local histograms integrated to form the globally orderless image as in the PW approach.  This introduces an asymmetry for $\alpha > 0$ in the joint densities making registration results inconsistent w.r.t.\ inversion.  This asymmetry has a direct influence on the marginal densities giving 3 different estimates of the marginal density: estimated from the histogram of a single image, or as the integral of either of the two joint histograms. I.e., ignoring the scale parameters, the histograms of, say, $J$ are given as,
\begin{equation}
h(j)=\int_\Omega P(J(\vec x)-j)\,dx,
\end{equation}
and the corresponding marginal in the GPV approximation is found either as,
\begin{align}
\tilde{h}(j)&=\int_\Omega \int_\Gamma P(J(\vec x)-j)[P(I(\vec x)-i)*W(\vec x)]\,di\,dx\\
&=\int_\Omega P(J(\vec x)-j)\int_\Gamma P(I(\vec x)-i)*W(\vec x)\,di\,dx,
\end{align}
or as
\begin{align}
h'(j)&=\int_\Omega \int_\Gamma P(I(\vec x)-i)[P(J(\vec x)-j)*W(\vec x)]\,di\,dx\\
&=\int_\Omega \int_\Gamma P(I(\vec x)-i)\,di\;P(J(\vec x)-j)*W(\vec x)\,dx.
\end{align}
The difference between these three estimates depends on the gradient of $I(\vec x)$, and due to the scale of $W$, the gradient will differ for the two estimates based on the joint histograms. The asymmetry in GPV causes $\mathcal M(A,B)\neq \mathcal M(B,A)$. In the limit of$\alpha \rightarrow 0$, and when using identical kernels and parameters as Parzen windows for $I$ and $J$, then GPV is symmetric, but unfortunately at the limit differentiability is lost and gradient based optimizations schemes have to be abandoned. The consequence of the asymmetry in the estimate of the joint distribution will be investigated further in the following section.

\section{Empirical investigations into the asymmetry in GPV}
\label{sec:asym}
\begin{figure}
  \centering
  \subfigure[][]{\label{fig:err:Gauss}\includegraphics[width=0.48\linewidth]{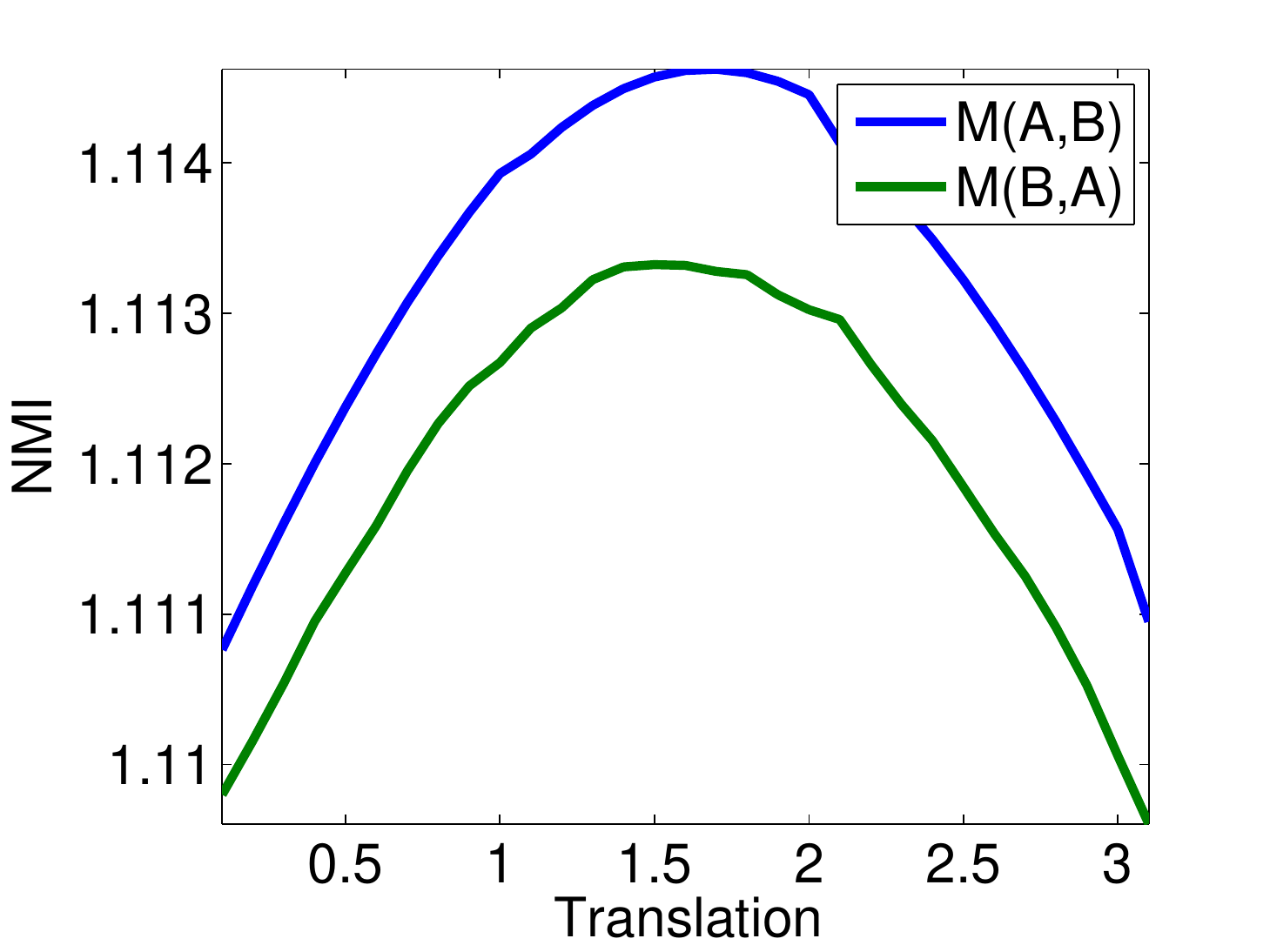}}
  \subfigure[][]{\label{fig:err:real}\includegraphics[width=0.48\linewidth]{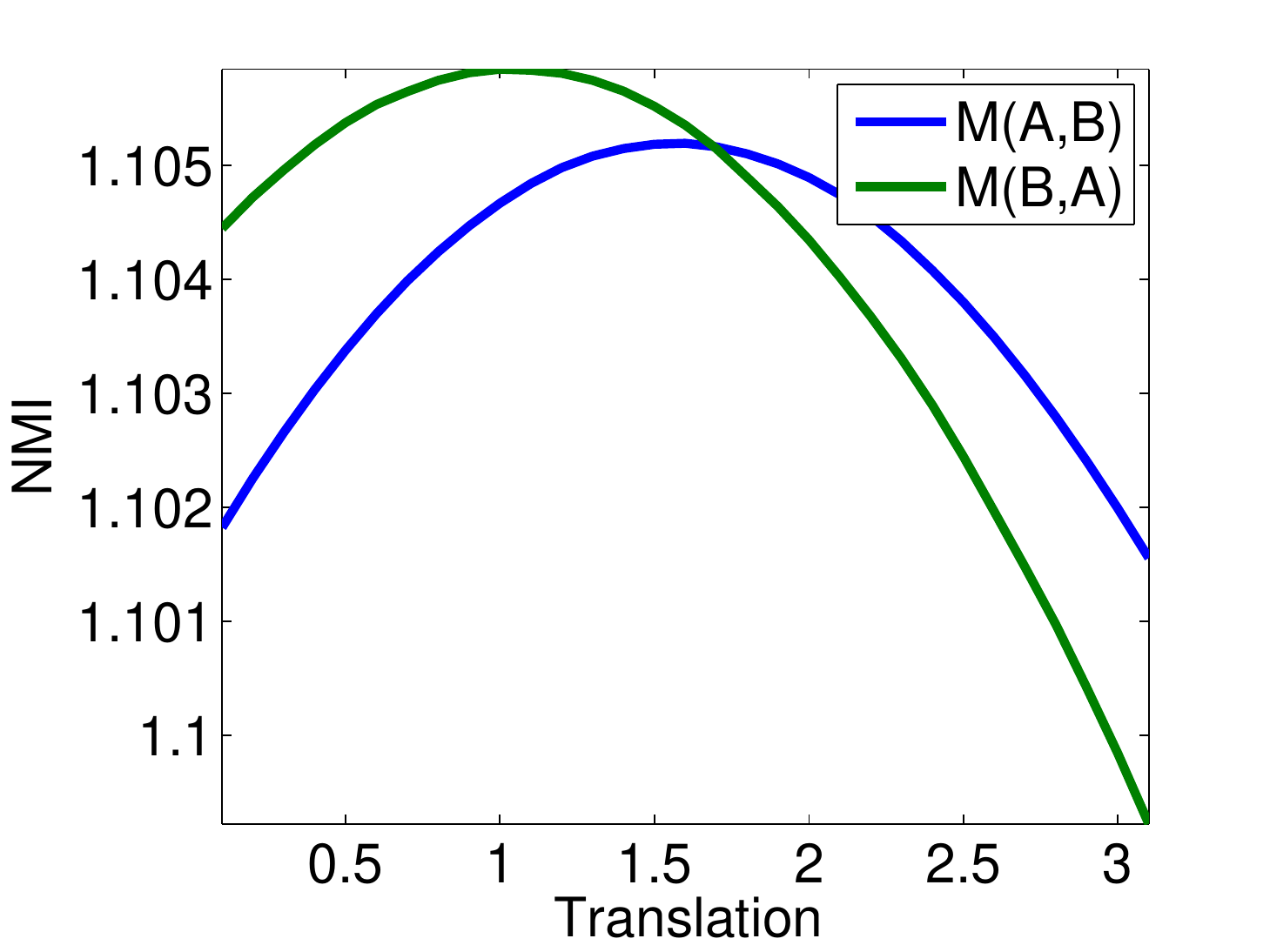}}
  \caption{GPV using NMI is asymmetric and has different optima, when comparing $\mathcal M(A,B)$ and $\mathcal M(B,A)$ under a translation along a fixed axis.  Images compared are (a) two 3-dimensional Gaussian of standard deviation $5$ and $11$, (b) baseline and followup of patient number 16 from the OASIS collection \cite{marcus2007open}.}
  \label{fig:err}
\end{figure}

GPV is asymmetric, i.e., $\mathcal M(A,B)\neq \mathcal M(B,A)$, when using GPV.  The asymmetry has been analyzed in the previous section, and in this section we will demonstrate that the asymmetry not only has theoretical but also practical implications. We start by illustrating the asymmetry of GPV used for NMI. \figurename~\ref{fig:err:Gauss} show $\mathcal M(A\circ\vec\phi,B)$ and $\mathcal M(B,A\circ\vec\phi)$  for two 3-dimensional images of spatial Gaussian with standard deviation 5 and 11 and centered in the middle of the images of size $256\times 256\times 128$. We apply a translational motion, $\vec\phi$, one image wrt.\ the other along a fixed axis and due to the symmetry of the Gassians, the points of optima are nearly identical. However, on real medical images this is not the case: In \figurename~\ref{fig:err:real}, we have plotted the cost functional $\mathcal M(A\circ\vec\phi,B)$ and $\mathcal M(B,A\circ\vec\phi)$ for pure translation of two images of baseline and followup of patient 16 from the OASIS collection \cite{marcus2007open}. The points of optima are clearly different. 

\begin{figure}
  \centering
  \subfigure[][]{\label{fig:gradientimg:gi1}\includegraphics[width=0.48\linewidth]{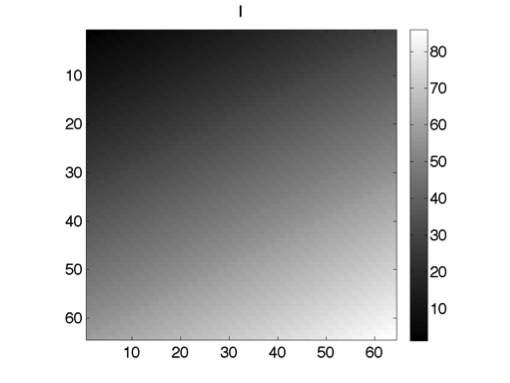}}
  \subfigure[][]{\label{fig:gradientimg:gi2}\includegraphics[width=0.48\linewidth]{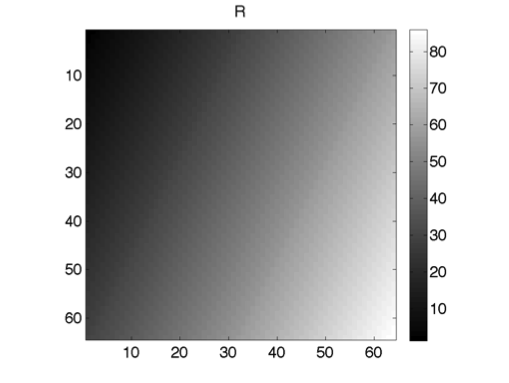}}
  \subfigure[][]{\label{fig:iso:iso1}\includegraphics[width=0.48\linewidth]{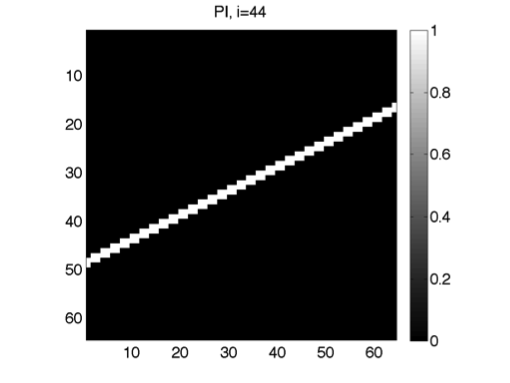}}
  \subfigure[][]{\label{fig:iso:iso2}\includegraphics[width=0.48\linewidth]{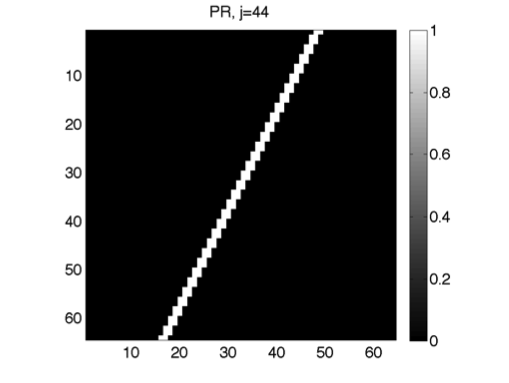}}
  \caption{ Two artificially generated images with same gradient magnitude but different directions. (a) and (b) show the images, and (c) and (d) show corresponding single isophotes extracted using the same Boxcar function.}
  \label{fig:gradientimg}
\end{figure}

\begin{figure}
  \centering
  \subfigure[][]{\label{fig:iso:iso12}\includegraphics[width=0.32\linewidth]{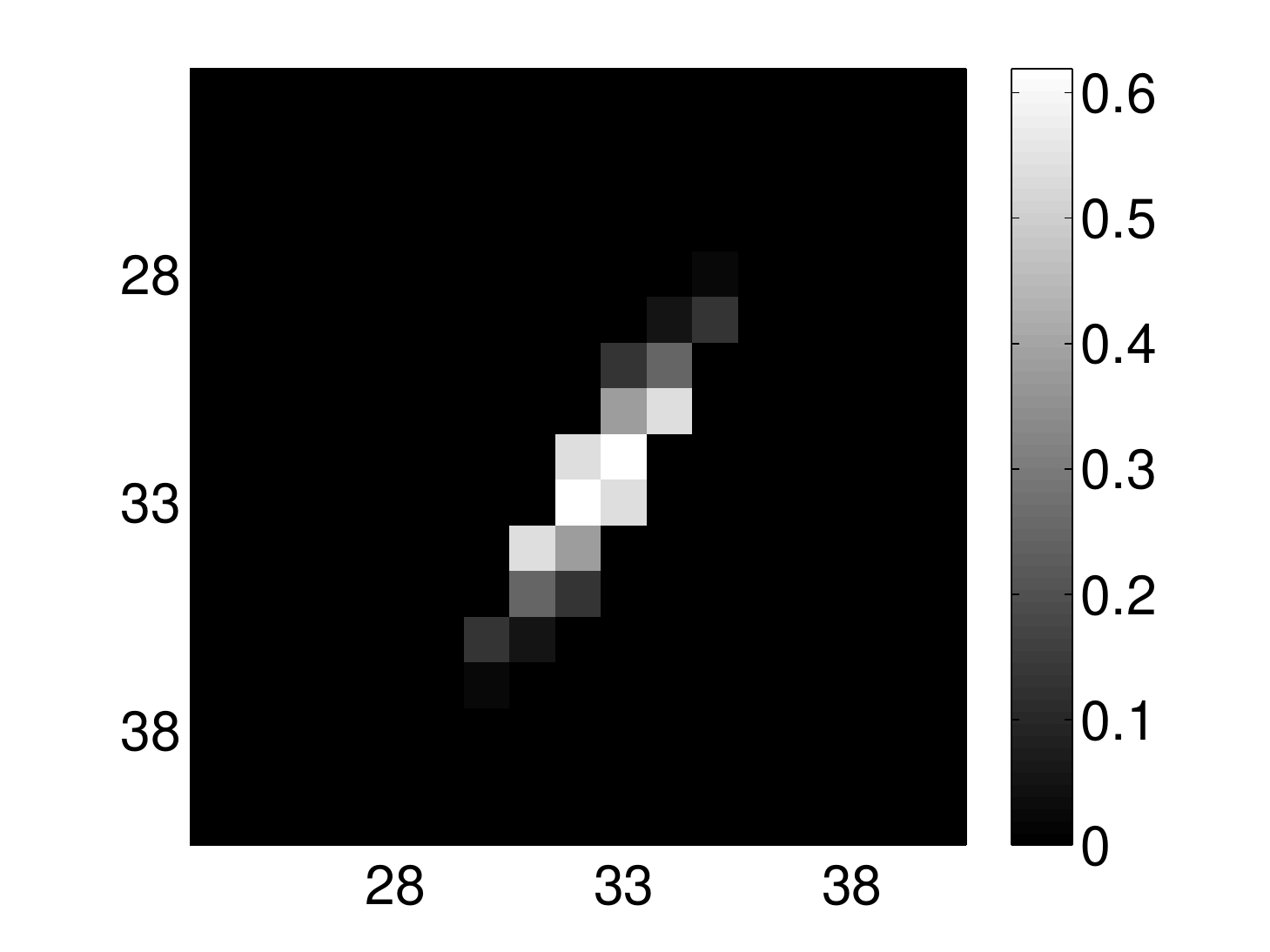}}
  \subfigure[][]{\label{fig:iso:iso_true}\includegraphics[width=0.32\linewidth]{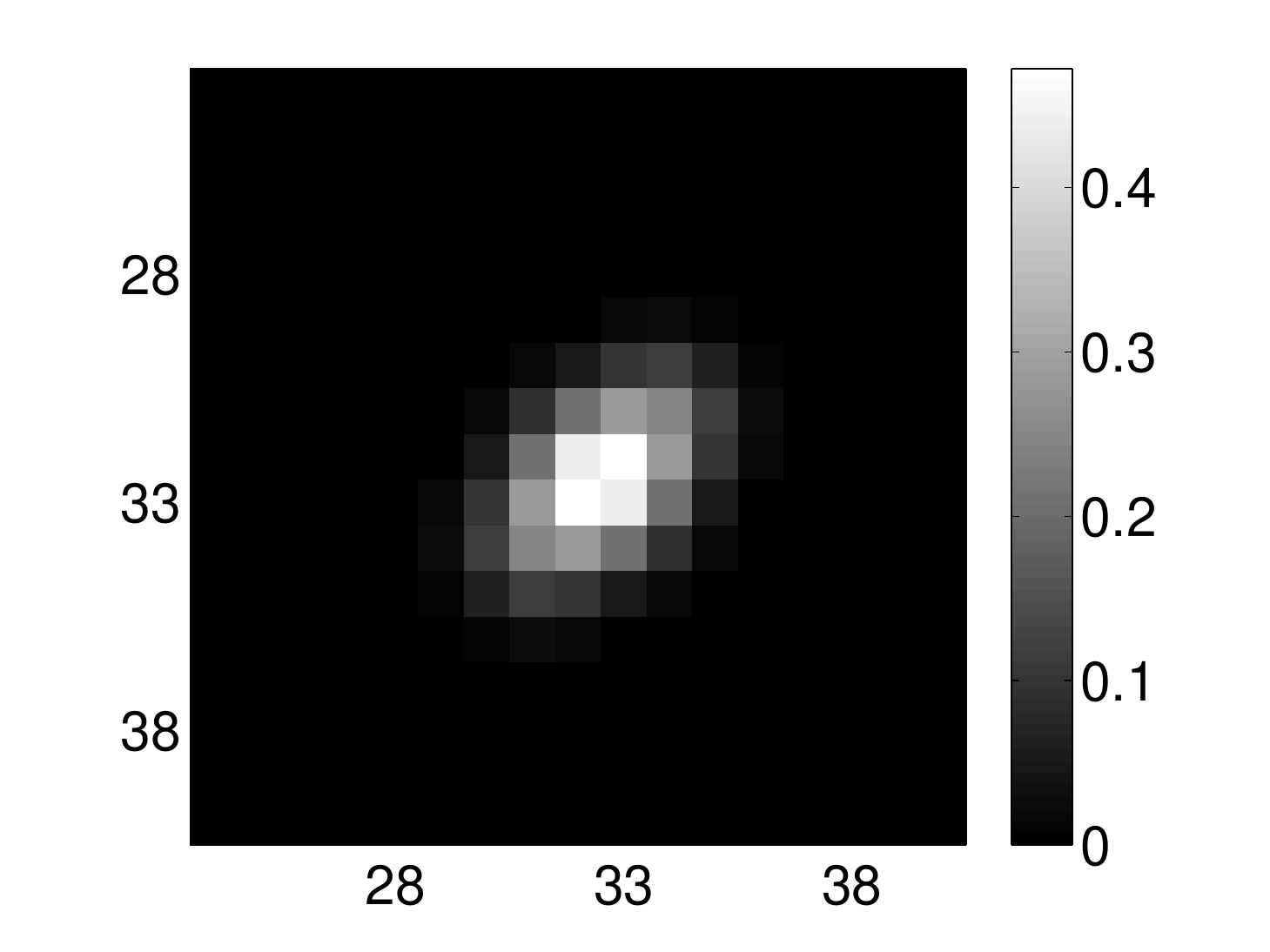}}
  \subfigure[][]{\label{fig:iso:iso21}\includegraphics[width=0.32\linewidth]{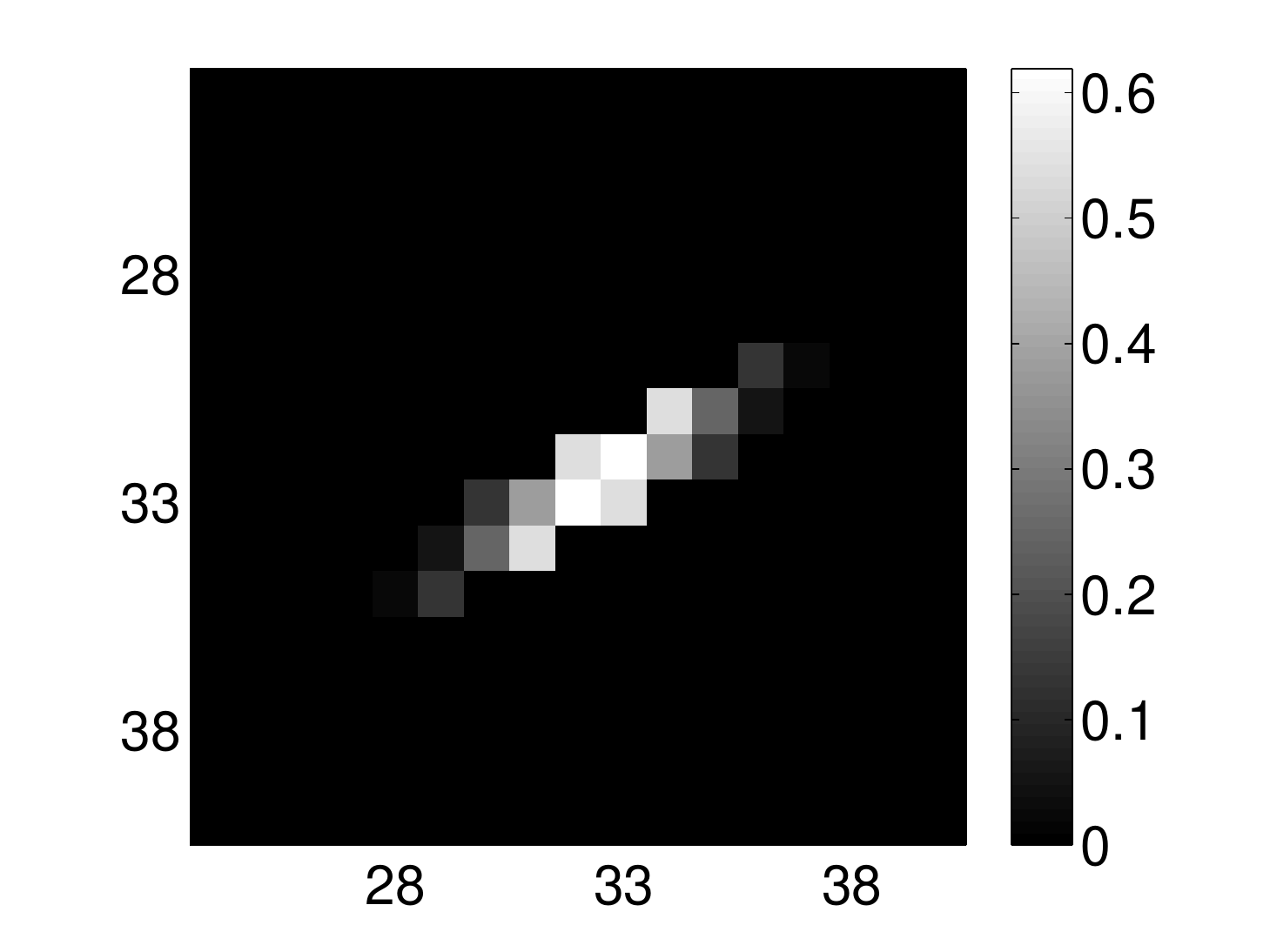}}
  \caption{The GPV approximation is asymmetric. (a) is $\mathcal M(A,B)$ and (c) is $ \mathcal M(B,A)$. (b) Is what the (a) and (c) are approximating. As the figure show the results are quite different due to the asymmetry, and as the geometrical differences between the compared isophotes, so does the asymmetry.}
  \label{fig:iso:joint}
\end{figure}

To empirically investigate this obvious asymmetry of GPV using NMI, we have constructed two images with a constant gradient, same magnitude but different direction for each as shown in \figurename~\ref{fig:gradientimg}.
We focus on a single isophote, $I(x,y)=I_0$ and $R(x,y)=R_0$, extracted using a Boxcar function.  These are shown in \figurename~\ref{fig:iso:iso1} and~\ref{fig:iso:iso2}.  The value of the joint histogram for these intensities $(I_0,R_0)$ is depicted in \figurename~\ref{fig:iso:joint} as a function of space and using various estimation techniques.  \figurename~\ref{fig:iso:joint}(a) shows the joint histogram's values when comparing \figurename~\ref{fig:gradientimg}(c) to \figurename~\ref{fig:gradientimg}(d) using GPV, i.e., where $I(x,y)=I_0$ is smoothed and intersected with $R(x,y)=R_0$ as $M(R,I)$ in GPV, and \figurename~\ref{fig:iso:joint}(c) shows the opposite case, $M(I,R)$.  For reference, in \figurename~\ref{fig:iso:joint}(b) is show the LOI estimate of the intersection of isophote $I_0$ and $R_0$.
As it can be seen, the spatial distribution of intensities is oriented according to the non-smoothed isophote.
 
\begin{figure}
  \centering
\subfigure[][]{\label{fig:gpvpw:ex2a}\includegraphics[width=0.43\linewidth]{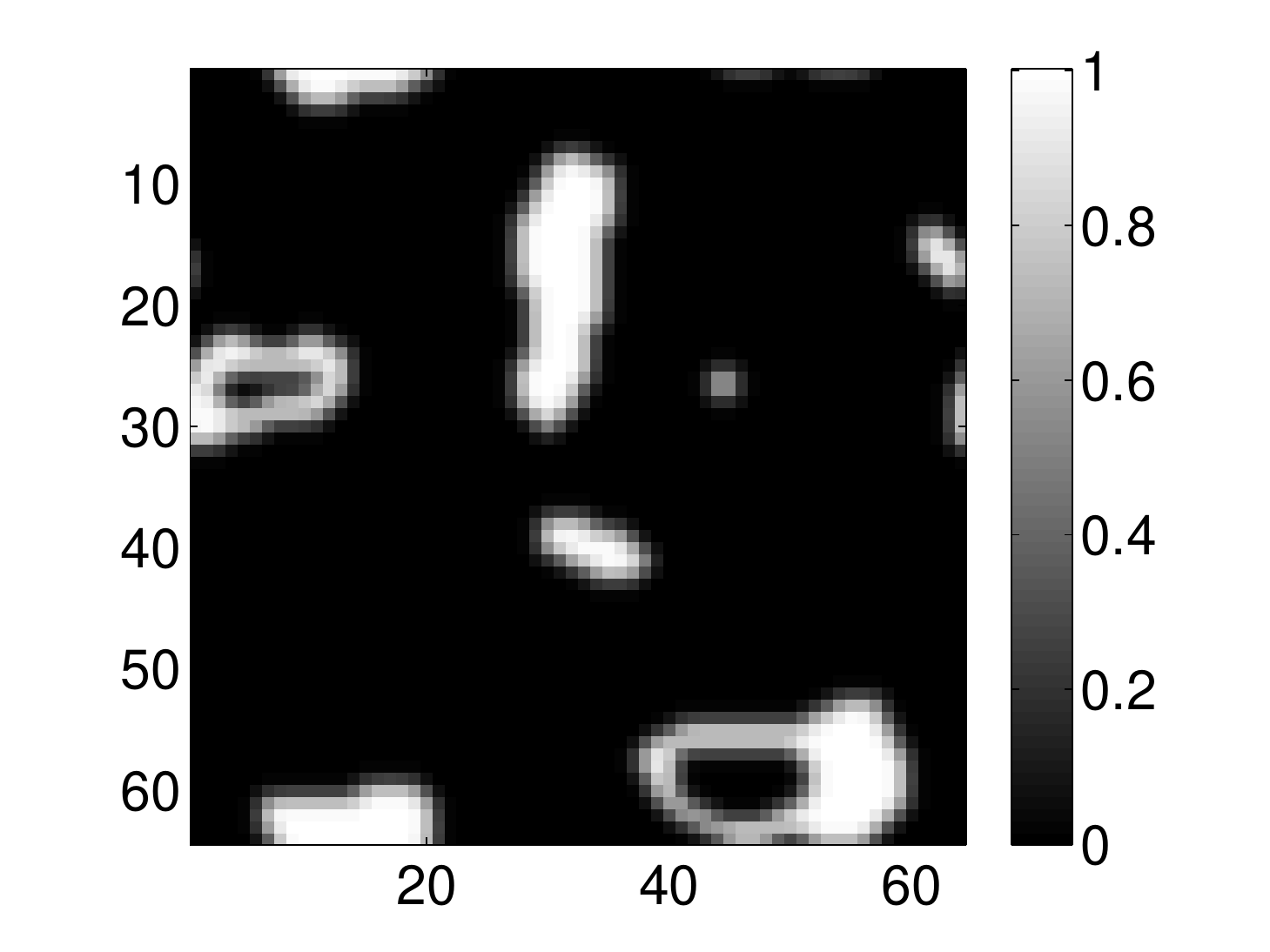}}
\subfigure[][]{\label{fig:gpvpw:ex2b}\includegraphics[width=0.43\linewidth]{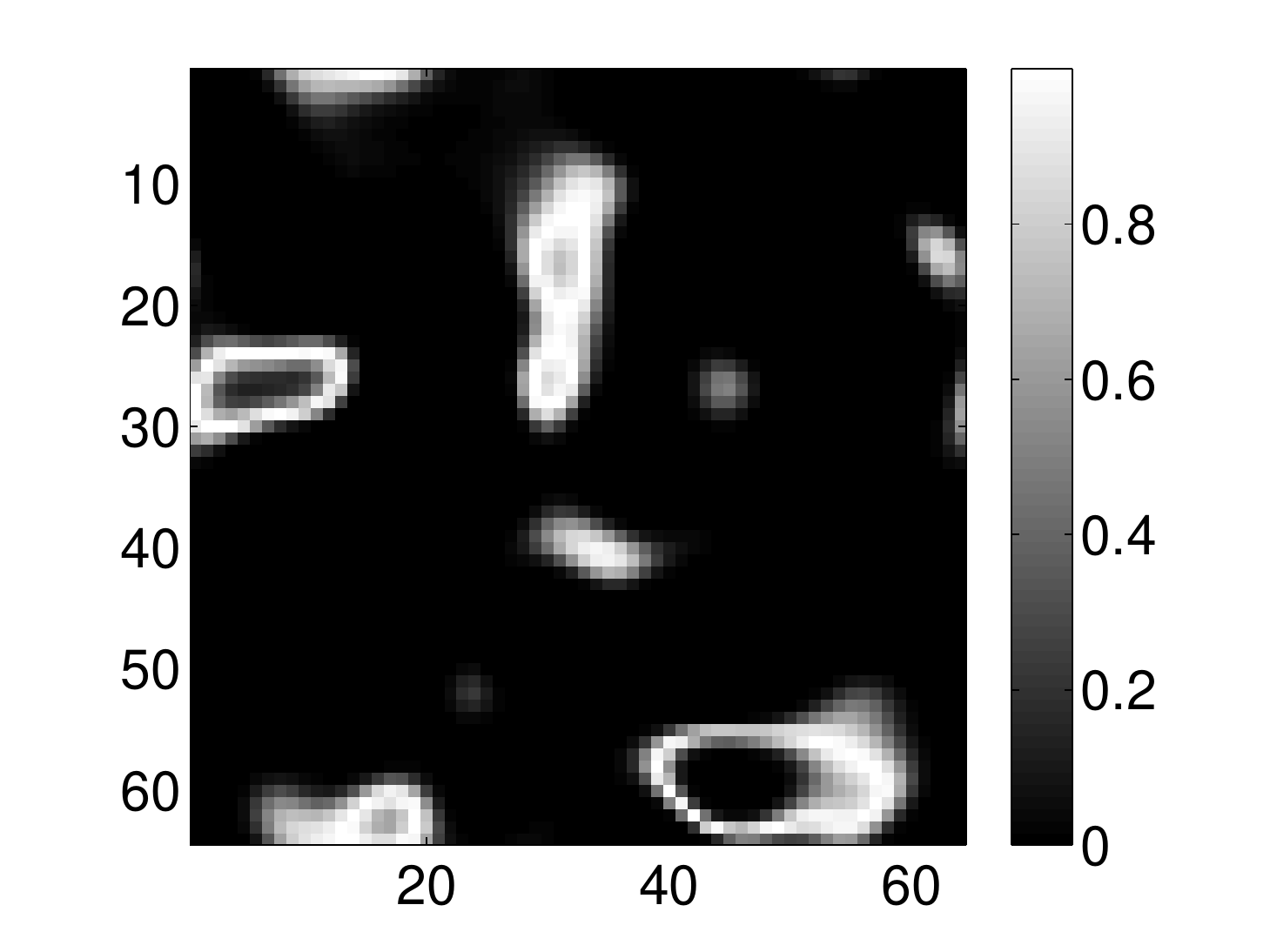}}
\subfigure[][]{\label{fig:gpvpw:ex2c}\includegraphics[width=0.43\linewidth]{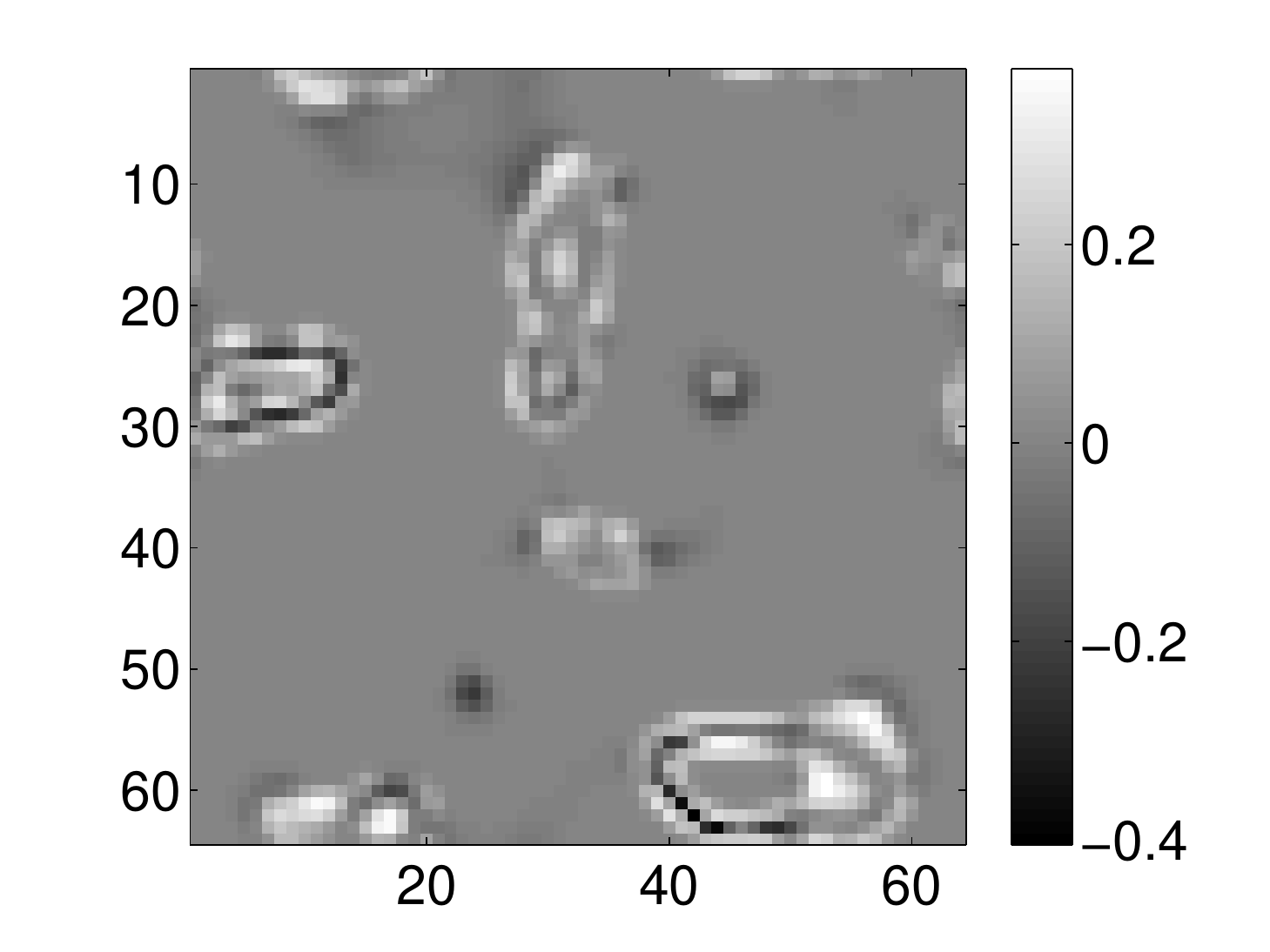}}
\subfigure[][]{\label{fig:gpvpw:ex2d}\includegraphics[width=0.43\linewidth]{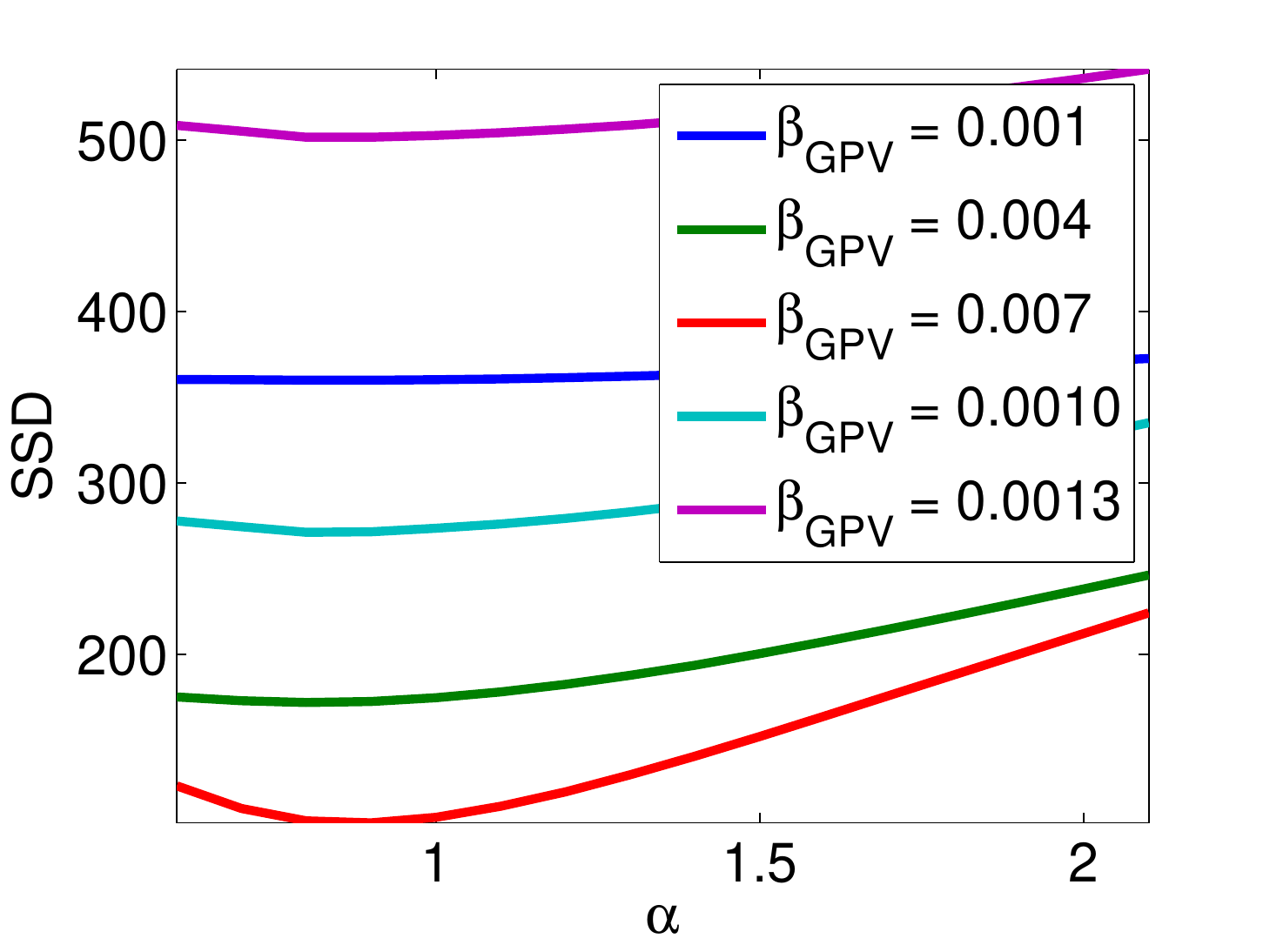}}
  \caption{ The difference between smoothing Boxcar isophotes and soft isophotes appear near high isophote curvature.  (a) The GPV isophote using$\beta_\text{GPV}=0.0013$, smoothed with $W$ using $\alpha=0.9$. (b) The PW isophote using $\beta_{\text{PW}}=0.005$. (c) the signed difference of (a) and (b), and (d) the absolute difference for a range of $\alpha$ and $\beta_{\text{GPV}}$.}
  \label{fig:gpvpw:ex2}
\end{figure}

Curvature adds further asymmetry, since the intensity moves in the direction of the center of the osculating circle, when smoothed spatially.  Thus, unless the two images curve in the exact same manner, the asymmetric smoothing of the GPV method will introduce further asymmetry in the similarity measure.  This is illustrated in \figurename~\ref{fig:gpvpw:ex2}, where an isophote is first extracted using Boxcar function, then smoothed spatially to give the image shown in \figurename~\ref{fig:gpvpw:ex2a}, and this is to be compared with isophote extracted with as a soft isophote as shown in \figurename~\ref{fig:gpvpw:ex2b}.  It can be seen that the images differ especially, where isophotes have high curvature.  To substantiate this qualitative conclusion, we have conducted the following experiment:  For a fixed image, an image of a given isophote is extracted using the 2 different methods, 1) PW as a soft isophote with fixed width $\beta_{\text{PW}}=0.005$, and 2) GPV as an isophote extracted using a Boxcar with varying width $\beta_{\text{GPV}}$ followed by spatial smoothing with a Gaussian of varying width $\alpha$. Thus, for a fixed image with PW isophote width $\beta_{\text{PW}}$, we have searched for the values of $\beta_{\text{GPV}}$ and $\alpha$ such that they minimize the sum of squared differences between the two isophote images shown in \figurename~\ref{fig:gpvpw:ex2c}.
Notice in particular, the difference between the two images of the isophotes is largest near high curvature of the original isophote.

\begin{figure*}
  \centering
  \subfigure[][]{\label{fig:asym:scale_1}\includegraphics[width=0.32\linewidth]{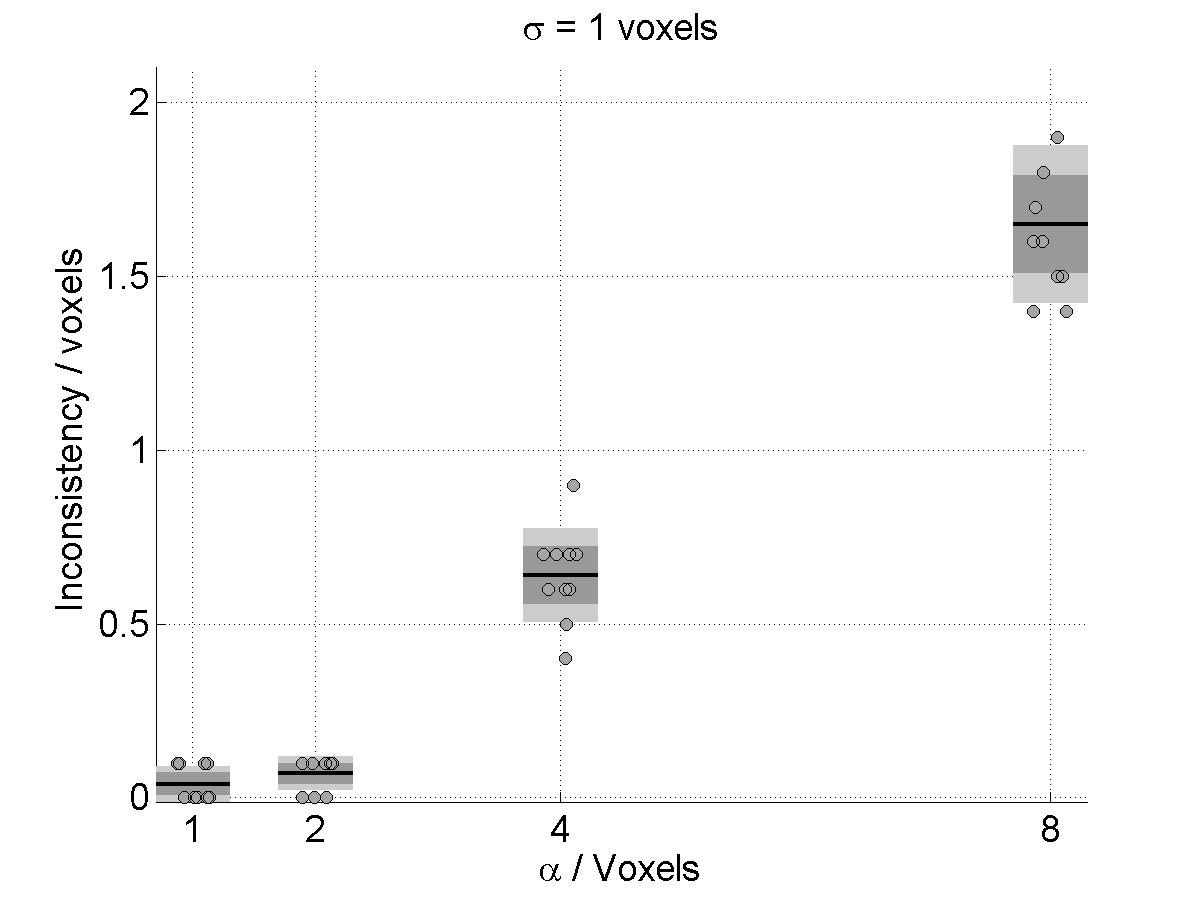}}
  \subfigure[][]{\label{fig:asym:scale_2}\includegraphics[width=0.32\linewidth]{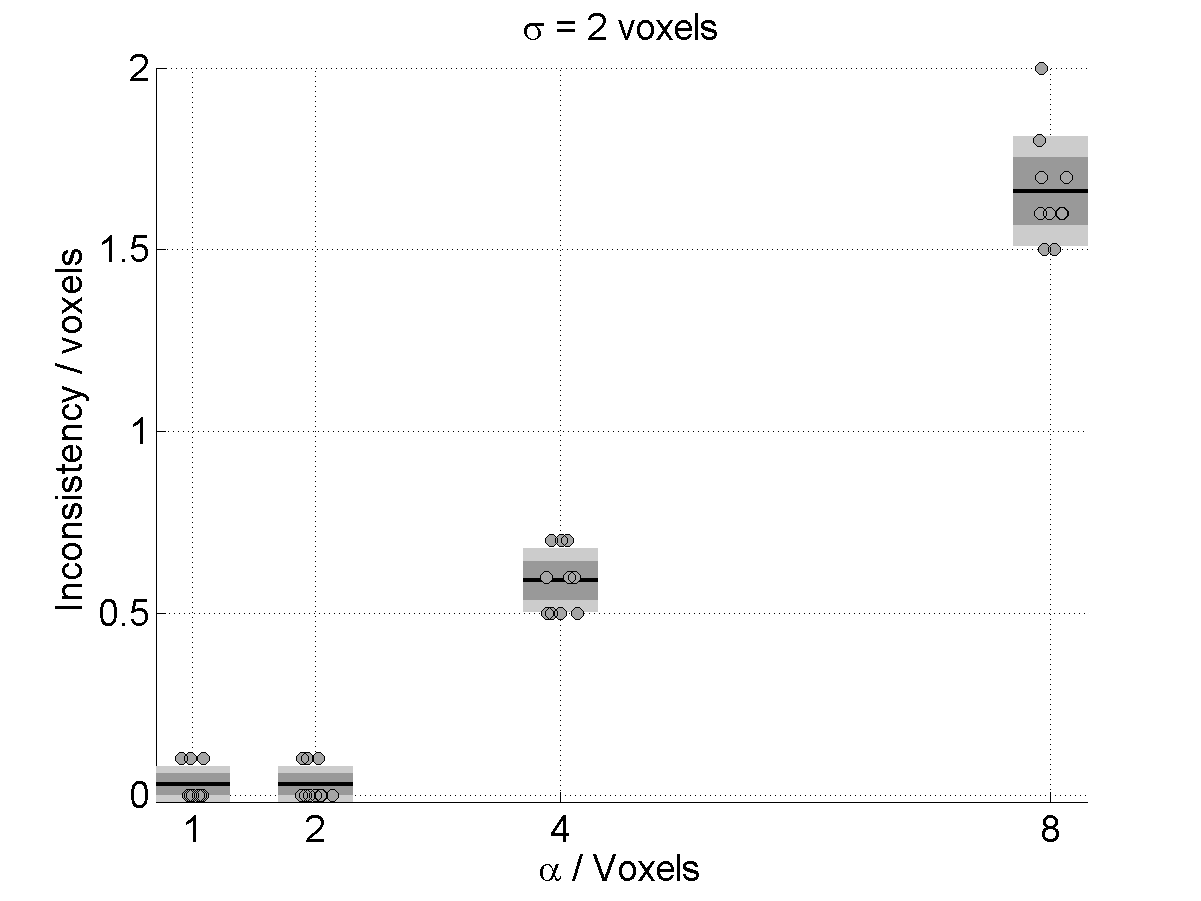}}
  \subfigure[][]{\label{fig:asym:scale_4}\includegraphics[width=0.32\linewidth]{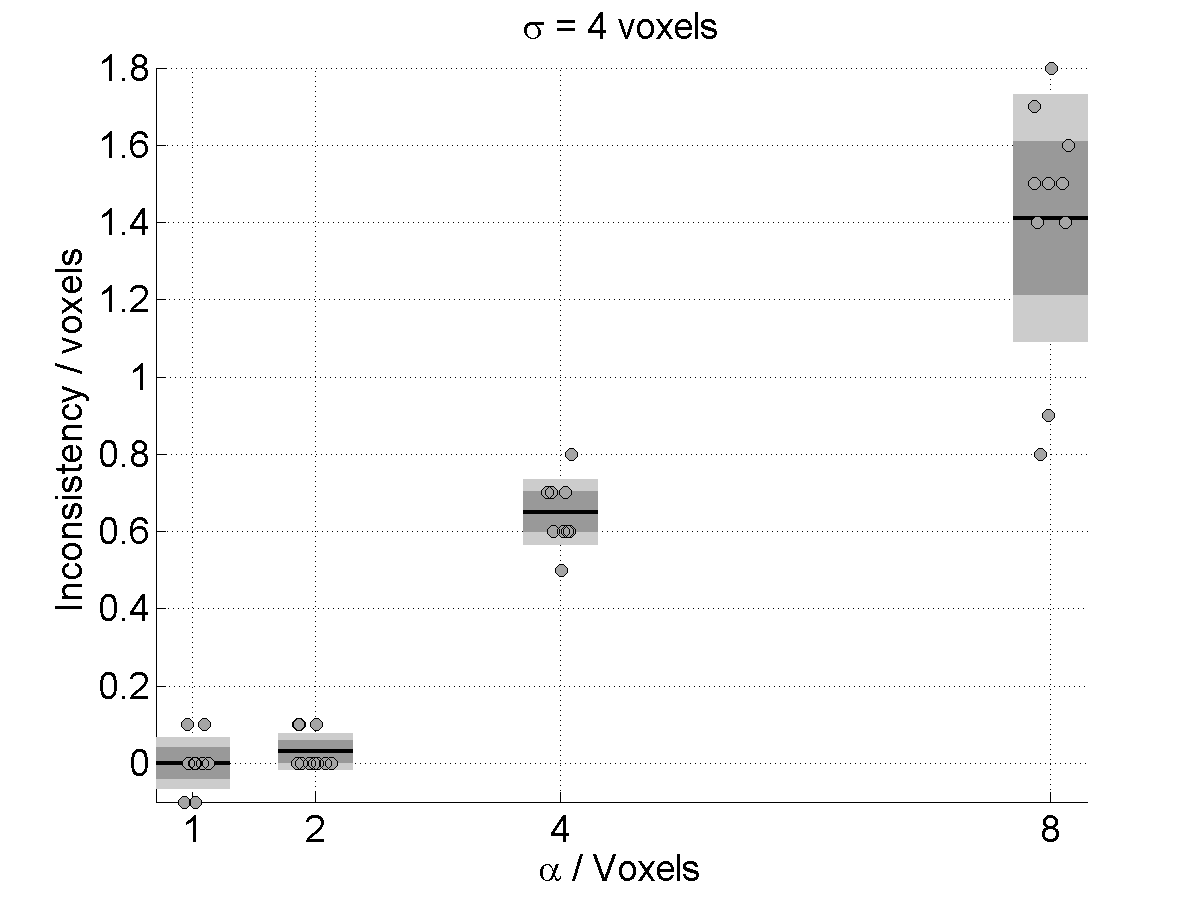}}
  \caption{GPV using NMI gives inconsistent optimization results for a simple, artificial translation, and the inconsistency depends linearly on $\alpha$ but not on $\sigma$. For each boxplot, the circles represent individual measurement with slight noise added in the horizontal direction for legibility, the black line denotes the mean, the dark and light gray areas denote the 50\% and 75\% fractiles.}
  \label{fig:asym}
\end{figure*}

\begin{figure}
  \centering
  \subfigure[][]{\label{fig:asym:gpv_k02}\includegraphics[width=0.75\linewidth]{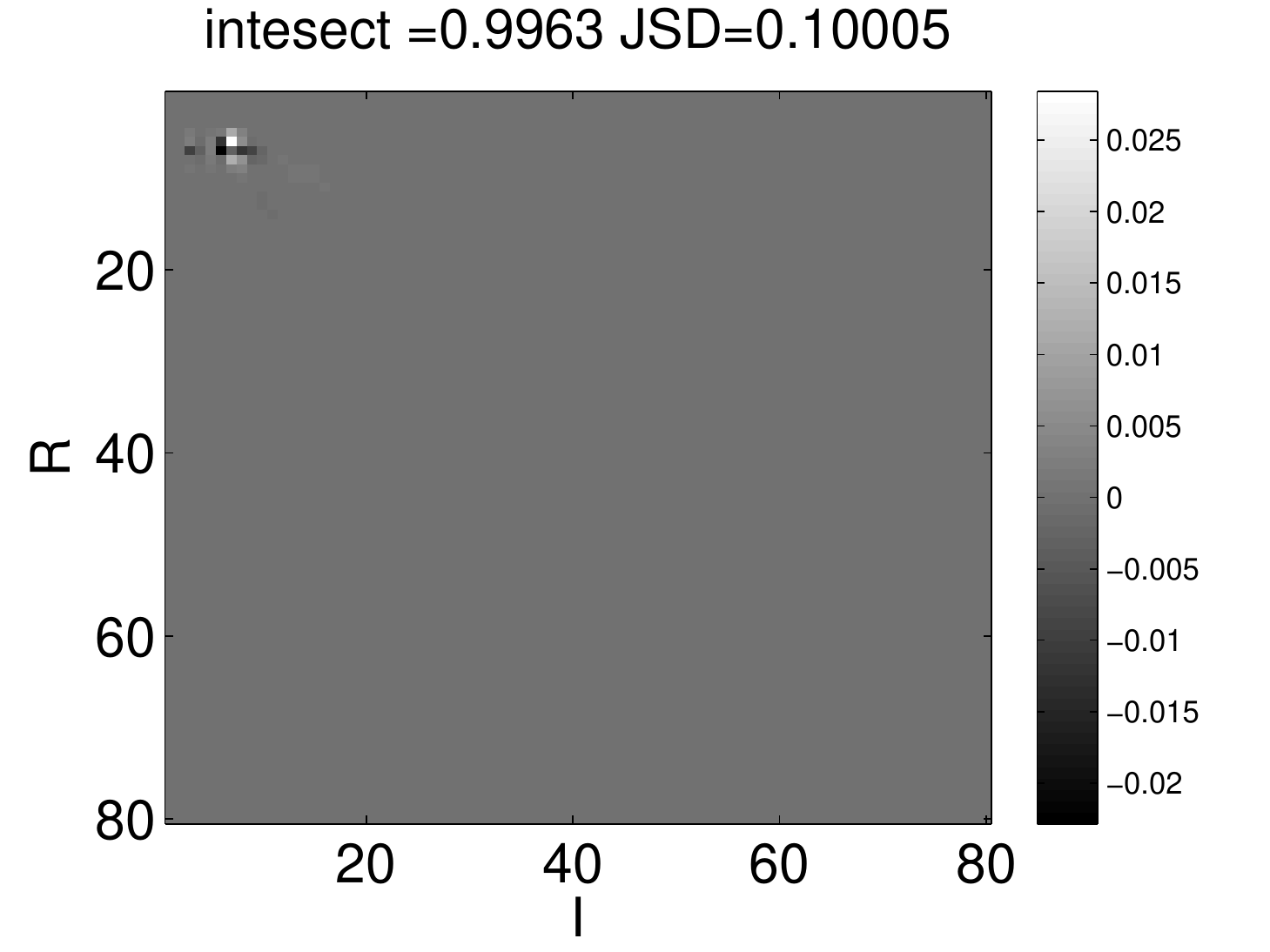}}
  \subfigure[][]{\label{fig:asym:gpv_k8}\includegraphics[width=0.75\linewidth]{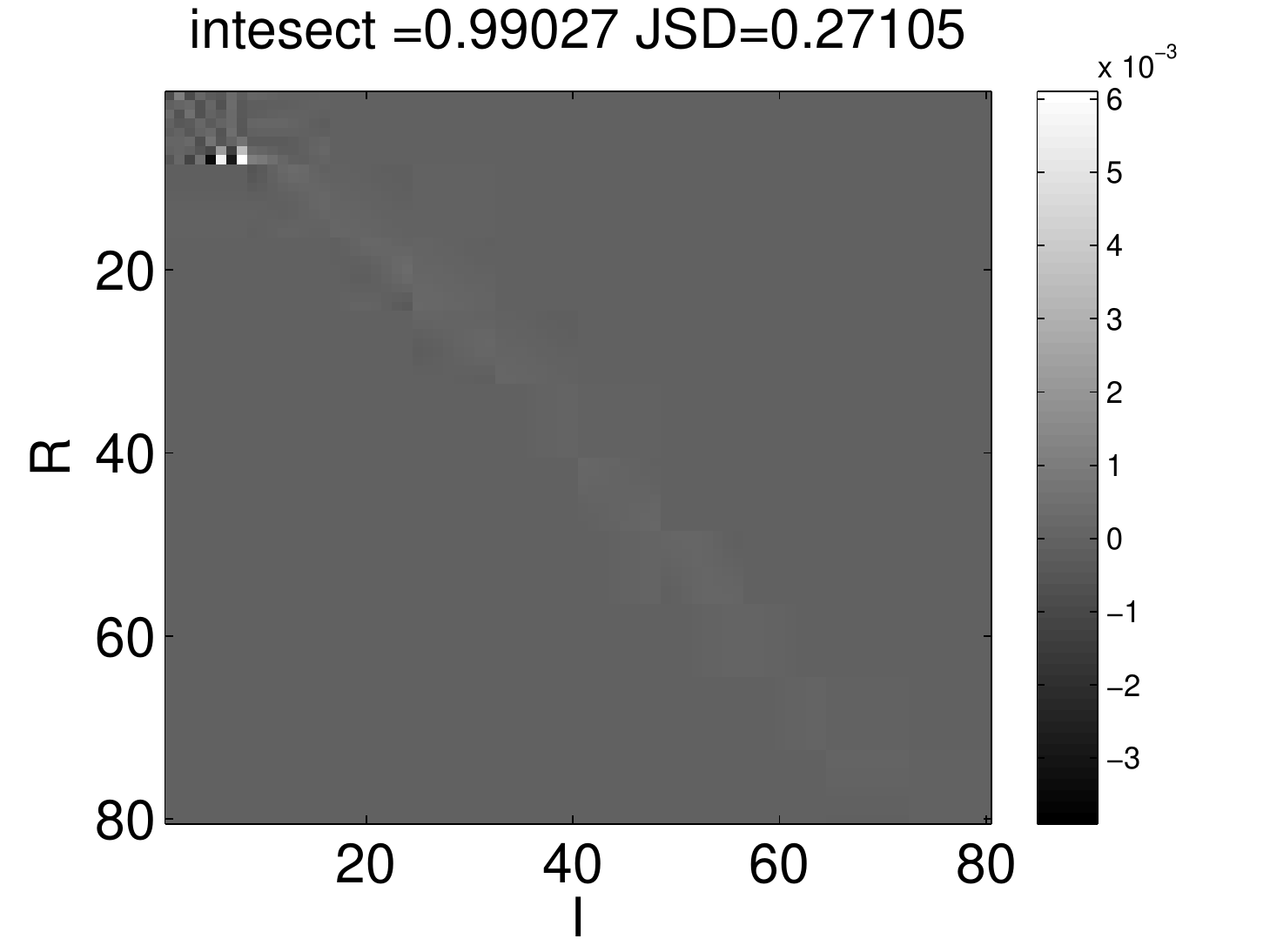}}
  \caption{ The asymmetry of GPV in the estimated densities. We have subtracted the joint density distribution estimated in $\mathcal M(A,B)$ from the estimated density distribution in $\mathcal M(B,A)$, at 2 different image scales with $\sigma=1$, $4$ and $\alpha=0.2$. The Jensen-Shannon divergence is $0.10005$ (a) and $0.27105$ for (b) }
  \label{fig:asym2}
\end{figure}

To empirically evaluate the degree of asymmetry as a function of $\alpha$, we have conducted the following experiment: For 10 baseline and followup images from \cite{marcus2007open}, we have rigidly registered the baseline and followup pair using NMI and GPV with a very small $\alpha$, and then for a range of $\alpha$s measuring the spatial asymmetry in the similarity measure along the x-axis caused by the increase in $\alpha$. This is repeated for a range of $\sigma$ values. The result is illustrated in \figurename~\ref{fig:asym}.  The experiment reveals that smoothing of the image does not eliminate the problem, and as our investigations show, asymmetry persist over all image scales.
The asymmetry can also be observed in the joint density estimates. In \figurename~\ref{fig:asym2} is shown the difference between the joint density used to evaluate $\mathcal M(B,A)$ and $\mathcal M(B,A)$ for 2 different values of $\alpha$.
The difference is seen to be non-negligible for both scales, and thus cannot be ignored.

To summarize, the GPV is asymmetric, and the degree of asymmetry increases proportionally to curvature of the isophotes as well as to $\alpha$. The asymmetry cannot be alleviated using image smoothing, and we conclude that GPV does not offer inverse consistent registration.

\section{Empirical comparison of PW and GPV by Scales}
\label{sec:scales}
In the following and using NMI, we will empirically evaluate and compare PW and GPV in terms of scales, i.e., the influences of the different kernels on the similarity measure, NMI, and the estimated joint density distribution to give intuition about the influence of different scales on NMI. Two types of algorithms for GPV and PW have been implemented: A fast cubic uniform B-spline approach (hereafter referred to as B-spline), which is described and analyzed in the next section, and a version based on Gaussian kernels. For direct comparison of B-splines and Gaussians we have estimated the variance of a B-spline to be $\sigma\approx 0.6$. This allows us to investigate the effect of tuning the standard deviations of each of the kernels for both PW and GPV. We note here that some computational restrictions imposed on GPV due to computational complexity, thus a Gaussian with local support has been used, i.e., very small values are truncated.  We have performed intra subject registration using rigid registration on a series of T1 weighted MRI of the human brain from different subjects \cite{marcus2007open}. For each subject we registered a followup to the baseline, such that the pair the two volumes are aligned close to optimally (within 0.5 voxel). For a given direction (x-axis) we have translated one of the two with $+/-1.5$ voxels in steps of 0.1 voxel and calculated the NMI similarity. This has been repeated for a wide range of kernels in the different spaces, i.e., different $\sigma$, $\beta$, and $\alpha$ including our fast B-spline based algorithm for 10 different subjects. 

\subsection{Spatial scale, $\sigma$}
\begin{figure}
  \centering
  \subfigure[][]{\label{fig:imgsmooth:PW}\includegraphics[width=0.48\linewidth]{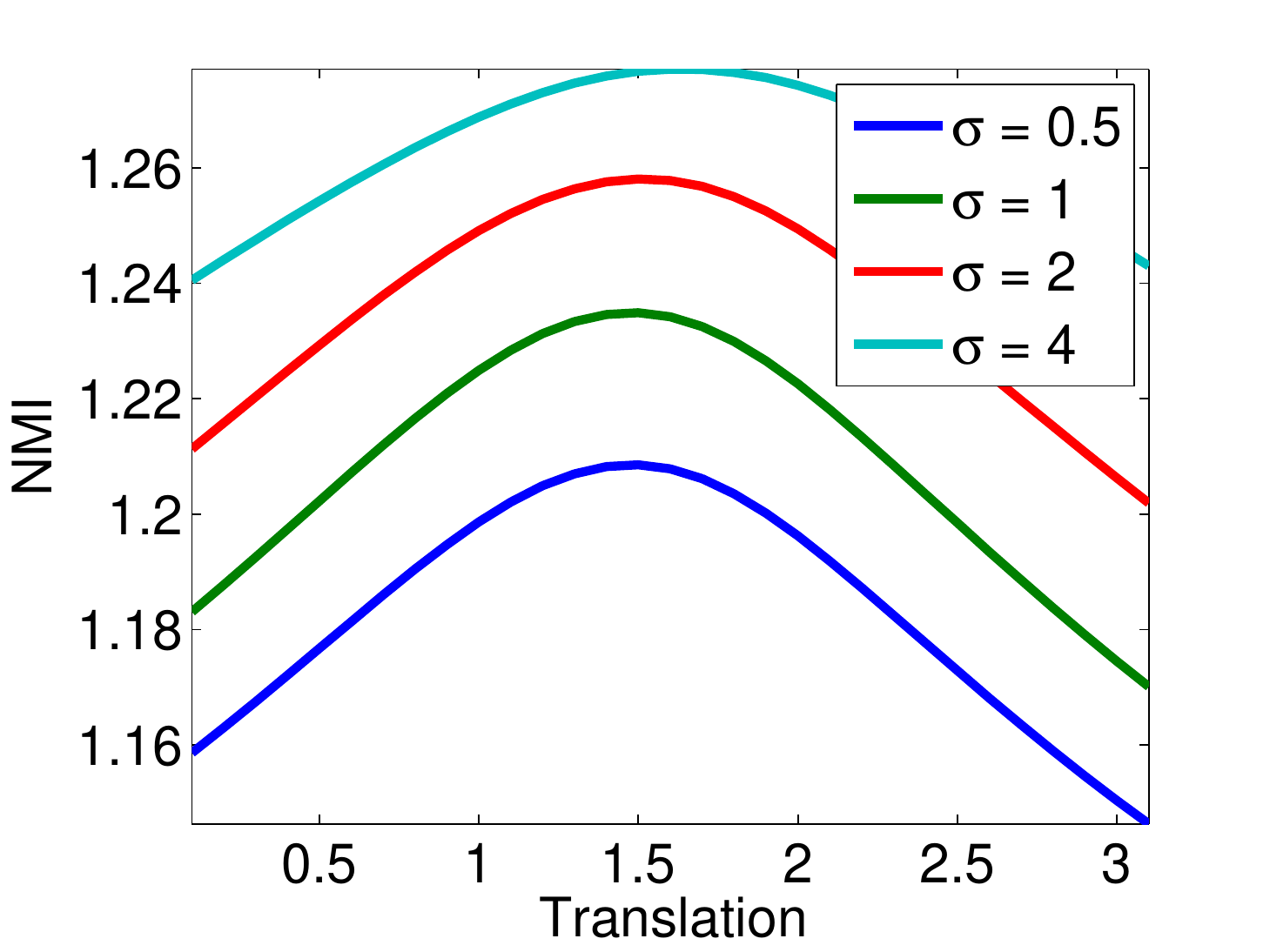}}
  \subfigure[][]{\label{fig:imgsmooth:PW_spline}\includegraphics[width=0.48\linewidth]{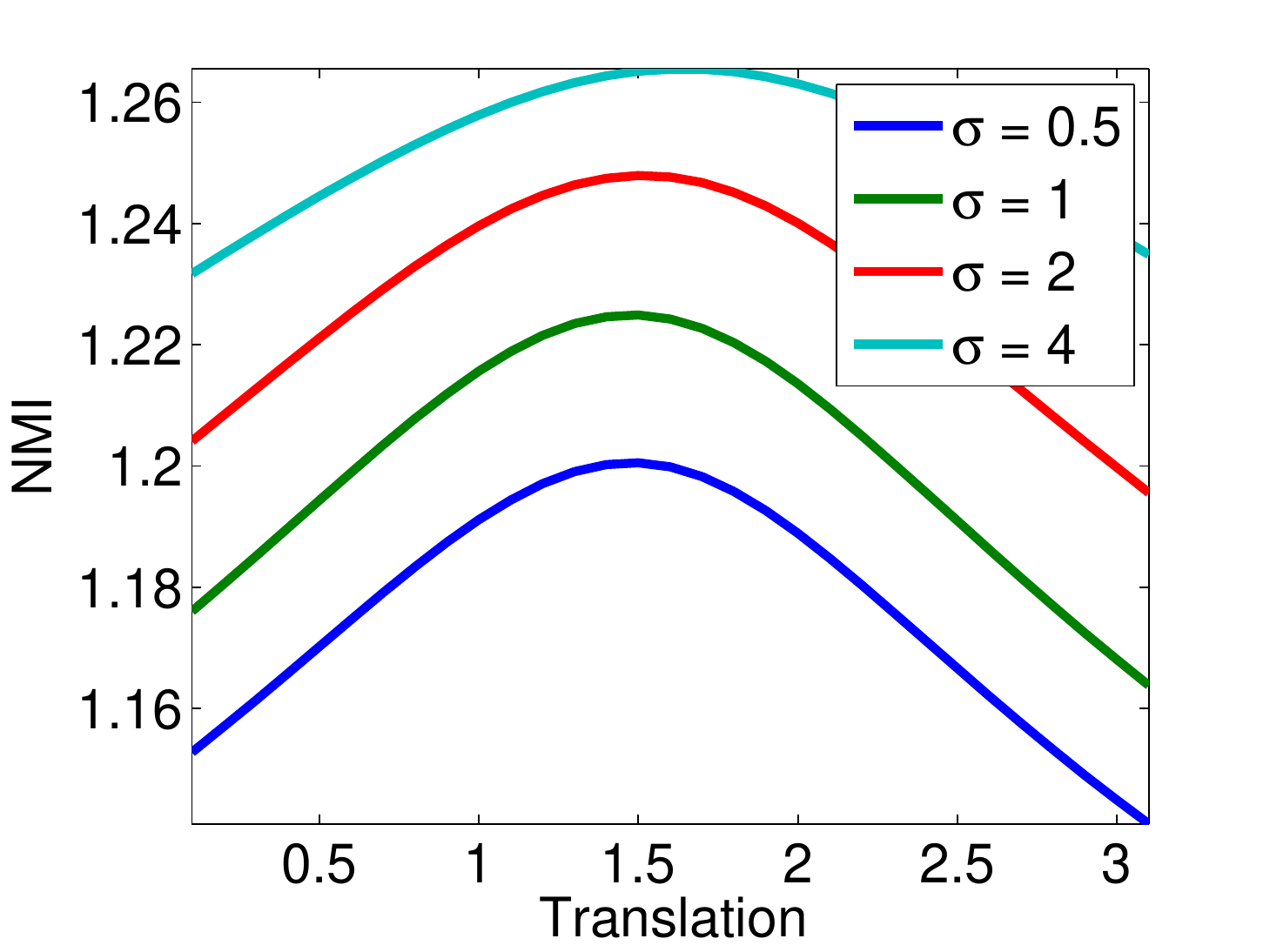}}
  \subfigure[][]{\label{fig:imgsmooth:GPV}\includegraphics[width=0.48\linewidth]{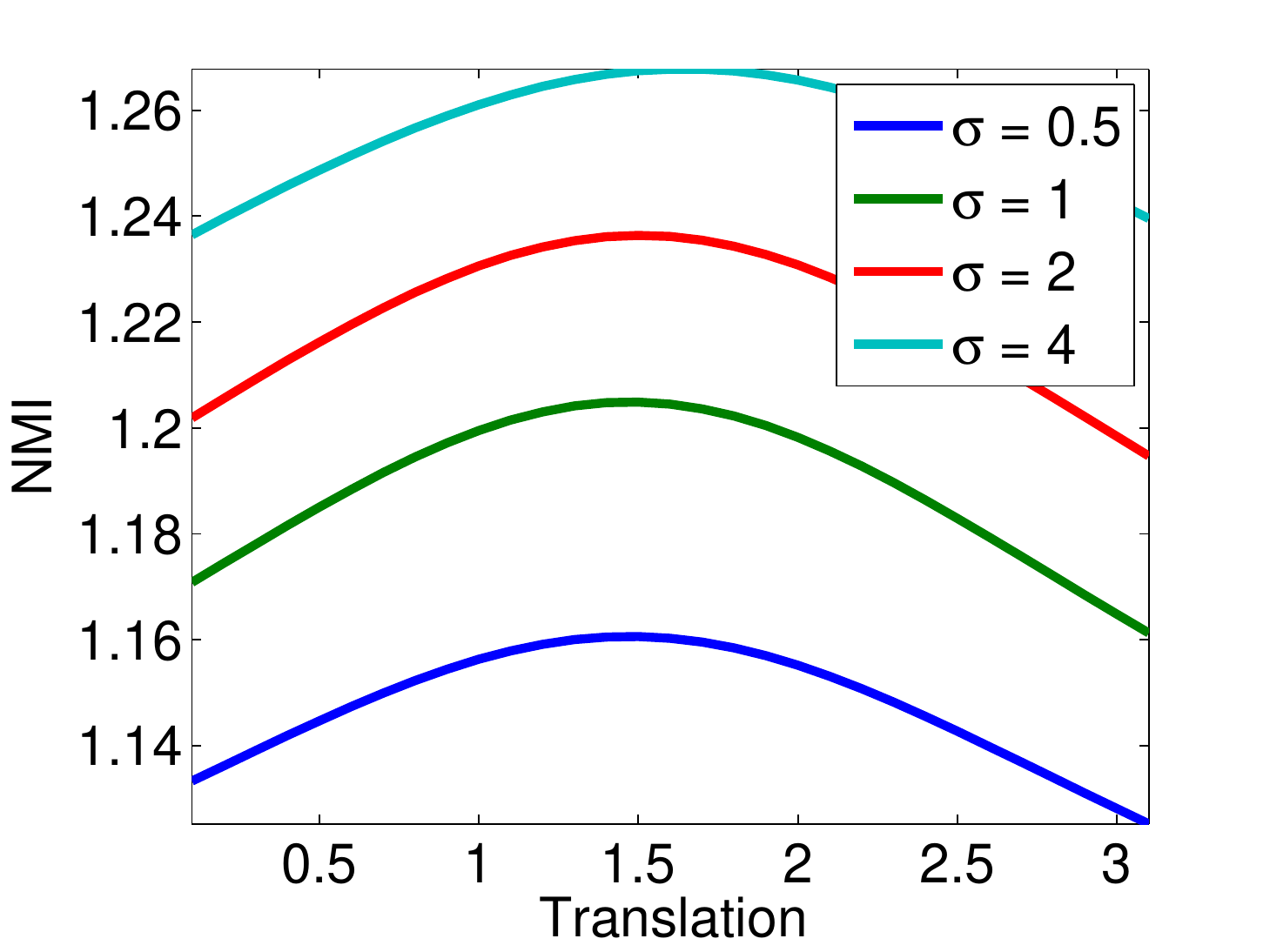}}
  \subfigure[][]{\label{fig:imgsmooth:GPV_spline}\includegraphics[width=0.48\linewidth]{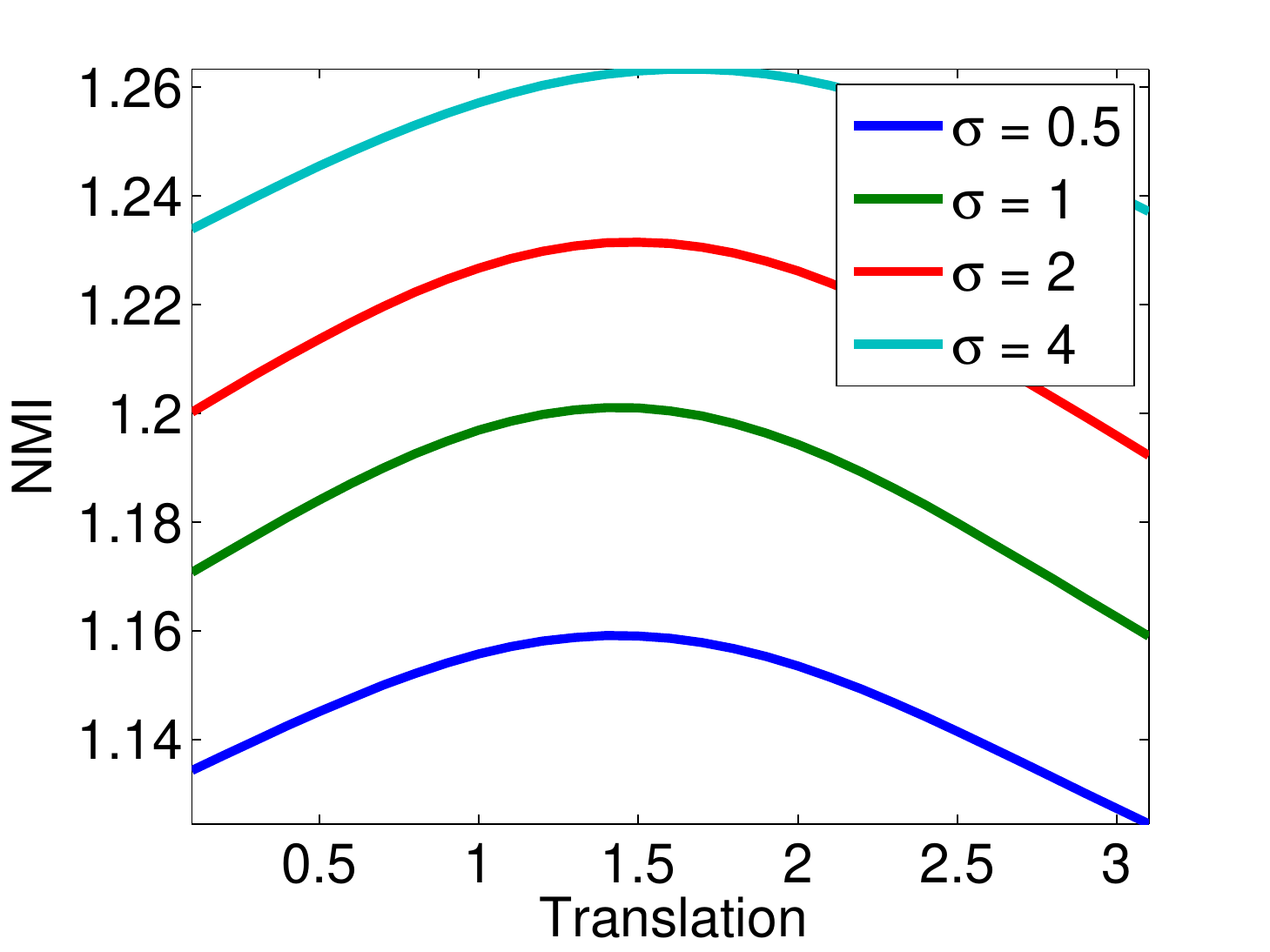}}
  \caption{ The effect of image smoothing on the objective function (NMI) using the different density estimation schemes: (a) The PW using a Gaussian kernel $\beta=0.6$, (b) PW using cubic b-spline, and (c) GPV using a Gaussian $\alpha=0.6$, (d) GPV using cubic B-spline.}
  \label{fig:imgsmooth}
\end{figure}

\begin{figure}
  \centering
  \subfigure[][]{\label{fig:imgsmoothHRI:PW_k02}\includegraphics[width=0.48\linewidth]{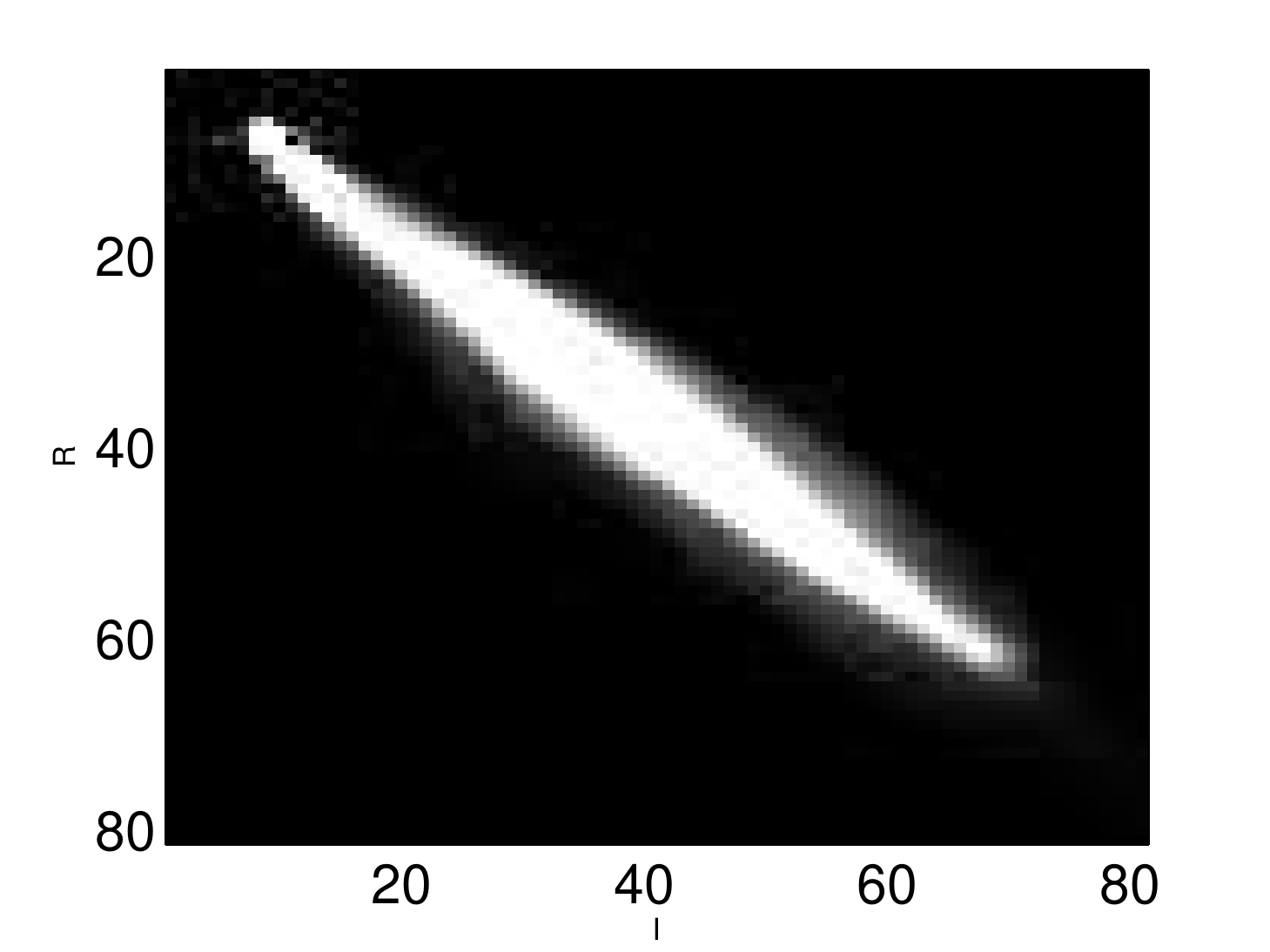}}
  \subfigure[][]{\label{fig:imgsmoothHRI:PW_k8}\includegraphics[width=0.48\linewidth]{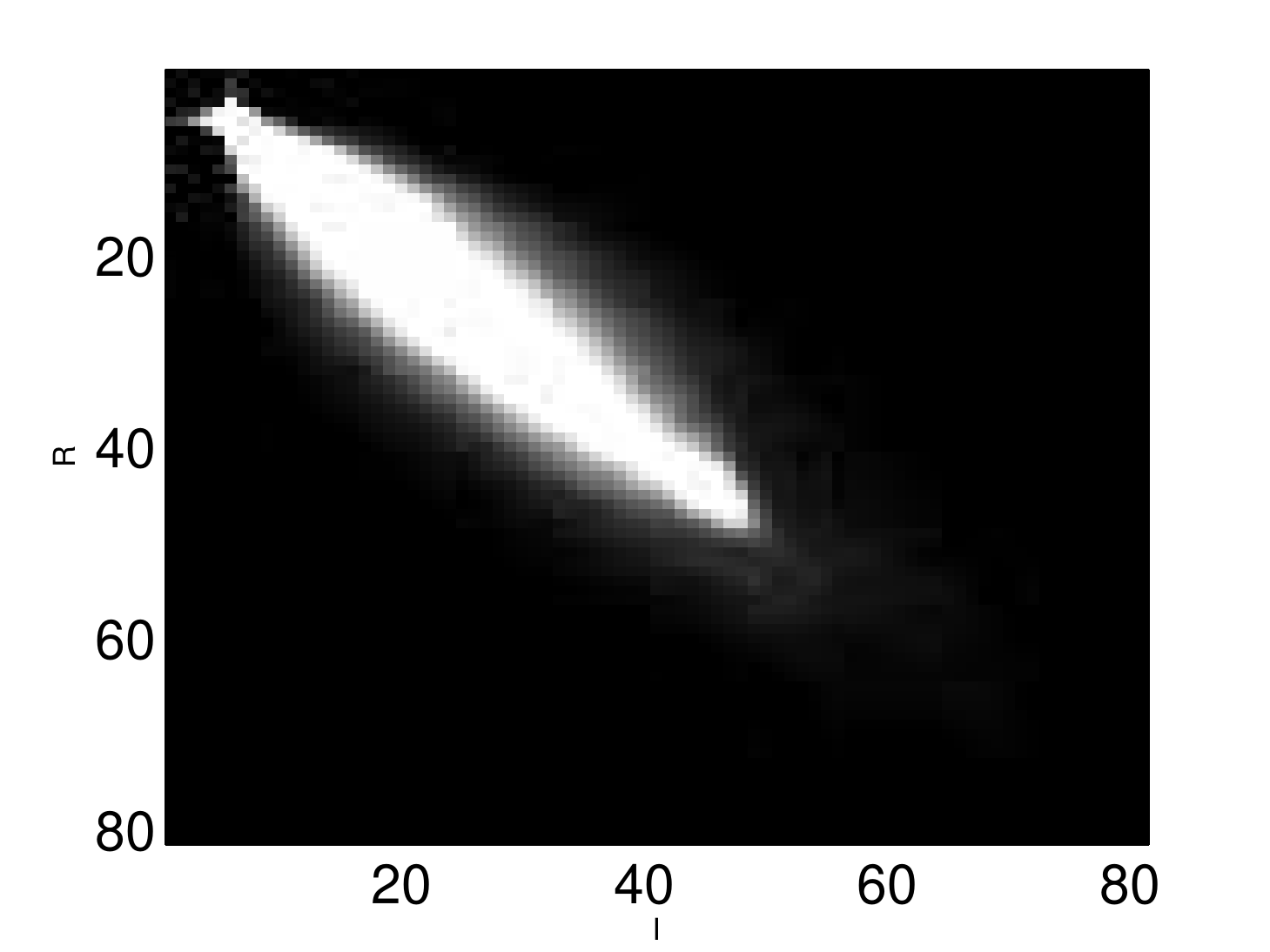}}
  \subfigure[][]{\label{fig:imgsmoothHRI:GPV_k02}\includegraphics[width=0.48\linewidth]{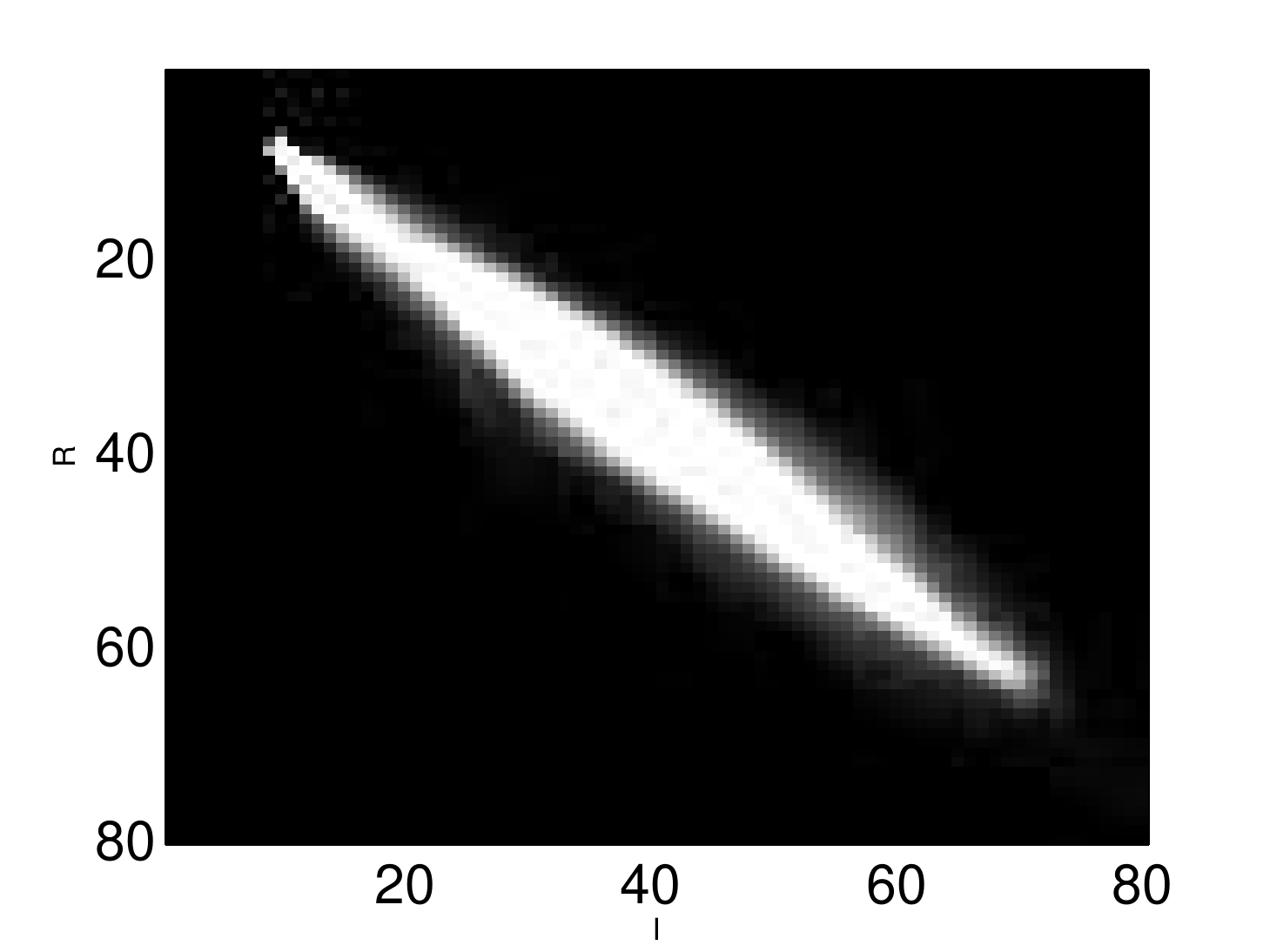}}
  \subfigure[][]{\label{fig:imgsmoothHRI:GPV_k8}\includegraphics[width=0.48\linewidth]{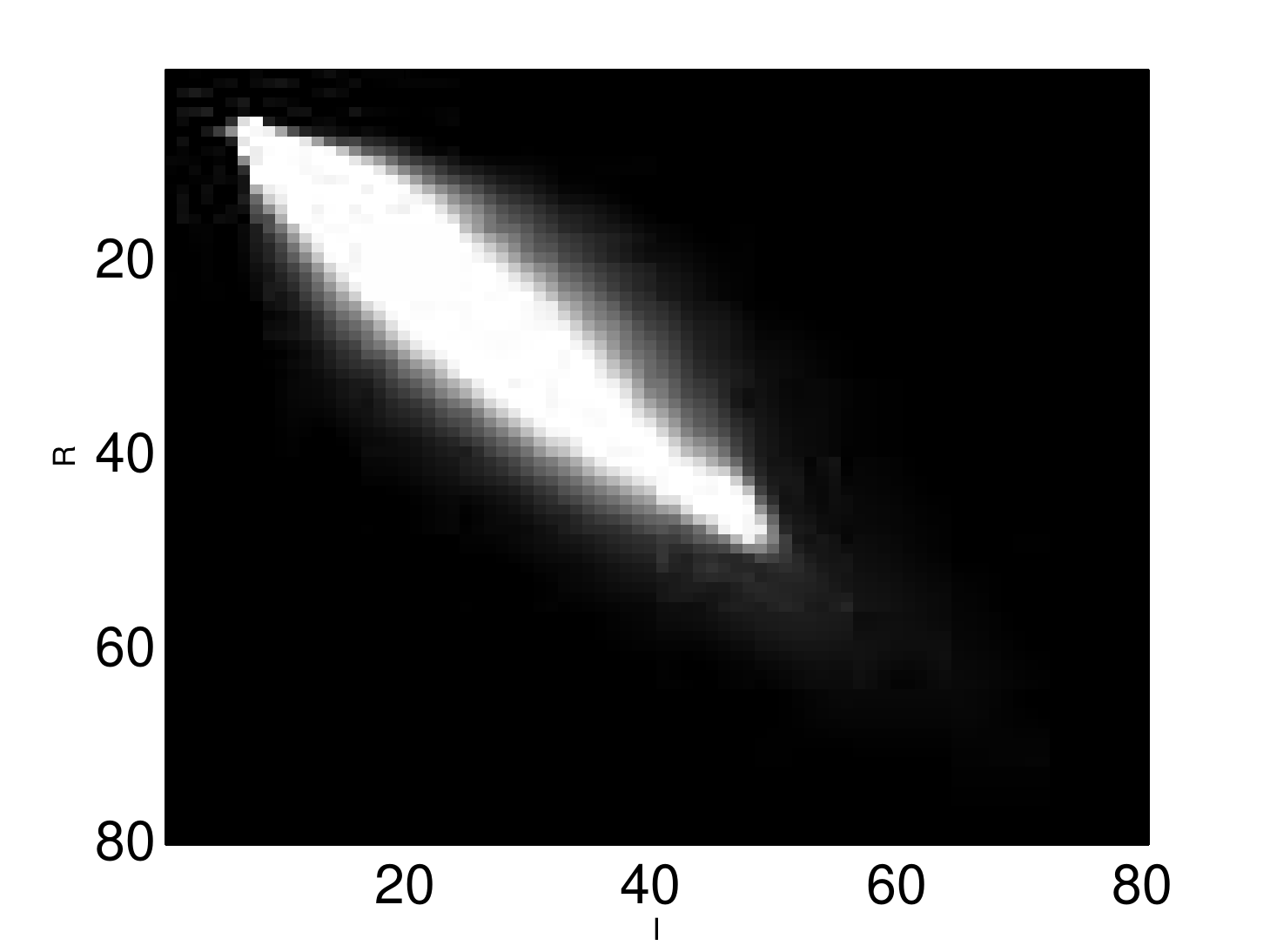}}
  \caption{ The effect of image smoothing on the joint density using different estimation schemes: (a) \& (b) The PW using $\beta=0.3$, and (a) $\sigma=0.5 $ and (b) $\sigma=2;$ (c) \& (d) GPV using $\alpha=0.6$,, $\beta=0.3$ and (c) $\sigma=0.5 $ and (d) $\sigma=2$.}
  \label{fig:imgsmoothHRI}
\end{figure}

When registering images, most algorithms exploit the scale space of the images by smoothing of the image with the kernel $K$. The idea is to capture large scale structures of the images to get closer to the optima before switching scale in order to capture structure at a finer scale. The actual influence on the different similarity measures has only been vaguely investigated in the literature. In spite of this uncertainty, smoothing the images is an often used technique, and it has been empirically shown to yield good results, e.g., in \cite{Rueckert1999a}. We have examined the effect of image smoothing on NMI, and the results can be seen in \figurename~\ref{fig:imgsmooth} for PW and GPV respectively.
Furthermore, \figurename~\ref{fig:imgsmoothHRI} shows the estimated joint probability distribution for both PW and GPV. 
As can be seen, the distribution is more concentrated in a smaller area and NMI increases, when $\sigma$ is large.   The figures indicate that PW in general has a more pronounced peak than GPV for NMI, and that the optima is not shifted much over scales for this particular set of T1-weighted MRI of brains and using NMI.

\subsection{Intensity scale, $\beta$}
\begin{figure}
  \centering
  \includegraphics[width=0.64\linewidth]{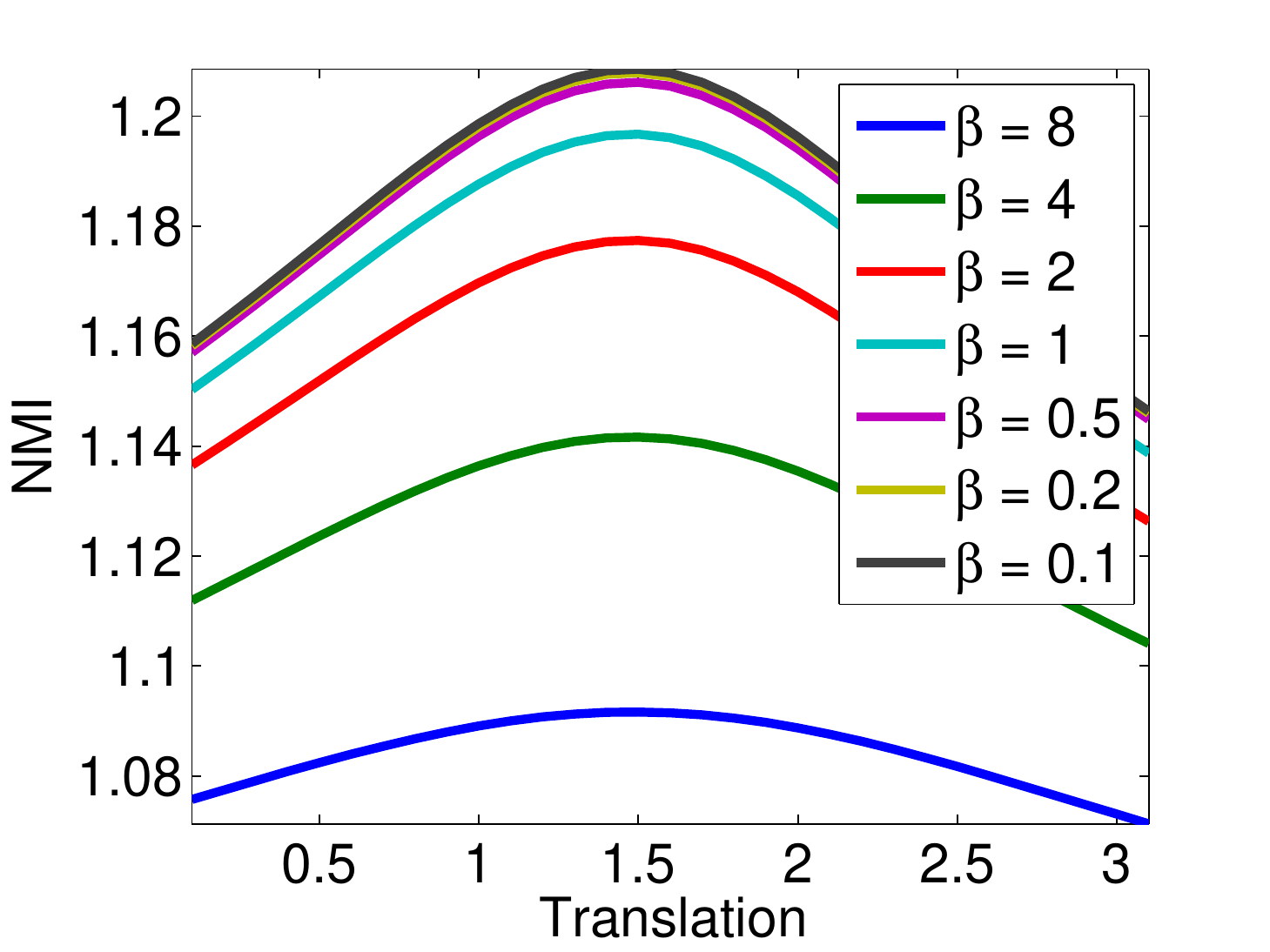}
  \caption{ The effect of varying $\beta$ for PW and NMI.}
  \label{fig:PWscales}
\end{figure}

\begin{figure}
  \centering
  \subfigure[][]{\includegraphics[width=0.48\linewidth]{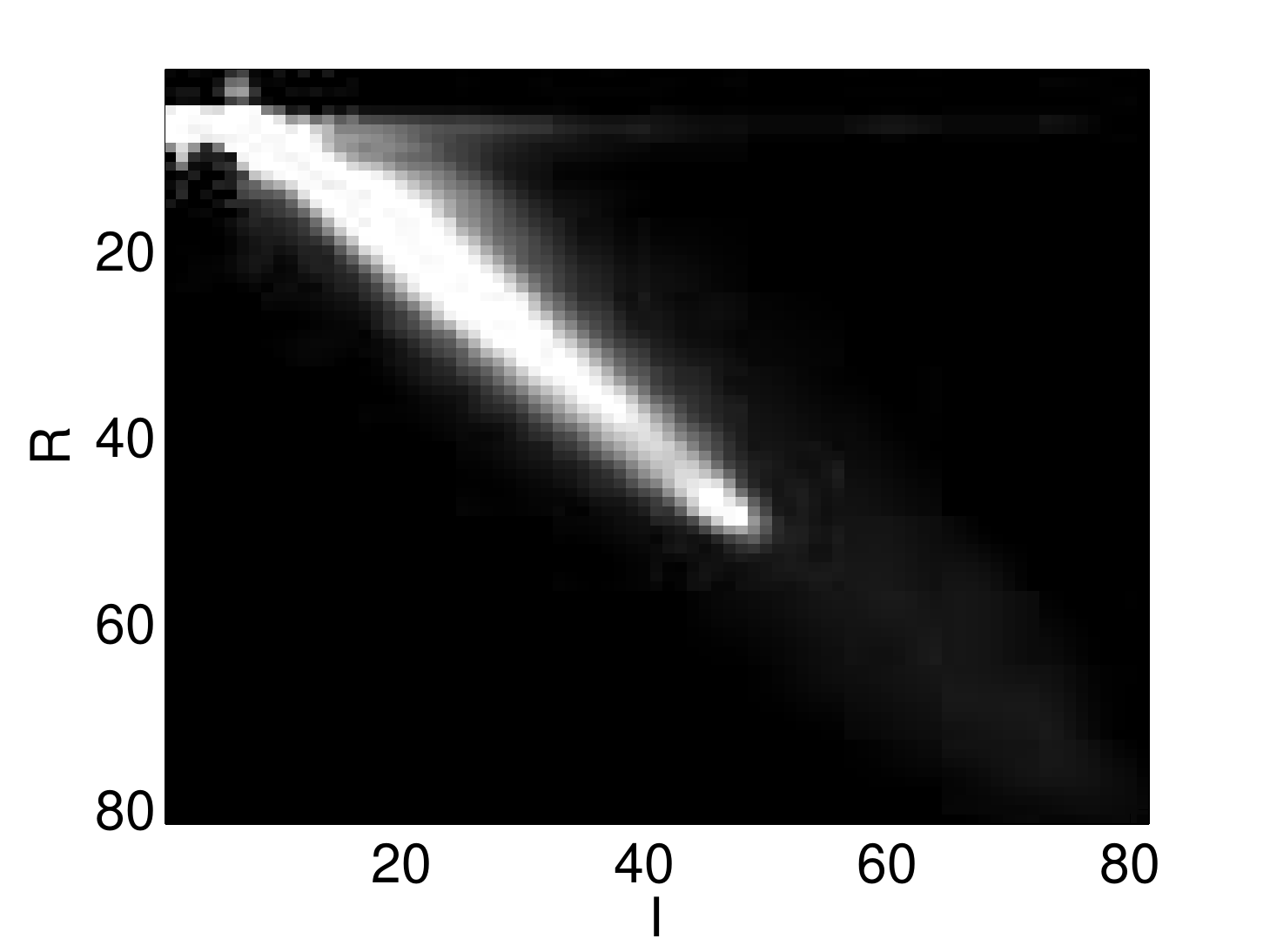}}
  \subfigure[][]{\includegraphics[width=0.48\linewidth]{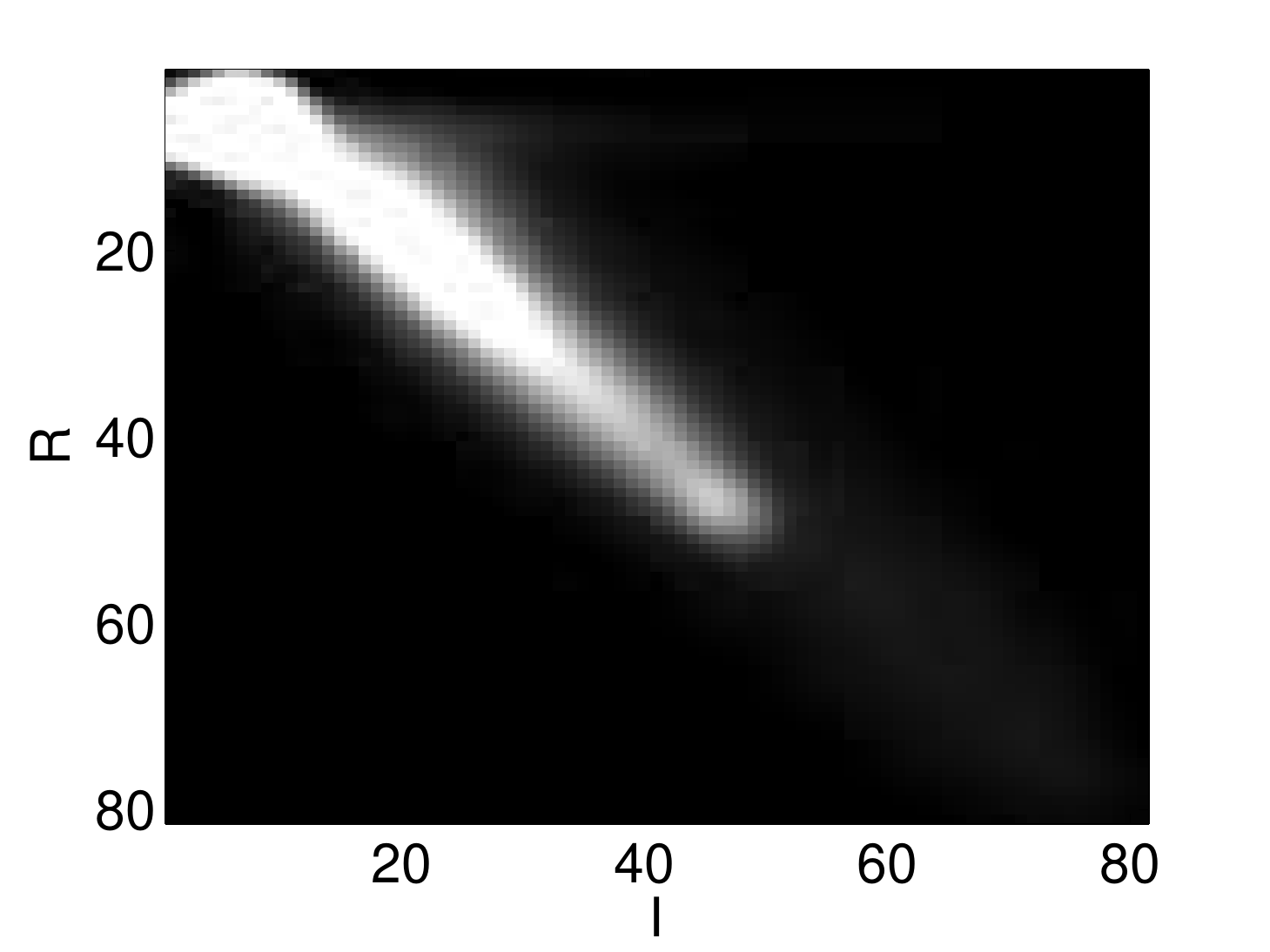}}
  \caption{ The effect of image smoothing with PW on the joint density estimate: The PW using $\sigma=1$ and (a) $\beta=0.5 $ and (b) $\beta=2$.}
  \label{fig:imgsmoothHRI:PW}
\end{figure}

The intensity scale controls the resolution in intensity domain, and as PW is a smoothing kernel in the intensity domain, then entropy is increased \cite{sporring.weickert99} proportionally to $\beta$. The smoothing disperses the densities within the joint density, thus decreases the overall NMI scores, as can be seen in \figurename~\ref{fig:PWscales}. 
The effect of $\beta$ on the joint density is illustrated in \figurename~\ref{fig:imgsmoothHRI:PW}. 
As expected, the joint histogram becomes more smooth as $\beta$ is increased.The consequence of increasing $\beta$ is that small scale changes in the image become neglectable (see Section~\ref{sec:notesOnScales}), whereas large changes are preserved, i.e., putting more emphasis on large gradients with increasing $\beta$. We have not included GPV in this experiment, however, GPV also has an intensity scale i.e. the width of its Boxcar function. 

\subsection{Integration scale, $\alpha$}
\begin{figure}
  \centering
  \includegraphics[width=0.64\linewidth]{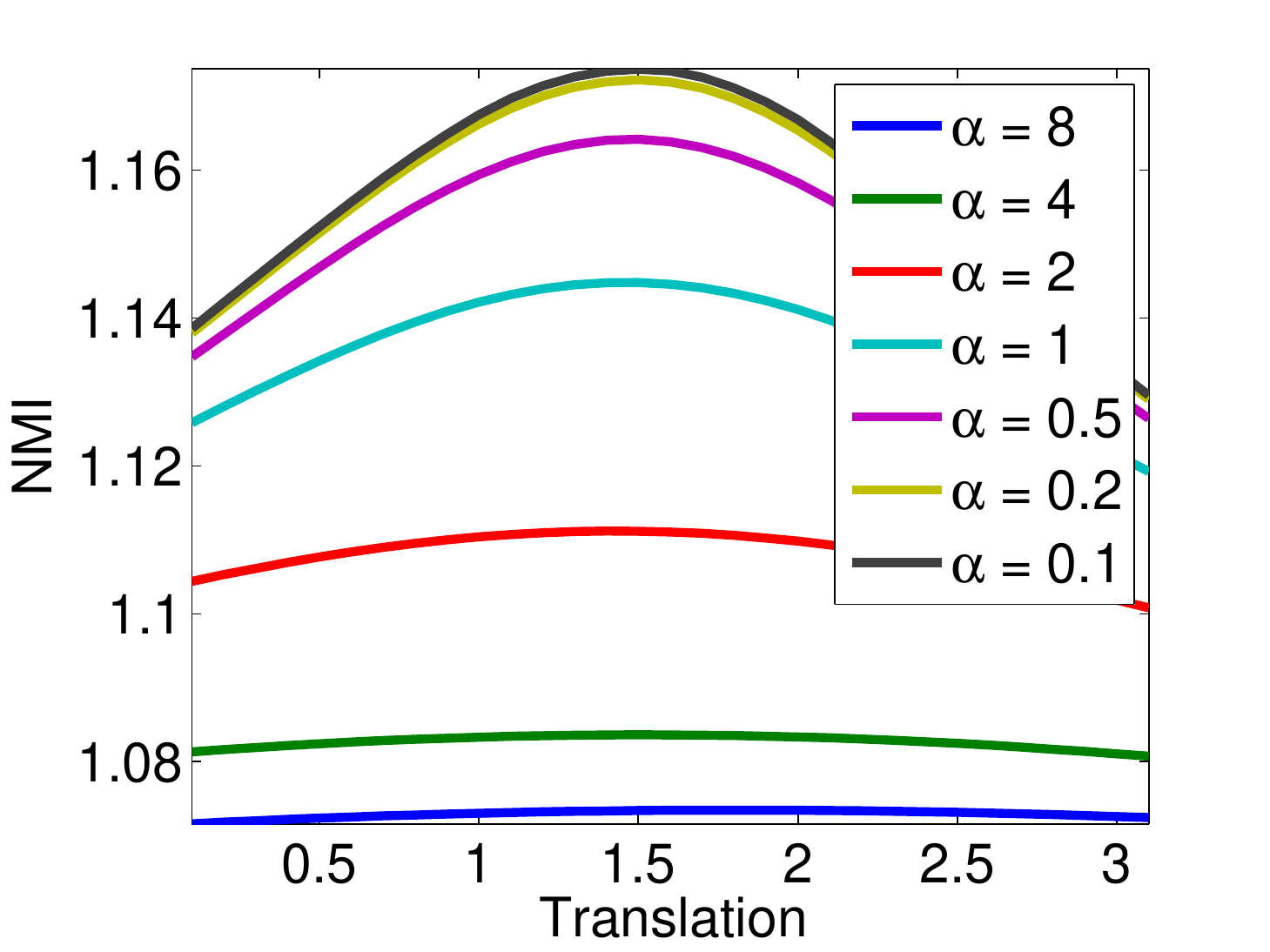}
  \caption{ The effect of varying $\alpha$ on the NMI functional using GPV with $\sigma=0.2$.}
  \label{fig:GPVscales}
\end{figure}

\begin{figure}
 \centering
  \subfigure[][GPV]{\label{fig:imgsmoothHRI:gpv_k02}\includegraphics[width=0.48\linewidth]{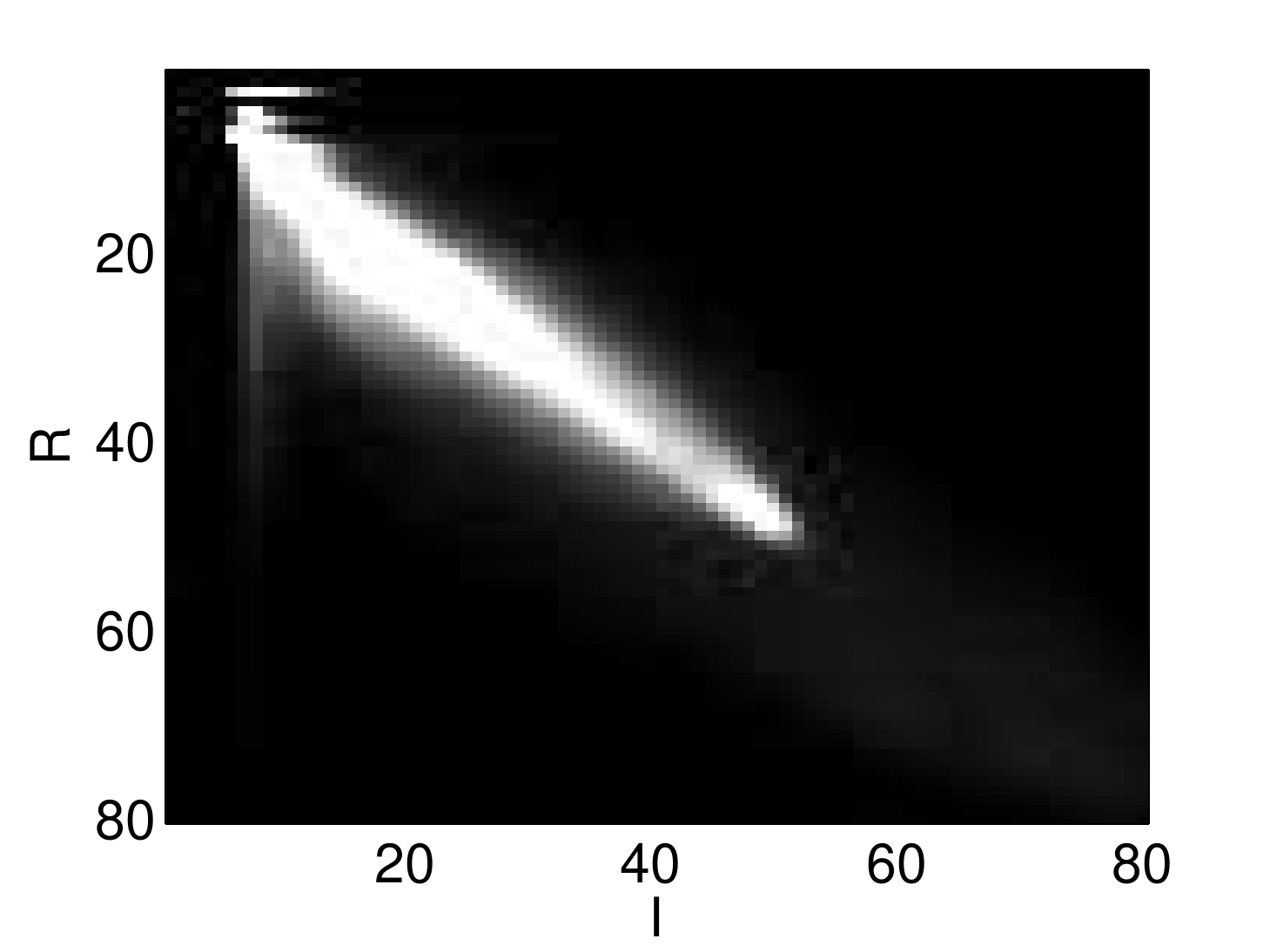}}
  \subfigure[][GPV]{\label{fig:imgsmoothHRI:gpv_k8}\includegraphics[width=0.48\linewidth]{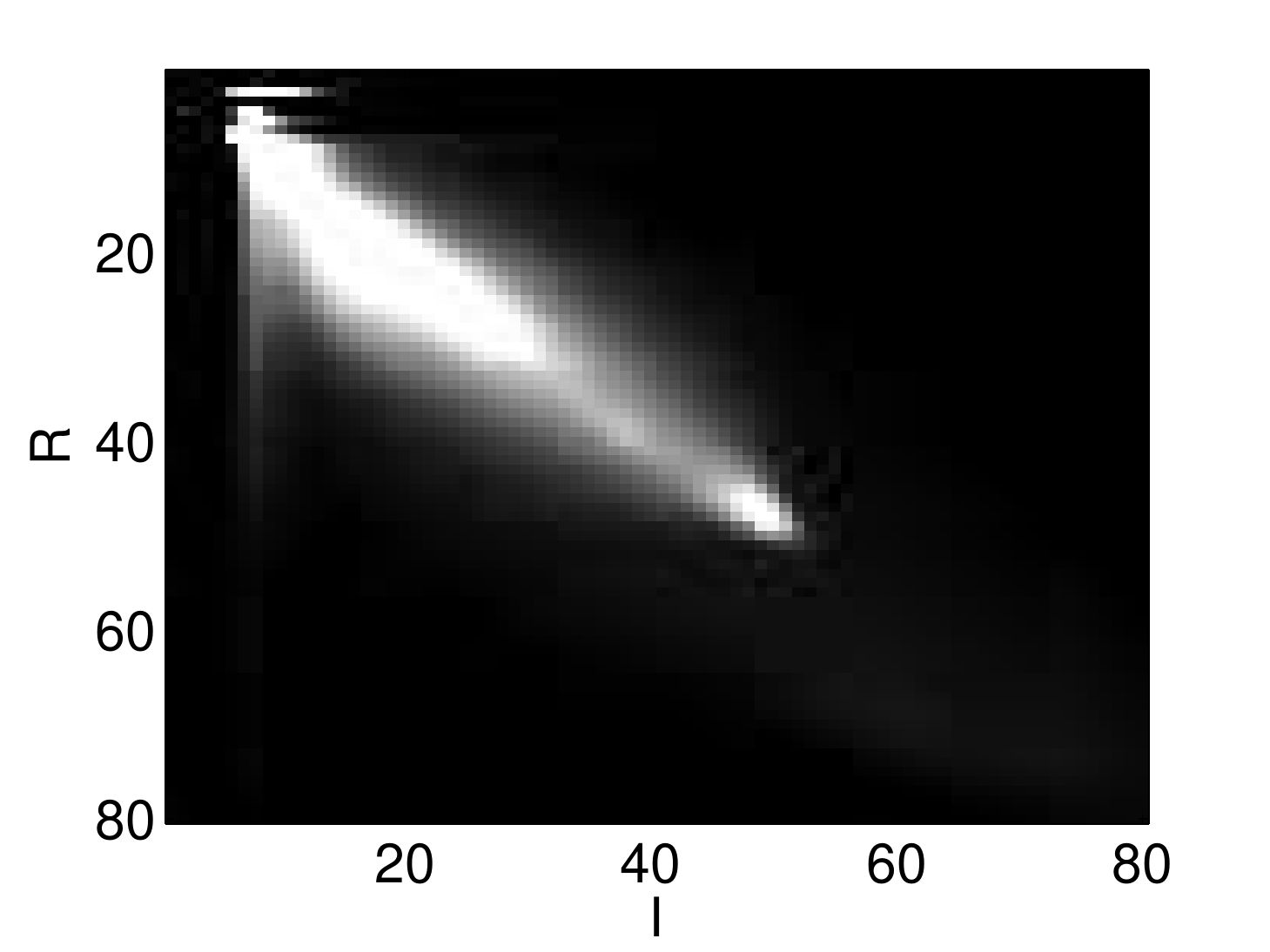}}
  \caption{ The effect of the integration scale on the joint density estimate for GPV and NMI using $\sigma=1$:  (a) $\alpha=0.5 $ and (b) $\alpha=2$.}
  \label{fig:imsmoothHRIGPV}
\end{figure}

PW is the special case of LOI, where $\alpha\rightarrow \infty$, thus a global density estimate, whereas GPV is an integration of local densities to become global. GPV uses a Boxcar function for $P$ and smoothes the isophotes with $W$, illustrated in \figurename~\ref{fig:iso:iso12} and~\ref{fig:iso:iso21}.  The effect of varying $\alpha$ on NMI using GPV is shown in \figurename~\ref{fig:GPVscales}.
It is seen that NMI decreases and becomes more dispersed as $\alpha$ is increased.  Comparing with \figurename~\ref{fig:PWscales} we note that the effect of $\alpha$ on GPV is similar to the effect of $\beta$ on PW:  it reduces the function value due to the dispersion effect.  Our theoretical investigation has revealed that smoothing is performed asymmetrically for GPV, and this is illustrated in \figurename~\ref{fig:imsmoothHRIGPV}, where we see horizontal dispersion but no vertical dispersion especially visible in the upper left corner.
Previous empirical investigations \cite{loeckx2006comparison} used the same B-spline kernel as PW $\beta$ and partial volume $\alpha$ and reported that PW is more precise, and that GPV has a larger convergence radius. From our experiments it is obvious that this difference is merely a consequence of the additional smoothing introduced by $W$ as discussed in Section~\ref{sec:notesOnScales}. This is supported by \figurename~\ref{fig:imgsmooth}: As can be seen the PW is significantly more peaked than the GPV, which appears superficially to be a smoothed version of PW.

The kernel $W$ can be used to describe local density estimates such as local MI or NMI \cite{hermosillo2002variational}, where each local histogram has its own NMI functional as in \eqref{eq:pdnonlinearIntegral}.

\subsection{Comparing GPV and PW by Scales}
The main difference between GPV and PW is: (1) the explicit modelling of the intensity coherence in PW with a Gaussian, thus assumes smooth images, where GPV has a disjoint view on the isophotes, i.e., no intensity coherence. (2) GPV views the template image intensities as a set of classes and the target as a continuous image, whereas PW views both images, template and target, as continuous.

\section{Fast Implementations}
\label{sec:implementation}
We use a quasi-Newton gradient descent algorithm for optimizing~\eqref{eq:F}.  This results in a very fast and general algorithm that with only a few changes works for many different loss-functions. 

In order to use quasi-Newton methods for optimization, we need to derive the gradient of~\eqref{eq:F} w.r.t.\ the parameters of the uniform cubic B-spline, $\mat\Phi$.  We use the notation of differentials, $dg(x)=Dg(x)\,dx$, where $D$ is the partial derivative operator.  Note that $d\vec x$ is a vector of differentials, not the wedge product of its elements as for integration.  Further, we will only write up non-zero terms that depend on $d\mat\Phi$. The differential of~\eqref{eq:F} is,
\begin{align}
  \label{eq:dF}
  d\mathcal{E}&=d\mathcal{M}+d\mathcal{S},
\end{align}
where arguments have been omitted for brevity.  Ignoring the regularization term we focus on the differential of the similarity measures. For~\eqref{eq:linearIntegral}, the differential is found to be,
\begin{align}
  \label{eq:dMlin}
  d\mathcal{M}_\text{lin} =\int_{\Gamma^2} F(i,j)dh_{I,J}\,di\wedge dj,
\end{align}
under the mild Leibnitz integration rule, and where
\begin{align}
    dh&=d\big(P(I(\vec x)-i)P(J(\vec x)-j)*W(\vec x)\big)\\
    &=\big(DP(I(\vec x)-i)dI\big)P(J(\vec x)-j)*W(\vec x),
\end{align}
avoiding irrelevant argument for brevity.  In contrast, the differential of~\eqref{eq:pdnonlinearIntegral} is $d\mathcal{M}_\text{lin} =\int_{\Omega} DF(\vec x, I(\vec x),J(\vec x))dI(\vec x)\,d\vec x$, where smoothness typically is imposed on $F$ and/or $I$. In comparison, our formulation~\eqref{eq:linearIntegral} naturally allows for the added smoothing in intensity and integration spaces and replaces technical difficulties in evaluating $DF$ with $Dh$. One advantage is thus that it becomes easier to compare loss-functions directly.  For~\eqref{eq:pdnonlinearIntegral} the differential is found to be,
\begin{equation}
\begin{split}
  \label{eq:dMnonlin}
  &d\mathcal{M}=\\
  &\int_{\Gamma^2} DF(\vec x, h_{I,J}(\vec x,i,j))\,dh_{I,J}(\vec x,i,j) \,d\vec x\wedge di\wedge dj,
  \end{split}
\end{equation}
similarly under the mild Leibnitz integration rule.  As shown in Section~\ref{sec:implementation}, the form of~\eqref{eq:dMnonlin} suggests only a slight computational overhead as compared to~\eqref{eq:dMlin}.  The derivatives for a range of $F$'s are given in \cite{sporring.darkner.11c}.

Using Leibniz integration rule, the differentials of the distributions are given as
\begin{align}
  &dp_I(i,\mat\Phi) = \frac{1}{|\Omega |}\int_\Omega dp_I(i|\vec x,\mat\Phi)\,d\vec x,\\
  &dp_I(i|\vec x,\mat\Phi)\simeq \frac{dh_I(i,\vec x,\mat\Phi)}{\int_\Gamma h_I(j,\vec x,\mat\Phi) dj}\nonumber\\&\qquad\qquad-\frac{h_I(i,\vec x,\mat\Phi)\int_\Gamma dh_I(j,\vec x,\mat\Phi) dj}{\left(\int_\Gamma h_I(j,\vec x,\mat\Phi) dj\right)^2},\\
  &dh_I(i,\vec x,\mat\Phi) = \left(dP(I(\vec x,\mat\Phi,\sigma)-i,\beta)*W(\vec x, \alpha)\right),
\end{align}
where irrelevant arguments have been omitted for brevity. Likewise, we have:
\begin{align}
  &dp_{I,R}(i,j) = \frac{1}{|\Omega |}\int_\Omega dp_{I,R}(i,j|\vec x)\,d\vec x,\\
  &dp_{I,R}(i,j|\vec x)\simeq \frac{dh_{I,R}(i,j,\vec x)}{\int_{\Gamma^2} h_{I,R}(k,l,\vec x)\,dk\wedge dl}\nonumber\\&\qquad\qquad-\frac{h_{I,R}(i,j,\vec x) \int_{\Gamma^2} dh_{I,R}(k,l,\vec x)\,dk\wedge dl}{\left(\int_{\Gamma^2} h_{I,R}(k,l,\vec  x)\,dk\wedge dl\right)^2},\\
  &dh_{I,R}(i,j,\vec x) = \nonumber\\&\big(dP(I(\vec \psi,\mat\Phi,\sigma)-i,\beta)\,P(J(\vec \psi,\sigma)-j,\beta)\big)*W(\vec x-\vec \psi, \alpha).
\end{align}
In the context of Locally orderless images, GPV can be derived as follows: 
\begin{align}
  dh_I &= d\left(P(I(\vec x,\mat\Phi,\sigma)-i,\beta)*W(\vec x, \alpha)\right)\\
  &= P(I(\tilde{\vec x},\mat\Phi,\sigma)-i,\beta)*\left(D_{\vec x}W(\vec x, \alpha) \right),
\end{align}
and the differential w.r.t.\ $x$ is found to be, 
\begin{align}
  &dh_{I,R}(i,j,\vec x,\alpha,\beta,\sigma) \nonumber\\
  &\qquad=P(J(\vec x,\sigma)-j,\beta)\bigg(\big(P(I(\vec\phi(\tilde{\vec x}),\sigma)-i,\beta) \big)\nonumber\\&\qquad\qquad*\big(D_{\vec x}W( x,\alpha)\big)\bigg).
\end{align}

In \figurename~\ref{fig:code} is shown the pseudo code for sum of squared differences using a spatial integration (SSD), Parzen window approximation of the general sum of p-norms (PNORM), Parzen window and Generalized partial volume approximation of normalized mutual information (PW and GPV). 
\begin{figure}
  \lstset{tabsize=2,basicstyle=\footnotesize}
  \lstset{emph={SSD}, emphstyle=\color{red},
    emph={[2]PW},emphstyle={[2]\color{blue}},
    emph={[3]GPV},emphstyle={[3]\color{green}},
    emph={[4]PNorm},emphstyle={[4]\color{cyan}}}
\begin{lstlisting}[frame=single]
# Given 2 images, I and R, and the determinant of the
# transformation, det, as a function of space, 
# calculate PW for NMI and PNorm, GPV for NMI and
# SSD, based on N image evaluation points, and
# M marginal and M^2 joint histogram bins. Flops are
# based on cubic B-splines

FOR N evaluation points
	calculate image spline coeff. 
	(60 flops)
	IF(SSD || PW  || PNorm)
		calculate derivative of image spline coeff. 
		(48 flops)
	FOR 64 combinations of image spline coeff.
		IF(SSD || PW  || PNorm)
			update image at evaluation point 
			(4 flops)
			update image gradient at evaluation point 
			(12 flops)
		IF(GPV)
			update histograms 
			(4 flops)
	IF(SSD)
		update residual 
		(2 flops)
	IF(PW || PNorm)
		calculate histogram spline coeff. 
		(20 flops)
		FOR 16 histogram spline coeff. 
			IF(PNorm)
				compute P-norm
				update residual
				update derivative
				(5 flops) 
			ELSE
			update histograms 
			(2 flops)
IF(PW || GPV)
	calculate NMI and derivative on histograms 
	(9*M^2+6M flops)
	FOR N evaluation points
		IF(GPV) 
			calculate derivative of image spline coeff. 
			(48 flops)
			FOR 64 combinations of image spline coeff.
				update derivative of histogram 
				(16 flops)
		IF(PW)
			FOR 16 histogram spline coeff.
				update derivative of histogram 
				(9 flops)
	update derivatives 
	(3 flops)
# Total flop usage:
#   SSD: 		1134N flops
#   PW:  		1331N +9M^2 +6M flops
#   PNorm:	1379N flops
#   GPV: 		1383N +9M^2 +6M flops
\end{lstlisting}
  \caption{\label{fig:code} Pseudo code for SSD, NMI using PW and GPV and P-Norm using PW.}
\end{figure}
Binary code interfacing to Matlab is available \cite{webpublishedCode}. All kernels used in our implementation are 3rd order uniform B-splines as well as Boxcar functions in order to reduce computational complexity.  The code assumes 3D images, cubic B-splines for all kernels, and $M$ bins in the histograms.  We assume that today's processors have equal processing time of, e.g., sum, log, sin etc.  From the pseudo code in \figurename~\ref{fig:code} and the annotated computational complexity, we see that PW and GPV have almost identical computational complexity. Results by actual implementations may vary, but in general the computing times for NMI using either GPV or PW are comparable in computational complexity to SSD using B-splines. W.r.t.\ memory, GPV requires $192 \times N\times 8$ bytes of memory to obtain the speed, where the PW only requires $8 \times N \times 8$ bytes (on 64-bit, double precision).

To substantiate our theoretical computation we have performed some empirical experiments.  First we note that the overhead of GPV and PW is in general small. The histogram calculations will only dominate in the special case of a small number of samples and many histogram bins.   We have compared computational complexity empirically for PW and GPV registration and SSD. We use cubic B-spline for $K$, $P$, and $W$, and histograms with $256$ bins for marginal histograms and $256^2$ for the joint histograms. We perform the computations on a laptop with i7-core Q820 (Quad-core) operating at 1.7 GHz and 12 GB shared memory. All similarity measures have been implemented in parallel using the Intel Threading Building Blocks library. As the code runs multi-threaded we believe that most of the $13\%$ overhead seen in Table~\ref{tab:performance} comes from the threads, which are initialized twice as many times in PW as in SSD. 
\begin{table}
  \center
  \begin{tabular}{|c||c||c||c|}
    \hline
    \# samples&Similarity measure&SSD&PW\\
    \hline
    1000000&Avg. execution time (in sec) &1.21&1.63\\
    \cline{2-4}
    &Relative exec. time to SSD&1&1.34\\
    \cline{2-4}
    &Theoretic relative exec. time to SSD&1&1.17\\
    \hline
  &Overhead&1&1.13\\
    \hline
  \end{tabular}
  \caption{The table shows the average execution time across 100 function evaluations of SSD vs PW-NMI for 1000000 points using 256 bins.}
  \label{tab:performance}
\end{table}
Furthermore, thread blocking can cause further latency during histogram update, thus the estimated times for single threaded implementation are very close to our estimate for large $N$.

\section{Conclusion}
\label{sec:conclusion}
We have introduced Locally Orderless Registration, a framework that encompasses most of the currently used similarity measures. Our framework allow us to divide a wide range of similarity measures into 3 categories from simple global linear measures such as the P-norm or Huber norm over non-linear global measures such as Correlation Coefficient, Mutual Information and Normalized Mutual Information  to position dependent schemes, e.g., Correlation Ratio and spatially encoded Mutual Information. All of these measures or any combination are formulated in a scale space over measurement, intensity, and integration space offering the flexibility to easily create application specific similarity measures in a smooth formulation well suited for gradient based schemes. We have presented a thorough analysis of the scales in the different spaces both theoretically through the moments of the density distribution and a simple local image model and through rigorous empirically experiments. 

We have extended our previous work \cite{darkner.sporring.ipmi2011} on the difference between Parzen Window and Generalized Partial Volume. Our analysis clearly shows that Generalized Partial Volume is an asymmetric density estimator not suited for problems that require inverse consistency. Depending on the smoothing, we have shown that this error can become larger than a single voxel. Thus, Generalized Partial Volume achieves its computational speed by making an approximation to the local histogram and by using 0-order B-spline as Parzen estimator. In \cite{loeckx2006comparison} it is reported that Parzen Window is more accurate than Generalized Partial Volume for kernels $W$ with $\alpha>0$, and we show that this is due to the difference in smoothing and not due to the properties of the two density estimators. And worse, Generalized Partial Volume measures the dissimilarity of the images at two different scales, and thus the effect becomes more pronounced with increased $\alpha$ - histograms of larger areas. 

Our theoretical analysis of the computational complexity and the memory requirements demonstrate that the Parzen window is more attractive for intensity based registration.

We believe that the choice of density estimator should be based on the particular application. If no intensity context exist, e.g., registering classes (Correlation Ratio) Generalized Partial Volume is to be preferred over Parzen Window due to the fact that coherence in the joint and marginal density distributions is a key assumption in Parzen Window. However, if intensity images are to be registered, and computational efficiency or inverse consistency is a desired property, then our analysis reveals that Parzen Window is a far more attractive density estimator as compared to Generalized Partial Volume.

\section*{Acknowledgment} \addcontentsline{toc}{section}{Acknowledgment}
Sune Darkner would like to thank the Oticon foundation for supporting his work through the grant for the project "A Fast and Personalized Biomechanical Model".

\bibliographystyle{IEEEtran}
\bibliography{bibtex}
\begin{IEEEbiography}[{\includegraphics[width=1in,height=1.25in,clip,keepaspectratio]{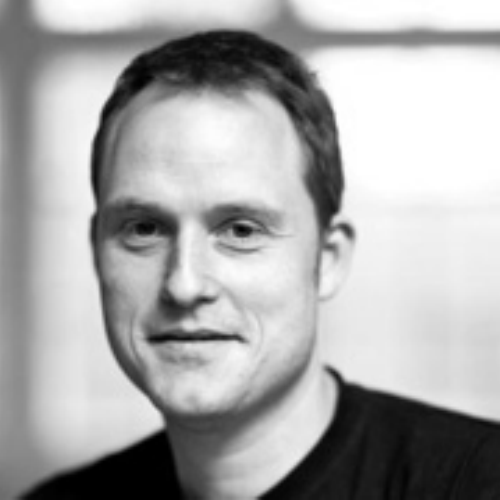}}]{Sune Darkner}
received his M.Sc. in applied mathematics in 2004 and founded a software company making databases for the telecommunication industry. In collaboration with Oticon A/S a large hearing aid manufacture he received his industrial PhD degree in 'Shape and Deformation Analysis of the Human Ear Canal'
in 2009 from the department of Informatics and Mathematical Modelling at the Technical University of Denmark (DTU). After a position at an Energy company as data analyst he rejoined the department of Informatics and Mathematical Modelling, Image group at DTU in 2009 as a post doc. Currently, he is Post Doc. in image analysis at DIKU Image, University of Copenhagen group. Research interest include, image registration and classification, optimization and regularization and computational physics.
\end{IEEEbiography}
\begin{IEEEbiography}[{\includegraphics[width=1in,height=1.25in,clip,keepaspectratio]{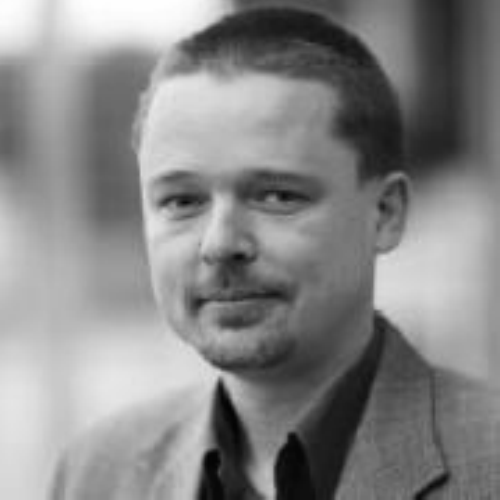}}]{Jon Sporring}
received his Master and Ph.D. degree from the Department of Computer Science, University of Copenhagen , Denmark in 1995 and 1998, respectively. Part of his Ph.D. program was carried out at IBM Research Center, Almaden, California, USA. Following his Ph.D, he worked as a visiting researcher at the Computer Vision and Robotics Lab at Foundation for Research \& Technology - Hellas, Greece, and as assistant research professor at 3D-Lab, School of Dentistry, University of Copenhagen. Since 2003 he has been employed as associate professor at the Department of Computer Science, University of Copenhagen, for a 1 year period starting in 2008 he was a part time Senior Researcher at Nordic Bioscience, and since 2008 he has been Vice-Chair for Research at Department of Computer Science. His primary research field is Computer Science, particularly image processing, computer graphics, information theory, and pattern recognition. This includes computer assisted physiotherapy, medical image registration, photon-mapping and analysis of Chronic Obstructive Pulmonary Disease (COPD).
\end{IEEEbiography}
\end{document}